\documentclass[english]{article}
\usepackage[T1]{fontenc}
\usepackage[latin9]{inputenc}
\usepackage{babel}
\usepackage{array}
\usepackage{float}
\usepackage{booktabs}
\usepackage{multirow}
\usepackage{amsmath}
\usepackage{amsthm}
\usepackage{amssymb}
\usepackage{cancel}
\usepackage{graphicx}
\PassOptionsToPackage{normalem}{ulem}
\usepackage{ulem}
\usepackage[unicode=true]
 {hyperref}

\makeatletter

\providecommand{\tabularnewline}{\\}
\floatstyle{ruled}
\newfloat{algorithm}{tbp}{loa}
\providecommand{\algorithmname}{Algorithm}
\floatname{algorithm}{\protect\algorithmname}

\PassOptionsToPackage{numbers, compress}{natbib}

\usepackage[preprint]{neurips_2024}

\PassOptionsToPackage{numbers, compress}{natbib}

\usepackage[unicode=true]{hyperref}       
\usepackage{url}            
\usepackage{booktabs}       
\usepackage{amsfonts}       
\usepackage{nicefrac}       
\usepackage{microtype}      
\usepackage{xcolor} 

\usepackage{microtype}
\usepackage{graphicx}
\usepackage{mathtools}

\usepackage{xcolor}
\usepackage{soul}

\definecolor{red}{HTML}{FF0000}
\definecolor{lawngreen}{HTML}{7CFC00}
\definecolor{orange}{HTML}{FFA500}
\definecolor{peru}{HTML}{CD853F}
\definecolor{gray}{HTML}{808080}
\definecolor{lightgray}{RGB}{240,240,240}

\usepackage{algpseudocode}

\@ifundefined{showcaptionsetup}{}{%
 \PassOptionsToPackage{caption=false}{subfig}}
\usepackage{subfig}
\makeatother

\begin{document}
\renewcommand{\contentsname}{Table of Content for Appendix}

\title{Bidirectional Diffusion Bridge Models}

\author{\textbf{Duc Kieu, Kien Do, Toan Nguyen, Dang Nguyen, Thin Nguyen}\\
Applied Artificial Intelligence Institute (A2I2), Deakin University,
Australia\\
\emph{\{v.kieu, k.do, k.nguyen, d.nguyen, thin.nguyen\}@deakin.edu.au}}

\maketitle
\global\long\def\Expect{\mathbb{E}}
\global\long\def\Real{\mathbb{R}}
\global\long\def\Data{\mathcal{D}}
\global\long\def\Loss{\mathcal{L}}
\global\long\def\Normal{\mathcal{N}}
\global\long\def\softmax{\text{softmax}}
\global\long\def\ELBO{\text{ELBO}}
\global\long\def\argmin#1{\underset{#1}{\text{argmin}}}
\global\long\def\argmax#1{\underset{#1}{\text{argmax}}}
\global\long\def\diag{\text{diag}}

\begin{abstract}
Diffusion bridges have shown potential in paired image-to-image
(I2I) translation tasks. However, existing methods are limited by
their unidirectional nature, requiring separate models for forward
and reverse translations. This not only doubles the computational
cost but also restricts their practicality. In this work, we introduce
the Bidirectional Diffusion Bridge Model (BDBM), a scalable approach
that facilitates bidirectional translation between two coupled distributions
using a single network. BDBM leverages the Chapman-Kolmogorov Equation
for bridges, enabling it to model data distribution shifts across
timesteps in both forward and backward directions by exploiting the
interchangeability of the initial and target timesteps within this
framework. Notably, when the marginal distribution given endpoints
is Gaussian, BDBM's transition kernels in both directions possess
analytical forms, allowing for efficient learning with a single network.
We demonstrate the connection between BDBM and existing bridge methods,
such as Doob\textquoteright s $h$-transform and variational approaches,
and highlight its advantages. Extensive experiments on high-resolution
I2I translation tasks demonstrate that BDBM not only enables bidirectional
translation with minimal additional cost but also outperforms state-of-the-art
bridge models. Our source code is available at \href{https://github.com/kvmduc/BDBM}{https://github.com/kvmduc/BDBM}.

\end{abstract}
\addtocontents{toc}{\protect\setcounter{tocdepth}{-1}}

\section{Introduction}

Diffusion models (DMs)~\cite{sohl2015deep,song2019generative,ho2020denoising}
have emerged as a powerful class of generative models, surpassing
GANs~\cite{goodfellow2014generative} and VAEs~\cite{kingma2013auto}
in generating high-quality data~\cite{dhariwal2021diffusion}. These
models learn to transform a Gaussian prior distribution into the data
distribution through iterative denoising steps. However, the Gaussian
prior assumption in diffusion models limits their application, particularly
in image-to-image (I2I) translation \cite{pix2pix2017}, where the
distributions of the two domains are non-Gaussian.

A straightforward solution is to incorporate an additional condition
related to one domain into diffusion models for guidance \cite{choi2021ilvr,saharia2022palette}.
This approach often overlooks the marginal distribution of each domain,
which may hinder its generalization ability, especially when the two
domains are diverse and significantly different. In contrast, methods
that construct an ODE flow \cite{lipman2022flow,liu2022flow,albergo2023stochastic}
or a Schrödinger bridge \cite{bortoli2021diffusion,shi2023diffusion,kim2024unpaired}
between two domains focus mainly on matching the marginal distributions
at the boundaries, neglecting the relationships between samples from
the two domains. Consequently, these methods are not well-suited for
paired I2I tasks.

To solve the paired I2I problem, recent methods \cite{LiuW0l23,zhou2024denoising}
leverage knowledge of the target sample $y$ in the pair $\left(x,y\right)$
and utilized Doob's $h$-transform \cite{doob1984classical} to construct
a bridge that converges to $y$. This involves learning either the
$h$ function \cite{SomnathPHM0B23} or the score function of the
$h$-transformed SDE \cite{zhou2024denoising}, both of which depend
on $y$. Other methods \cite{LiX0L23} extend the unconditional variational
framework for diffusion models to a conditional one given $y$ for
constructing such bridges, thereby learning a backward transition
distribution conditioned on $y$. Despite their success in capturing
the correspondence between $x$ and $y$, these methods share a common
limitation: they can only generate data in \emph{one direction}, from
$y$ to $x$. For the reverse from $x$ to $y$, a separate bridge
must be trained with $x$ being the target, which doubles computational
resources and modeling complexity. We argue that real-world applications
would greatly benefit from bidirectional generative models capable
of transitioning between two distributions using a single model.

Therefore, we introduce a novel bridge model called \textbf{\textit{\emph{B}}}\textit{\emph{idirectional
}}\textbf{\textit{\emph{D}}}\textit{\emph{iffusion }}\textbf{\textit{\emph{B}}}\textit{\emph{ridge
}}\textbf{\textit{\emph{M}}}\textit{\emph{odel}} (BDBM) that enables
\emph{bidirectional} transitions between two coupled distributions
using \textit{only a single network}. Our bridge is built on a framework
that highlights the symmetry between forward and backward transitions.
By utilizing the Chapman-Kolmogorov Equation (CKE) for conditional
Markov processes, we transform the problem of modeling the conditional
distribution $p\left(x_{T}=y|x_{0}=x\right)$ into modeling the forward
transition from $p\left(x_{t}|x,y\right)$ to $p\left(x_{s}|x,y\right)$
- the marginal distributions at times $t$ and $s$ ($0\leq t<s\leq T$)
of a \emph{double conditional Markov process} (DCMP) between two endpoints
$x,y\sim p\left(x,y\right)$. Given the interchangeability of the
two marginal distributions, we can model the conditional distribution
$p\left(x_{0}=x|x_{T}=y\right)$ simply by learning the backward transition
from $p\left(x_{s}|x,y\right)$ to $p\left(x_{t}|x,y\right)$ without
altering the DCMP. Notably, the forward and backward transition distributions
of the DCMP are connected through Bayes' rule and can be expressed
analytically as Gaussian distributions when the DCMP is a diffusion
process. This insight motivates us to reparameterize models of the
forward and backward transition distributions in a way that they share
a common term. Therefore, we can use a single network for modeling
this term and train it with a unified objective for both directions.

We evaluate our method on four popular paired I2I translation datasets~\cite{pix2pix2017,diode_dataset}
with image sizes up to 256$\times$256, considering both pixel and
latent spaces. Experimental results demonstrate that BDBM surpasses
state-of-the-art (SOTA) unidirectional diffusion bridge models in
terms of visual quality (measured by FID) and perceptual similarity
(measured by LPIPS) of generated samples, while requiring similar
or even fewer training iterations. These promising results showcase
the clear advantages of our method, which not only facilitates bidirectional
translation at minimal additional cost but also improves performance.

\section{Preliminaries\label{sec:Preliminaries}}

\subsection{Markov Processes and Diffusion Processes}

A Markov process is a stochastic process satisfying the Markov property,
i.e., the future (state) is independent of the past given the present:
\[
p\left(x_{s}|x_{t},x_{u}\right)=p\left(x_{s}|x_{t}\right)
\]
where $x_{u}$, $x_{t}$, $x_{t}$ denote random states at times $u$,
$t$, $s$ satisfying that $0\leq u<t<s$. Here, $p\left(x_{s}|x_{t}\right)$
is the transition distribution of the Markov process. 

Diffusion processes are special cases of Markov processes where the
transition distribution is typically a Gaussian distribution. A diffusion
process can be either discrete-time \cite{ho2020denoising} or continuous-time
\cite{song2020score}. A continuous-time diffusion process can be
described by the following (forward) stochastic differential equation
(SDE):
\begin{align}
dX_{t} & =\mu\left(t,X_{t}\right)dt+\sigma\left(t,X_{t}\right)dW_{t}\label{eq:SDE_1}
\end{align}
where $W_{t}$ denotes the Wiener process (aka Brownian motion) at
time $t$. Eq.~\ref{eq:SDE_1} can be solved via simulation provided
that the distribution of $X_{0}$ is known. One can derive the \emph{forward}
and \emph{backward Kolmogorov equations} (KFE and KBE) for this SDE
as follows:
\begin{equation}
\text{KFE:}\ \ \frac{\partial p\left(t,x\right)}{\partial t}=\mathcal{G}^{*}p\left(t,x\right);\ p\left(0,\cdot\right)\ \text{is given}\label{eq:KFE_1}
\end{equation}
\begin{equation}
\text{KBE:}\ \ \frac{\partial p\left(T,y|t,x\right)}{\partial t}=-\mathcal{G}p\left(T,y|t,x\right);\ p\left(T,\cdot\right)\ \text{is given}\label{eq:KBE_1}
\end{equation}
where $\mathtt{\mathcal{G}}$ denotes the \emph{generator} corresponding
to the SDE in Eq.~\ref{eq:SDE_1} and $\mathcal{G}^{*}$ is the adjoint
of $\mathtt{\mathcal{G}}$. When $\sigma\left(t,x\right)$ is a scalar
depending only on $t$ (i.e., $\sigma\left(t,x\right)\equiv\sigma\left(t\right)$,
for a real-valued function $f$, $\mathcal{G}f\left(t,x\right)$ and
$\mathcal{G}^{*}f\left(t,x\right)$ are given by:
\begin{align*}
\mathcal{G}f\left(t,x\right) & ={\nabla f\left(t,x\right)}^{\top}\mu\left(t,x\right)+\frac{{\sigma\left(t\right)}^{2}}{2}\Delta f\left(t,x\right)\\
\mathcal{G}^{*}f\left(t,x\right) & =-\nabla\cdot\left(f\left(t,x\right)\mu\left(t,x\right)\right)+\frac{{\sigma\left(t\right)}^{2}}{2}\Delta f\left(t,x\right)
\end{align*}
where $\nabla\cdot$ and $\Delta$ denote the divergence and Laplacian,
respectively.

\subsection{Chapman-Kolmogorov Equations\label{subsec:Chapman-Kolmogorov-Equations} }

A Markov process can be described via the \emph{Chapman-Kolmogorov
equation} (CKE) \cite{karush1961chapman} as follows:
\begin{equation}
p\left(x_{s}|x_{t}\right)=\int p\left(x_{s}|x_{r}\right)p\left(x_{r}|x_{t}\right)dx_{r}\label{eq:CKE}
\end{equation}

\noindent which holds for all times $t$, $r$, $s$ satisfying that
$0\le t<r<s\leq T$. The CKE in Eq.~\ref{eq:CKE} can be considered
as the integral form of the KFE and KBE in Eqs.~\ref{eq:KFE_1},
\ref{eq:KBE_1}. Compared to the Kolmogorov equations, the CKE is
easier to work with since (i) it does not involve the partial derivatives
of the transition kernel, (ii) it is applicable to both continuous-
and discrete-time Markov processes, and (iii) it encapsulates both
forward and backward transitions. Regarding the last point, we can
apply Eq.~\ref{eq:CKE} either in the forward manner (from $0$ to
$T$) to evaluate the distribution of the next state $x_{s}$ given
the distribution of the current state $x_{t}$:
\begin{equation}
p\left(x_{s}|x_{0}\right)=\int p\left(x_{s}|x_{t}\right)p\left(x_{t}|x_{0}\right)dx_{t};\ p\left(x_{t}|x_{0}\right)\ \text{is given}\label{eq:forward_CKE}
\end{equation}
or in the backward manner (from $T$ to $0$) to evaluate the distribution
of the previous state $x_{t}$ given the distribution of current state
$x_{s}$:
\begin{equation}
p\left(x_{T}|x_{t}\right)=\int p\left(x_{T}|x_{s}\right)p\left(x_{s}|x_{t}\right)dx_{s};\ p\left(x_{T}|x_{s}\right)\ \text{is given}\label{eq:backward_CKE}
\end{equation}
In the discrete-time setting, Eq.~\ref{eq:forward_CKE} can be interpreted
as given a Markov process with $p\left(x_{t+1}|x_{t}\right)$ specified
for every time $t$. If we have known the marginal distribution $p\left(x_{t}|x_{0}\right)$
at time $t$, then by solving the CKE forwardly, we can compute $p\left(x_{t+1}|x_{0}\right)$
at time $t+1$. Similarly, in Eq.~\ref{eq:backward_CKE}, if we have
known $p\left(x_{T}|x_{t+1}\right)$ at time $t+1$, then by solving
the CKE backwardly, we can compute $p\left(x_{T}|x_{t}\right)$ at
time $t$. For example, in DDPM \cite{ho2020denoising}, given $p\left(x_{t}|x_{0}\right)=\Normal\left(x_{t}\mid\sqrt{\bar{\alpha}_{t}}x_{0},\left(1-\bar{\alpha}_{t}\right)\mathrm{I}\right)$
and $p\left(x_{t+1}|x_{t}\right)=\Normal\left(x_{t+1}\mid\sqrt{1-\beta_{t+1}}x_{t},\beta_{t+1}\mathrm{I}\right)$,
we can use Eq.~\ref{eq:forward_CKE} to compute $p\left(x_{t+1}|x_{0}\right)$
as $\Normal\left(x_{t+1}\mid\sqrt{\bar{\alpha}_{t+1}}x_{0},\left(1-\bar{\alpha}_{t+1}\right)\mathrm{I}\right)$.

Interestingly, the backward CKE in Eq.~\ref{eq:backward_CKE} can
be written in another way according to Bayes' rule:
\begin{equation}
p\left(x_{t}|x_{T}\right)=\int p\left(x_{t}|x_{s}\right)p\left(x_{s}|x_{T}\right)dx_{s};\ p\left(x_{s}|x_{T}\right)\ \text{is given}\label{eq:backward_CKE_2}
\end{equation}
The mathematical derivation is detailed in Appdx.~\ref{subsec:Derivation-of-the-backward-CKE}.
Eq.~\ref{eq:backward_CKE_2} is akin to the forward CKE in Eq.~\ref{eq:forward_CKE}
but in reverse time.

\section{Method}

\noindent 
\begin{figure*}[t]
\begin{centering}
\includegraphics[width=0.8\textwidth]{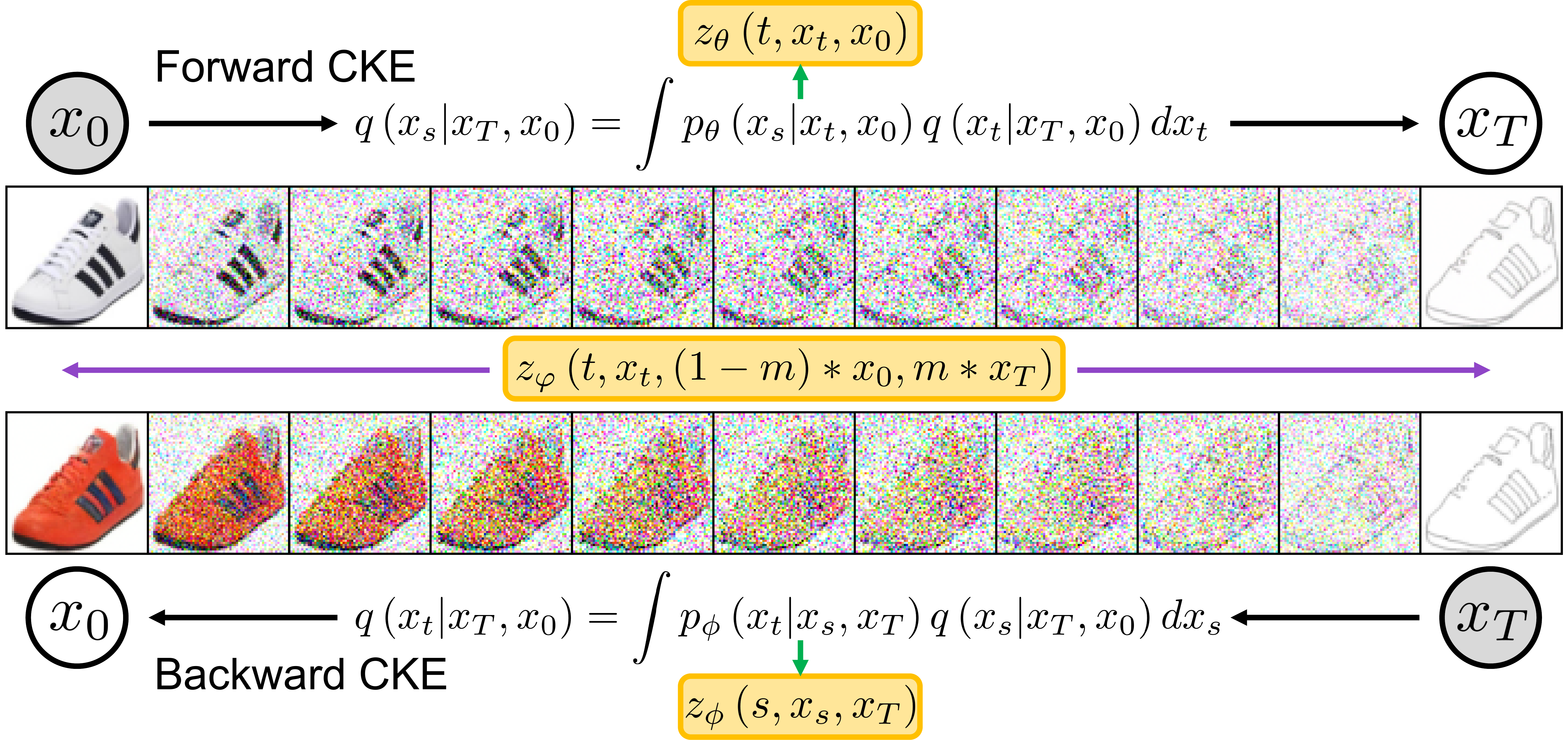}
\par\end{centering}
\caption{An illustration of \emph{Bidirectional Diffusion Bridge Models (BDBM)}.
Instead of learning two separate models $z_{\theta}\left(t,x_{t},x_{0}\right)$
and $z_{\phi}\left(s,x_{s},x_{T}\right)$ for the forward and backward
transitions, we learn a single model $z_{\varphi}\left(t,x_{t},\left(1-m\right)*x_{0},m*x_{T}\right)$
with a binary mask $m$ that enables transition in both directions.
Grey and white nodes denote initial and generated samples, respectively.}
\end{figure*}

\subsection{Chapman-Kolmogorov Equations for Bridges\label{subsec:Chapman-Kolmogorov-equations-for-bridges}}

In many real-world problems (e.g., paired/unpaired image translation),
the joint boundary distribution $p\left(y_{A},y_{B}\right)$ of samples
from two domains $A$, $B$ is given in advance rather than just either
$p\left(y_{A}\right)$ or $p\left(y_{B}\right)$, and we need to design
a stochastic process such that if we start from $y_{A}$ ($y_{B}$),
we should reach $y_{B}$ ($y_{A}$) with a predefined probability
$p\left(y_{B}|y_{A}\right)$ ($p\left(y_{A}|y_{B}\right)$). Such
stochastic processes are referred to as \emph{stochastic bridges}
or simply \emph{bridges }\cite{liu2022let,LiuW0l23,LiX0L23,zhou2024denoising}.
In this section, we will develop mathematical models for stochastic
bridges based on the CKEs for Markov processes in Section \ref{sec:Preliminaries}.

Without loss of generality, we associate two domains $A$, $B$ with
samples at time $0$, $T$, respectively. Let $\left\{ X_{t}\right\} $
be a stochastic process in which the initial distribution $p\left(x_{0}|y_{A}\right)$
is a Dirac distribution at $y_{A}$ (i.e., $p\left(x_{0}|y_{A}\right)=\delta_{y_{A}}$).
To conform to the notation used in prior works, we denotes $\hat{x}_{0}:=y_{A}$.
The symbol $^{\wedge}$ indicates that $\hat{x}_{0}$ is a specified
value rather than a random state like $x_{0}$\footnote{This allows us to write $p\left(x_{0}|\hat{x}_{0}\right)=\delta_{\hat{x}_{0}}=\delta_{y_{A}}$}.
For modeling simplicity, we assume that the process is a \emph{conditional
Markov process} described by the following CKE:
\begin{equation}
p\left(x_{v}|x_{t},\hat{x}_{0}\right)=\int p\left(x_{v}|x_{s},\text{\ensuremath{\hat{x}_{0}}}\right)p\left(x_{s}|x_{t},\text{\ensuremath{\hat{x}_{0}}}\right)dx_{s}\label{eq:bridge_CKE}
\end{equation}
where $t<s<v$. Interestingly, if we start this process from an arbitrary
time $t$ with the marginal distribution $p\left(x_{t}|\hat{x}_{0}\right)$,
we will always reach the same distribution at time $T>t$. To see
this, we represent $p\left(x_{T}|\hat{x}_{0}\right)$ using two different
starting times $t,s$ with $0\leq t<s<T$ as follows:
\begin{align}
 & p\left(x_{T}|\hat{x}_{0}\right)\nonumber \\
=\  & \int p\left(x_{T}|x_{s},\hat{x}_{0}\right)p\left(x_{s}|\hat{x}_{0}\right)dx_{s}\\
=\  & \int p\left(x_{T}|x_{s},\hat{x}_{0}\right)\left(\int p\left(x_{s}|x_{t},\hat{x}_{0}\right)p\left(x_{t}|\hat{x}_{0}\right)dx_{t}\right)dx_{s}\\
=\  & \int\underbrace{\left(\int p\left(x_{T}|x_{s},\hat{x}_{0}\right)p\left(x_{s}|x_{t},\hat{x}_{0}\right)dx_{s}\right)}_{p\left(x_{T}|x_{t},\hat{x}_{0}\right)}p\left(x_{t}|\hat{x}_{0}\right)dx_{t}\\
=\  & \int p\left(x_{T}|x_{t},\hat{x}_{0}\right)p\left(x_{t}|\hat{x}_{0}\right)dx_{t}
\end{align}
The intuition here is the associativity of the (functional) inner
product between $p\left(x_{T}|x_{s},\hat{x}_{0}\right)$, $p\left(x_{s}|x_{t},\hat{x}_{0}\right)$,
and $p\left(x_{t}|\hat{x}_{0}\right)$. Let us consider the problem
of learning the transition kernel $p_{\theta}\left(x_{s}|x_{t},\hat{x}_{0}\right)$
of the above process such that $p_{\theta}\left(x_{T}|\hat{x}_{0}\right)$
equals $p\left(y_{B}|y_{A}\right)$. Clearly, $p_{\theta}\left(x_{s}|x_{t},\hat{x}_{0}\right)$
should satisfy:
\begin{equation}
p\left(x_{T}|x_{t},\hat{x}_{0}\right)=\int p\left(x_{T}|x_{s},\hat{x}_{0}\right)p_{\theta}\left(x_{s}|x_{t},\hat{x}_{0}\right)dx_{s}\label{eq:bridge_CKE_not_learnable}
\end{equation}
for all $0\leq t<s$. However, Eq.~\ref{eq:bridge_CKE_not_learnable}
does not facilitate easy learning of $p_{\theta}\left(x_{s}|x_{t},\hat{x}_{0}\right)$
because determining the values of $p\left(x_{T}|x_{t},\hat{x}_{0}\right)$
and $p\left(x_{T}|x_{s},\hat{x}_{0}\right)$ can be challenging in
practice, which usually requires another parameterized model. Therefore,
we utilize the equivalent formula below:
\begin{equation}
q\left(x_{s}|\hat{x}_{T},\hat{x}_{0}\right)=\int p_{\theta}\left(x_{s}|x_{t},\hat{x}_{0}\right)q\left(x_{t}|\hat{x}_{T},\hat{x}_{0}\right)dx_{t}\label{eq:bridge_CKE_learnable}
\end{equation}
with $\hat{x}_{T}\sim p\left(y_{B}|y_{A}\right)$. The derivation
of Eq.~\ref{eq:bridge_CKE_learnable} is presented in Appdx.~\ref{subsec:Proof_Chapman-Kolmogorov-equations-for-bridges}.
Eq.~\ref{eq:bridge_CKE_learnable} implies that if we can construct
a \emph{double conditional Markov process} between $\hat{x}_{0}$
and $\hat{x}_{T}$ such that the marginal distribution at time $t$
is $q\left(x_{t}|\hat{x}_{T},\hat{x}_{0}\right)$ and the two boundary
distributions at times $0$ and $T$ are Dirac distributions at $\hat{x}_{0}$
and $\hat{x}_{T}$, respectively (i.e., $q\left(x_{0}|\hat{x}_{T},\hat{x}_{0}\right)=\delta_{\hat{x}_{0}}\left(x_{0}\right)$
and $q\left(x_{T}|\hat{x}_{T},\hat{x}_{0}\right)=\delta_{\hat{x}_{T}}\left(x_{T}\right)$),
then by learning $p_{\theta}\left(x_{s}|x_{t},\hat{x}_{0}\right)$
to match the transition probability $q\left(x_{s}|x_{t},\hat{x}_{T},\hat{x}_{0}\right)$
of this process, $p_{\theta}\left(x_{s}|x_{t},\hat{x}_{0}\right)$
will serve as the transition probability of a bridge starting from
$\hat{x}_{0}$ and ending at $\hat{x}_{T}$ with $p\left(\hat{x}_{T}|\hat{x}_{0}\right)=p\left(y_{B}|y_{A}\right)$.
There are various ways to align $p_{\theta}\left(x_{s}|x_{t},\hat{x}_{0}\right)$
with $q\left(x_{s}|x_{t},\hat{x}_{T},\hat{x}_{0}\right)$ and the
loss below is commonly used due to its link to variational inference
\cite{ho2020denoising,LiX0L23}:
\begin{equation}
\Loss=\Expect_{t,s,\hat{x}_{0},\hat{x}_{T}}\left[D_{\text{KL}}\left(q\left(x_{s}|x_{t},\hat{x}_{T},\hat{x}_{0}\right)\|p_{\theta}\left(x_{s}|x_{t},\hat{x}_{0}\right)\right)\right]\label{eq:loss_bridge_1}
\end{equation}
where $t\sim\mathcal{U}\left(0,T-\Delta t\right)$, $s=t+\Delta t$,
$\hat{x}_{0}\sim p\left(y_{A}\right)$, $\hat{x}_{T}\sim p\left(y_{B}|y_{A}\right)$.

In practice, we often choose $q\left(x_{t}|\hat{x}_{T},\hat{x}_{0}\right)$
and $q\left(x_{s}|\hat{x}_{T},\hat{x}_{0}\right)$ to be Gaussian
distributions, which results in $q\left(x_{s}|x_{t},\hat{x}_{T},\hat{x}_{0}\right)$
being a Gaussian. Therefore, if $p_{\theta}\left(x_{s}|x_{t},\hat{x}_{0}\right)$
is also modeled as a Gaussian distribution, then Eq.~\ref{eq:loss_bridge_1}
can be expressed in closed-form. Details about this will be presented
in Section~\ref{subsec:Generalized-Diffusion-Bridge-Models}. In
Appdx.~\ref{subsec:Connection-between-the-CKE-framework}, we provide
the connection of this framework to variational inference, score matching,
and Doob's $h$-transform.

\subsection{Generalized Diffusion Bridge Models\label{subsec:Generalized-Diffusion-Bridge-Models}}

To simplify our notation, from this section onward, we will use $x_{0}$,
$x_{T}$ in place of $\hat{x}_{0}$, $\hat{x}_{T}$ in the conditional
distributions $q\left(x_{t}|\hat{x}_{0},\hat{x}_{T}\right)$ and $q\left(x_{s}|x_{t},\hat{x}_{0},\hat{x}_{T}\right)$
with a note that they should be interpreted as specified values rather
than random states. As discussed in Section~\ref{subsec:Chapman-Kolmogorov-equations-for-bridges},
$q\left(x_{t}|x_{0},x_{T}\right)$ should be chosen as a Gaussian
distribution with zero variance at $t\in\left\{ 0,T\right\} $ to
facilitate learning the transition kernel. A general formula of $q\left(x_{t}|x_{0},x_{T}\right)$
is $q\left(x_{t}|x_{0},x_{T}\right)=\Normal\left(\alpha_{t}x_{0}+\beta_{t}x_{T},\sigma_{t}^{2}\mathrm{I}\right)$
where $\alpha_{t},\beta_{t},\sigma_{t}$ are continuously differentiable
functions of $t\in\left[0,T\right]$ satisfying $\alpha_{0}=\beta_{T}=1$
and $\alpha_{T}=\beta_{0}=\sigma_{0}=\sigma_{T}=0$. According to
this formula, $x_{t}\sim q\left(x_{t}|x_{0},x_{T}\right)$ can be
computed as follows:
\begin{equation}
x_{t}=\alpha_{t}x_{0}+\beta_{t}x_{T}+\sigma_{t}z\label{eq:ref_bridge_xt_decomp}
\end{equation}
with $z\sim\Normal\left(0,\mathrm{I}\right)$. Similarly, we have
$q\left(x_{s}|x_{0},x_{T}\right)=\Normal\left(\alpha_{s}x_{0}+\beta_{s}x_{T},\sigma_{s}^{2}\mathrm{I}\right)$.
This means $q\left(x_{s}|x_{t},x_{0},x_{T}\right)$ has the form $\Normal\left(x_{s}\big|\mu\left(s,t,x_{t},x_{0},x_{T}\right),\delta_{s,t}^{2}\mathrm{I}\right)$
where:
\begin{align}
 & \mu\left(s,t,x_{t},x_{0},x_{T}\right)\nonumber \\
=\  & \alpha_{s}x_{0}+\beta_{s}x_{T}+\sqrt{\sigma_{s}^{2}-\delta_{s,t}^{2}}\frac{\left(x_{t}-\alpha_{t}x_{0}-\beta_{t}x_{T}\right)}{\sigma_{t}}\label{eq:ref_forward_mean_1}\\
=\  & \frac{\beta_{s}}{\beta_{t}}x_{t}+\left(\alpha_{s}-\alpha_{t}\frac{\beta_{s}}{\beta_{t}}\right)x_{0}+\left(\sqrt{\sigma_{s}^{2}-\delta_{s,t}^{2}}-\sigma_{t}\frac{\beta_{s}}{\beta_{t}}\right)z\label{eq:ref_forward_mean_2}
\end{align}
and $\delta_{s,t}$ can vary arbitrarily within the (half-)interval
$[0,\sigma_{s})$. Eq.~\ref{eq:ref_forward_mean_2} is derived from
Eq.~\ref{eq:ref_forward_mean_1} by setting $x_{T}=\frac{1}{\beta_{t}}\left(x_{t}-\alpha_{t}x_{0}-\sigma_{t}z\right)$
according to Eq.~\ref{eq:ref_bridge_xt_decomp}.

To match $p_{\theta}\left(x_{s}|x_{t},x_{0}\right)$ with $q\left(x_{s}|x_{t},x_{0},x_{T}\right)$
($t<s$), we should be able to infer $x_{T}$ from $x_{t}$, $x_{0}$
in $p_{\theta}\left(x_{s}|x_{t},x_{0}\right)$. A straightforward
approach is to formulate $p_{\theta}\left(x_{s}|x_{t},x_{0}\right)$
as $\Normal\left(x_{s}\big|\mu_{\theta}\left(s,t,x_{t},x_{0}\right),\delta_{s,t}^{2}\mathrm{I}\right)$
and reparameterize $\mu_{\theta}\left(s,t,x_{t},x_{0}\right)$ to
match with $\mu\left(s,t,x_{t},x_{0},x_{T}\right)$, where $x_{T}$
replaced by its approximation $x_{T,\theta}\left(t,x_{t},x_{0}\right)$
in Eq.~\ref{eq:ref_forward_mean_1} (or $z$ replaced by $z_{\theta}\left(t,x_{t},x_{0}\right)$
in Eq.~\ref{eq:ref_forward_mean_2}). When $z_{\theta}\left(t,x_{t},x_{0}\right)$
is modeled, we regard $x_{T,\theta}\left(t,x_{t},x_{0}\right)$ as
$\frac{1}{\beta_{t}}\left(x_{t}-\alpha_{t}x_{0}-\sigma_{t}z_{\theta}\left(t,x_{t},x_{0}\right)\right)$,
and the loss in Eq.~\ref{eq:loss_bridge_1} simplifies to:
\begin{equation}
\Loss=\Expect_{t,x_{0},x_{T},z,x_{t}}\left[w_{t}\left\Vert z_{\theta}\left(t,x_{t},x_{0}\right)-z\right\Vert _{2}^{2}\right]\label{eq:loss_bridge_2}
\end{equation}
where $t\sim\mathcal{U}\left(0,T\right)$, $x_{0}\sim p\left(y_{A}\right)$,
$x_{T}\sim p\left(y_{B}|y_{A}\right)$, $z\sim\Normal\left(0,\mathrm{I}\right)$,
and $x_{t}=\alpha_{t}x_{0}+\beta_{t}x_{T}+\sigma_{t}z$. $w_{t}$
is set to 1 in our work. This loss is a weighted version of the score
matching loss for bridges \cite{zhou2024denoising}. Once $z_{\theta}$
has been learned, it will approximate $-\sigma_{t}\nabla\log p\left(x_{t}|x_{0}\right)$,
and $x_{T,\theta}$ derived from $z_{\theta}$ approximates $\Expect_{p\left(x_{T}|x_{t},x_{0}\right)}\left[x_{T}\right]$
due to Tweedie's formula for bridges (Appdx\@.~\ref{subsec:Tweedie's-formula-for-bridges}).

\begin{table*}
\begin{centering}
\begin{tabular}{cccccccccc}
\toprule 
\multirow{2}{*}{Model} & \multicolumn{3}{c}{Edges$\rightarrow$Shoes$\times64$} & \multicolumn{3}{c}{Edges$\rightarrow$Handbags$\times64$} & \multicolumn{3}{c}{Normal$\rightarrow$Outdoor$\times256$}\tabularnewline
\cmidrule{2-10} 
 & FID $\downarrow$ & IS $\uparrow$ & LPIPS $\downarrow$ & FID $\downarrow$ & IS $\uparrow$ & LPIPS $\downarrow$ & FID $\downarrow$ & IS $\uparrow$ & LPIPS $\downarrow$\tabularnewline
\midrule
\midrule 
BBDM & 2.11 & 3.23 & \uline{0.05} & 6.38 & 3.71 & 0.19 & 8.79 & 5.48 & 0.29\tabularnewline
\midrule 
$\text{I}^{2}\text{SB}$ & 2.14 & \textbf{3.41} & 0.06 & 6.05 & \uline{3.73} & 0.17 & \uline{5.48} & 5.71 & 0.37\tabularnewline
\midrule 
DDBM & 6.42 & 3.26 & 0.12 & 3.89 & 3.58 & 0.23 & 6.16 & 5.74 & 0.35\tabularnewline
\midrule
\midrule 
BDBM-1 (ours) & \uline{1.78} & \uline{3.28} & 0.07 & \uline{3.83} & 3.71 & \uline{0.11} & 7.17 & \textbf{5.97} & \textbf{0.11}\tabularnewline
\midrule 
BDBM (ours) & \textbf{1.06} & \uline{3.28} & \textbf{0.02} & \textbf{3.06} & \textbf{3.74} & \textbf{0.08} & \textbf{4.67} & \uline{5.91} & \uline{0.16}\tabularnewline
\bottomrule
\end{tabular}
\par\end{centering}
\caption{Quantitative comparison between BDBM and unidirectional bridge models
on translation tasks from sketch/normal maps to color images. The
best results are highlighted in bold, while the second-best results
are underlined.\label{tab:main_result_1}}
\end{table*}

\subsection{Bidirectional Diffusion Bridge Models\label{subsec:Bidirectional-Diffusion-Bridge-Models}}

Leaning $p_{\theta}\left(x_{s}|x_{t},x_{0}\right)$ with $t<s$ in
Section~\ref{subsec:Generalized-Diffusion-Bridge-Models} leads to
a bridge that maps samples at time $0$ (domain $A$) to those at
time $T$ (domain $B$). Unfortunately, we cannot travel in the reverse
direction (i.e., generate $x_{0}$ from $x_{T}$) with this bridge.
It is because the reverse transition kernel derived from $p_{\theta}\left(x_{s}|x_{t},x_{0}\right)$
requires the knowledge of $x_{0}$, which is not available if starting
from time $T$. A straightforward solution to this problem is constructing
another bridge with $x_{T}$ as the source by learning $p_{\phi}\left(x_{t}|x_{s},x_{T}\right)$
($t<s$). This results in two separate models for forward and backward
travels, which doubles the resources for training and deployment.
To overcome this limitation, we propose a novel Bidirectional Diffusion
Bridge Model (BDBM) that enables bidirectional travel while requiring
the training of only a single network. In our model, $p_{\theta}\left(x_{s}|x_{t},x_{0}\right)$
and $p_{\phi}\left(x_{t}|x_{s},x_{T}\right)$ are transition kernels
operating in opposite directions along the same bridge that connects
$x_{0}$ and $x_{T}$. Due to the interchangeability between $x_{t}$
and $x_{s}$ in Eq.~\ref{eq:bridge_CKE_learnable}, it follows that
if $p_{\theta}\left(x_{s}|x_{t},x_{0}\right)$ approximates $q\left(x_{s}|x_{t},x_{0},x_{T}\right)$,
then $p_{\phi}\left(x_{t}|x_{s},x_{T}\right)$ should approximate
$q\left(x_{t}|x_{s},x_{0},x_{T}\right)$, which is derived from $q\left(x_{s}|x_{t},x_{0},x_{T}\right)$
via the Bayes' rule:
\begin{equation}
q\left(x_{t}|x_{s},x_{0},x_{T}\right)=q\left(x_{s}|x_{t},x_{0},x_{T}\right)\frac{q\left(x_{t}|x_{0},x_{T}\right)}{q\left(x_{s}|x_{0},x_{T}\right)}\label{eq:Bayes_rule_bridge}
\end{equation}
Since $q\left(x_{t}|x_{0},x_{T}\right)$, $q\left(x_{s}|x_{0},x_{T}\right)$,
and $q\left(x_{s}|x_{t},x_{0},x_{T}\right)$ are Gaussian distributions
specified in Eqs.~\ref{eq:ref_bridge_xt_decomp}, \ref{eq:ref_forward_mean_1},
$q\left(x_{t}|x_{s},x_{0},x_{T}\right)$ is also a Gaussian distribution
of the form $\Normal\left(x_{t}\Big|\tilde{\mu}\left(t,s,x_{s},x_{0},x_{T}\right),\frac{\delta_{s,t}^{2}\sigma_{t}^{2}}{\sigma_{s}^{2}}\mathrm{I}\right)$
with $\tilde{\mu}\left(t,s,x_{s},x_{0},x_{T}\right)$ given by:
\begin{align}
 & \tilde{\mu}\left(t,s,x_{s},x_{0},x_{T}\right)\nonumber \\
=\  & \alpha_{t}x_{0}+\beta_{t}x_{T}+\sigma_{t}\sqrt{\sigma_{s}^{2}-\delta_{s,t}^{2}}\frac{\left(x_{s}-\alpha_{s}x_{0}-\beta_{s}x_{T}\right)}{\sigma_{s}^{2}}\label{eq:ref_backward_mean_1}\\
=\  & \frac{\alpha_{t}}{\alpha_{s}}x_{s}+\left(\beta_{t}-\beta_{s}\frac{\alpha_{t}}{\alpha_{s}}\right)x_{T}+\left(\frac{\sigma_{t}\sqrt{\sigma_{s}^{2}-\delta_{s,t}^{2}}}{\sigma_{s}}-\sigma_{s}\frac{\alpha_{t}}{\alpha_{s}}\right)z'\label{eq:ref_backward_mean_2}
\end{align}
where Eq.~\ref{eq:ref_backward_mean_2} is derived from Eq.~\ref{eq:ref_backward_mean_1}
by setting $x_{0}=\frac{1}{\alpha_{s}}\left(x_{s}-\beta_{s}x_{T}-\sigma_{s}z'\right)$.
We can align $p_{\phi}\left(x_{t}|x_{s},x_{T}\right)$ with $q\left(x_{t}|x_{s},x_{0},x_{T}\right)$
by reparameterizing the mean $\tilde{\mu}_{\phi}\left(t,s,x_{s},x_{T}\right)$
of $p_{\phi}\left(x_{t}|x_{s},x_{T}\right)$ such that it has the
same formula as $\tilde{\mu}\left(t,s,x_{s},x_{0},x_{T}\right)$ in
Eq.~\ref{eq:ref_backward_mean_1} but with $x_{0}$ replaced by $x_{0,\phi}\left(s,x_{s},x_{T}\right)$
(or $z'$ replaced by $z_{\phi}\left(s,x_{s},x_{T}\right)$ in Eq.~\ref{eq:ref_backward_mean_2}). 

In the case where $p_{\theta}\left(x_{s}|x_{t},x_{0}\right)$ and
$p_{\phi}\left(x_{t}|x_{s},x_{T}\right)$ are modeled via $z_{\theta}\left(t,x_{t},x_{0}\right)$
and $z_{\phi}\left(s,x_{s},x_{T}\right)$, respectively, it is possible
to use a single network $z_{\varphi}$ instead of two separate networks
$z_{\theta}$ and $z_{\phi}$ because they both represent the same
noise variable $z\sim\Normal\left(0,\mathrm{I}\right)$ (given $t=s$).
To deal with the problem that the forward transition depends on $x_{0}$
while the backward transition depends on $x_{T}$, we feed both $x_{0}$
and $x_{T}$ as inputs to $z_{\varphi}$ and mask one of them using
a mask $m$ associated with the transition direction. This results
in the model $z_{\varphi}\left(t,x_{t},\left(1-m\right)*x_{0},m*x_{T}\right)$
where $m=0$ (1) if we move forward from 0 to $T$ (backward from
$T$ to 0). We learn $z_{\varphi}$ by minimizing the following loss:
\begin{align}
\Loss_{\text{BDBM}}= & \Expect_{t,x_{0},x_{T},z,x_{t},m}\left[w_{t}\left\Vert z_{\varphi}\left(t,x_{t},\left(1-m\right)*x_{0},m*x_{T}\right)-z\right\Vert _{2}^{2}\right]\label{eq:loss_BDBM}
\end{align}
where $x_{0}$, $x_{T}$, $t$, $z$, $x_{t}$ are sampled in the
same way as in Eq.~\ref{eq:loss_bridge_2}, and the mask $m$ is
sampled from the Bernoulli distribution with $p\left(m=1\right)=0.5$.

On the other hand, when $x_{T,\theta}\left(t,x_{t},x_{0}\right)$
and $x_{0,\phi}\left(s,x_{s},x_{T}\right)$ serve as the parameterized
models for $p_{\theta}\left(x_{s}|x_{t},x_{0}\right)$ and $p_{\phi}\left(x_{t}|x_{s},x_{T}\right)$,
respectively, we propose to use a unified model to predict $x_{0}+x_{T}$.
We denote this model as $s_{\varphi}\left(t,x_{t},\left(1-m\right)*x_{0},m*x_{T}\right)$
and learn it with the loss:
\begin{align}
\Loss_{\text{BDBM}}^{(2)} & =\Expect_{t,x_{0},x_{T},z,x_{t},m}\Big[w_{t}\big\| s_{\varphi}\left(t,x_{t},\left(1-m\right)*x_{0},m*x_{T}\right)-\left(x_{0}+x_{T}\right)\big\|_{2}^{2}\Big]\label{eq:loss_BDBM_2}
\end{align}
When traveling from 0 to $T$ (from $T$ to 0), we set $m$ to $0$
(1) and use $s_{\varphi}\left(t,x_{t},x_{0},0\right)-x_{0}$ ($s_{\varphi}\left(s,x_{s},0,x_{T}\right)-x_{T}$)
to mimic $x_{T,\theta}\left(t,x_{t},x_{0}\right)$ ($x_{0,\phi}\left(s,x_{s},x_{T}\right)$).
We can also train $s_{\varphi}$ to predict $x_{0}+x_{T}$. In Appdx.~\ref{subsec:Training-and-sampling_alg},
we provide detailed training and sampling algorithms for BDBM. We
also discuss several important variants of BDBM in Appdx.~\ref{subsec:Special-cases-of-BDBM}.

\section{Experiments}

\begin{figure*}
\begin{centering}
\includegraphics[width=1\textwidth]{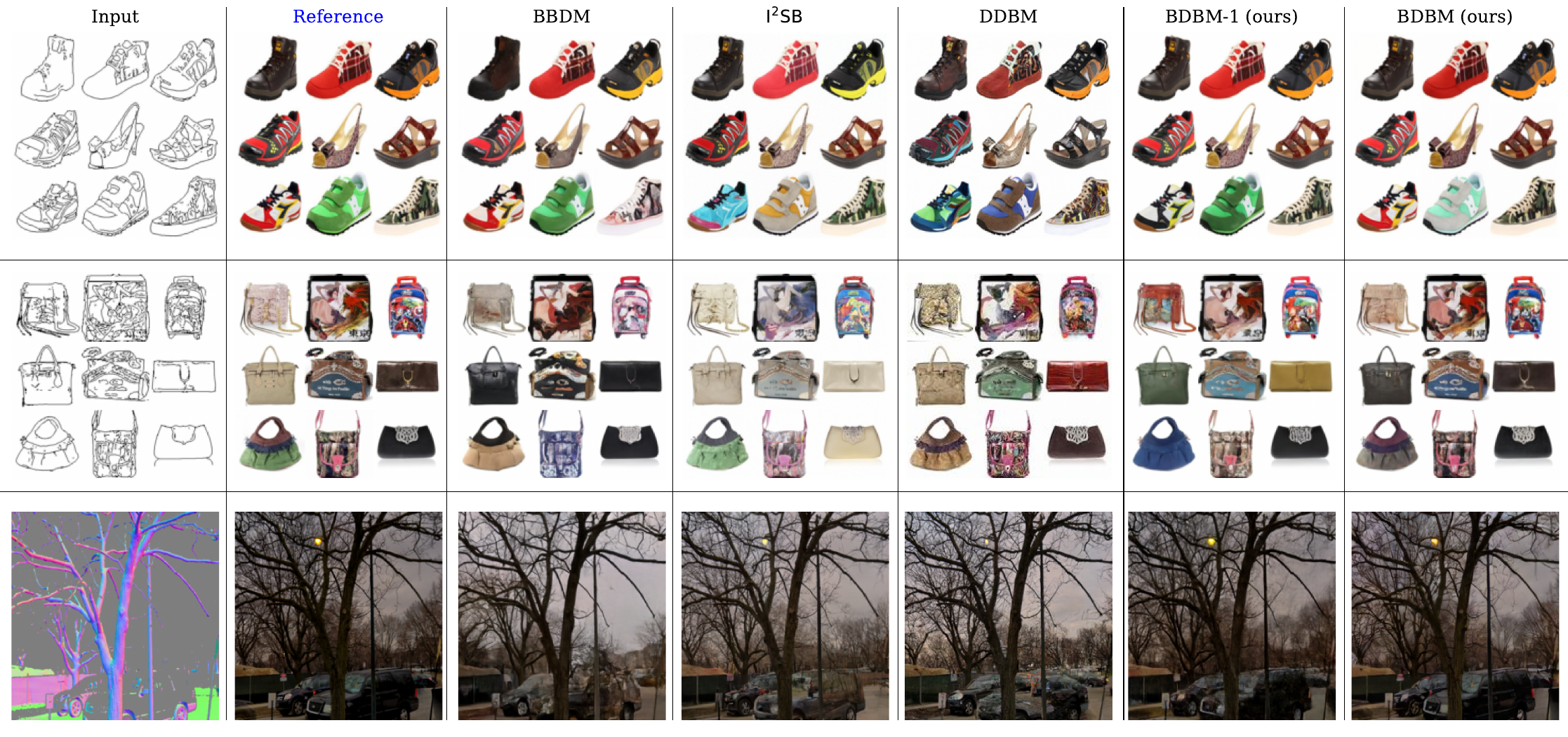}
\par\end{centering}
\caption{Images generated by BDBM and unidirectional baselines in the Edges$\rightarrow$Shoes,
Edges$\rightarrow$Handbags, and Normal$\rightarrow$Outdoor translation
tasks.\label{fig:main_qualitative_results}}
\end{figure*}

\subsection{Experimental Settings}

\subsubsection{Datasets and evaluation metrics}

We validate our method on 4 paired image-to-image (I2I) translation
datasets namely Edges$\leftrightarrow$Shoes, Edges$\leftrightarrow$Handbags,
DIODE Outdoor \cite{diode_dataset}, and Night$\leftrightarrow$Day
\cite{pix2pix2017}. Following \cite{zhou2024denoising}, we rescale
images to 64$\times$64 resolution for the first two datasets and
256$\times$256 for the latter two. We construct bridges in the pixel
space for the first three datasets and in the latent space of dimensions
32$\times$32$\times$4 for the Night$\leftrightarrow$Day dataset.
To map images to latent representations, we use a pretrained VQ-GAN
encoder \cite{rombach2022high}. Following prior work \cite{LiX0L23},
we use FID \cite{heusel2017gans}, IS \cite{salimans2016improved},
and LPIPS \cite{zhang2018perceptual} to measure the fidelity and
perceptual faithfulness of generated images. These metrics are computed
on training samples, as in \cite{zhou2024denoising}.

\begin{figure}
\begin{centering}
{\resizebox{0.70\textwidth}{!}{%
\par\end{centering}
\begin{centering}
\begin{tabular}{ccc}
\includegraphics[width=0.24\textwidth]{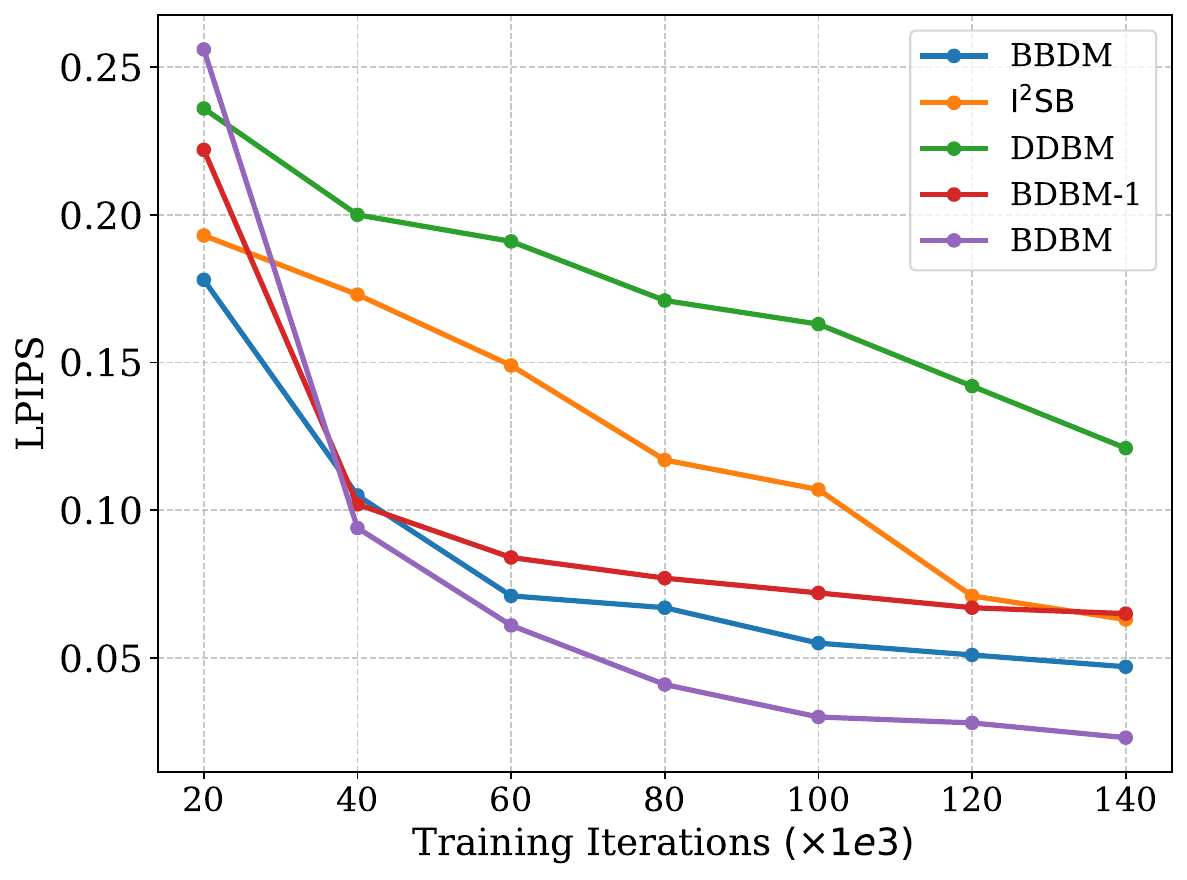} &  & \includegraphics[width=0.24\textwidth]{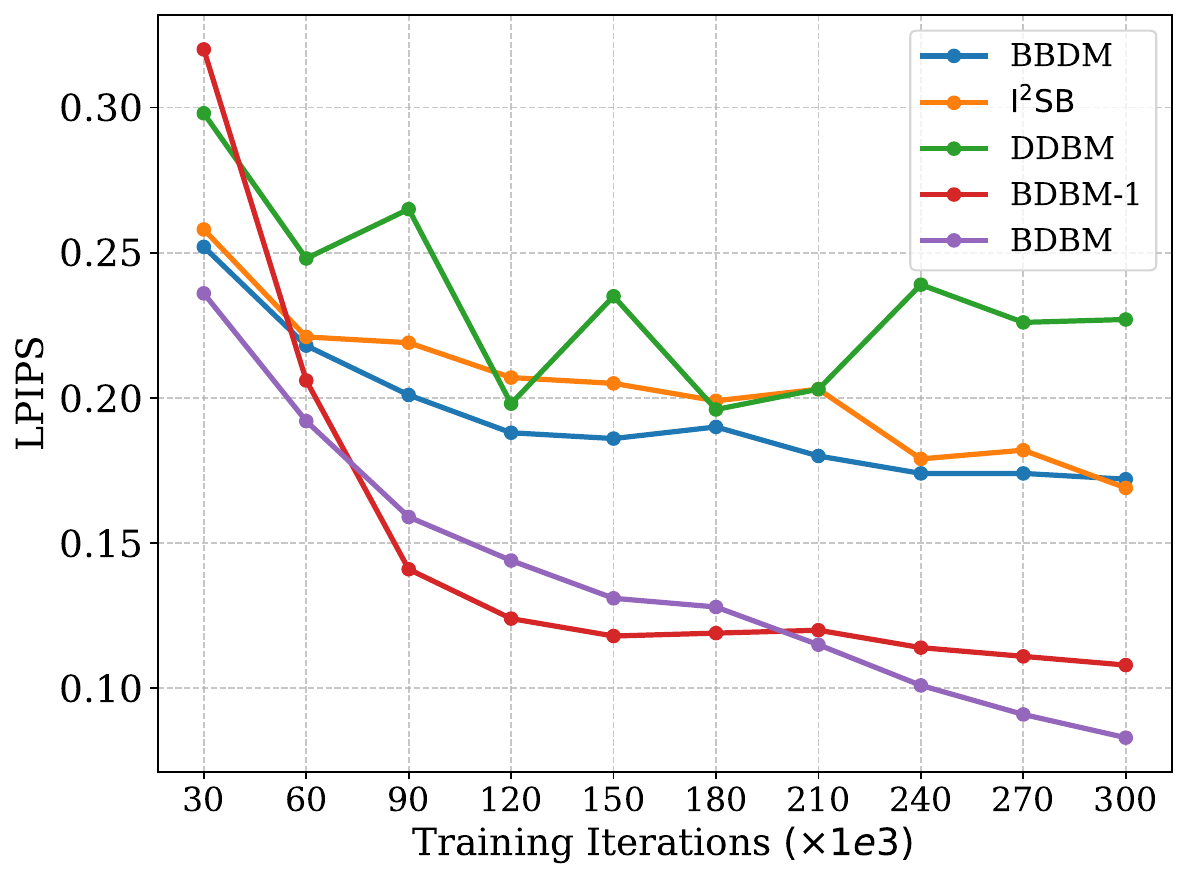}\tabularnewline
(a) Edges$\rightarrow$Shoes &  & (b) Edges$\rightarrow$Handbags\tabularnewline
\end{tabular}}}
\par\end{centering}
\caption{LPIPS curves of BDBM and unidirectional baselines on Edges$\rightarrow$Shoes
and Edges$\rightarrow$Handbags.\label{fig:LPIPS-curves-main}}
\end{figure}

\subsubsection{Model and training configurations\label{subsec:Model-and-training}}

Unless stated otherwise, we use Brownian bridges, as described in
Appdx.~\ref{subsec:Brownian-Bridge_settings}, with $\alpha_{t}=1-\frac{t}{T}$,
$\beta_{t}=\frac{t}{T}$ and $\sigma_{t}^{2}=k\frac{t}{T}\left(1-\frac{t}{T}\right)$
for our experiments. We consider discrete-time models with $T=1000$,
$\Delta t=1$, and $k=2$. Comparison with the continuous-time counterpart
is provided in Appdx.~\ref{subsec:Continuous-vs-Discrete}. For generation,
we employ ancestral sampling with number of function evaluations (NFE)
being 200. The variance of the transition kernel $\delta_{s,t}^{2}$
is set to $\delta_{s,t}^{2}=\eta\left(\sigma_{s}^{2}-\sigma_{t}^{2}\frac{\alpha_{s}^{2}}{\alpha_{t}^{2}}\right)$
with $\eta=1$. Studies on different values of $k$ and $\eta$ are
presented in Sections~\ref{subsec:noise_variance_ablation}, \ref{subsec:transition_variance_ablation},
respectively. We model $z_{\varphi}\left(t,x_{t},\left(1-m\right)*x_{0},m*x_{T}\right)$
using UNets with ADM architectures \cite{dhariwal2021diffusion} customized
for different input sizes. For 64$\times$64 images, we use 2 residual
blocks with 128 base channels. This allows us to train with batch
size of 128 for 64$\times$64 images on an H100 80GB GPU. For 256$\times$256
images, we increase the base channels to 256 and train with batch
size of 8. For effective training with batch size of 32, we accumulate
gradients over 4 update steps. All models were trained for 140k iterations
on the Edges$\leftrightarrow$Shoes dataset and 300k iterations on
the other datasets. The reduced iterations for Edges$\leftrightarrow$Shoes
were due to its smaller training set of 50k samples, compared to 130k
for Edges$\leftrightarrow$Handbags, as well as its smaller image
sizes compared to DIODE Outdoor and Night$\leftrightarrow$Day. The
Adam optimizer \cite{KingmaB14} is employed with a learning rate
of 1e-4 and $\beta_{1}$ set to 0.9.

\subsubsection{Baselines}

We compare our method BDBM with both unidirectional and bidirectional
I2I translation baselines. The unidirectional baselines include state-of-the-art
(SOTA) diffusion bridge models such as $\text{I}^{2}\text{SB}$ \cite{LiuVHTNA23},
BBDM \cite{LiX0L23} and DDBM \cite{zhou2024denoising}. We also include
a unidrectional variant of our method, referred to as BDBM-1, for
comparison to highlight the impact of modeling both directions simultaneously.
The bidirectional baselines consist of DDIB \cite{su2023dual} and
Rectified Flow (RF) \cite{liu2022flow}. The baselines, excluding
RF, were trained using their official code repositories. Since the
official RF code does not support parallel training, we used the implementation
from \cite{lee2023minimizing} for parallel training. For all baselines,
we use the same architecture, training configurations, and NFE as
our method.

\begin{table*}
\begin{centering}
\begin{tabular}{ccccccc}
\toprule 
\multirow{2}{*}{Model} & \multicolumn{3}{c}{Edges$\leftrightarrow$Shoes$\times64$} & \multicolumn{3}{c}{Edges$\leftrightarrow$Handbags$\times64$}\tabularnewline
\cmidrule{2-7} 
 & FID $\downarrow$ & IS $\uparrow$ & LPIPS $\downarrow$ & FID $\downarrow$ & IS $\uparrow$ & LPIPS $\downarrow$\tabularnewline
\midrule
\midrule 
DDIB & 85.24/45.19 & 2.13/\textbf{3.40} & 0.38/0.45 & 77.95/31.50 & 2.81/3.59 & 0.49/0.52\tabularnewline
\midrule 
RF & 8.63/43.17 & \textbf{2.21}/2.79 & 0.03/0.16 & 5.98/48.53 & \textbf{3.19}/3.71 & 0.07/0.25\tabularnewline
\midrule
\midrule 
BDBM (ours) & \textbf{0.98}/\textbf{1.06} & 2.20/3.28 & \textbf{0.01}/\textbf{0.02} & \textbf{1.87}/\textbf{3.06} & 3.10/\textbf{3.74} & \textbf{0.02}/\textbf{0.08}\tabularnewline
\bottomrule
\end{tabular}
\par\end{centering}
\caption{Results of BDBM and bidirectional baselines on bidirectional translation
tasks. For each method and metric, we report two numbers, the left
is for color-to-sketch translation, and the right is for sketch-to-color
translation. The best results are highlighted in bold.\label{tab:main_result_2}}
\end{table*}

\subsection{Experimental Results}

\subsubsection{Unidirectional I2I translation\label{subsec:Unidirectional-I2I-Translation}}

Following \cite{zhou2024denoising}, we experiment with the Edges$\leftrightarrow$Shoes,
Edges$\leftrightarrow$Handbags, and DIODE Outdoor datasets, focusing
on translating sketches or normal maps to color images, as this translation
is more challenging than the reverse. Results for the reverse translation
are provided in Appdx.~\ref{subsec:Unidirectional-reverse-translation}.

As shown in Table~\ref{tab:main_result_1} and Fig.~\ref{fig:LPIPS-curves-main},
BDBM significantly outperforms BDBM-1 and other unidirectional baselines
in most metrics and datasets. This improvement is also evident in
the superior quality of samples generated by our method compared to
the baselines, as displayed in Fig.~\ref{fig:main_qualitative_results}.
Notably, BDBM was trained using the same number of iterations as the
baselines. This means that the actual number of model updates w.r.t.
a specific direction in BDBM is only \emph{half} that of the baselines,
as the two endpoints $x_{0}$, $x_{T}$ are sampled with equal probability
in the loss $\Loss_{\text{BDBM}}$ (Eq.~\ref{eq:loss_BDBM}). This
demonstrates the clear advantage of our proposed bidirectional training
over the unidirectional counterpart.

We hypothesize that allowing either $x_{0}$ or $x_{T}$ to serve
as the condition for the \emph{shared-parameter} noise model $z_{\varphi}$
during training enables the optimizer to leverage the endpoint that
yields more accurate predictions for effective parameter updates.
Intuitively, this endpoint is likely the one closer in time to the
input $x_{t}$ of the noise model. For instance, consider two noise
predictions $z_{\varphi}\left(t,x_{t},x_{0}\right)$ and $z_{\varphi}\left(t,x_{t},x_{T}\right)$
for $x_{t}$ at time $t$ closer to 0 than to $T$, where $x_{0}$
and $x_{T}$ are chosen with equal probability. Since $x_{0}$ generally
provides more reliable information about the noise in $x_{t}$ compared
to $x_{T}$, the optimizer tends to prioritize the output of $z_{\varphi}\left(t,x_{t},x_{0}\right)$
when updating the shared parameters $\varphi$. This update not only
improves the accuracy of $z_{\varphi}\left(t,x_{t},x_{0}\right)$
but also enhances $z_{\varphi}\left(t,x_{t},x_{T}\right)$ due to
the shared parameter structure. In contrast, unidirectional training
can only use a single endpoint, for example $x_{T}$, as the condition,
which reduces it effectiveness in learning model parameters at times
$t$ far from $T$. As $x_{t}$ becomes increasingly different from
$x_{T}$, the information provided by $x_{T}$ becomes less useful
for accurately predicting the noise in $x_{t}$.

\subsubsection{Bidirectional I2I translation\label{subsec:Bidirectional-I2I-Translation}}

\begin{figure}
\begin{centering}
{\resizebox{0.70\textwidth}{!}{%
\par\end{centering}
\begin{centering}
\begin{tabular}{c|c|c|c}
\textcolor{blue}{Reference} & DDIB & RF & BDBM (ours)\tabularnewline
\includegraphics[width=0.12\textwidth]{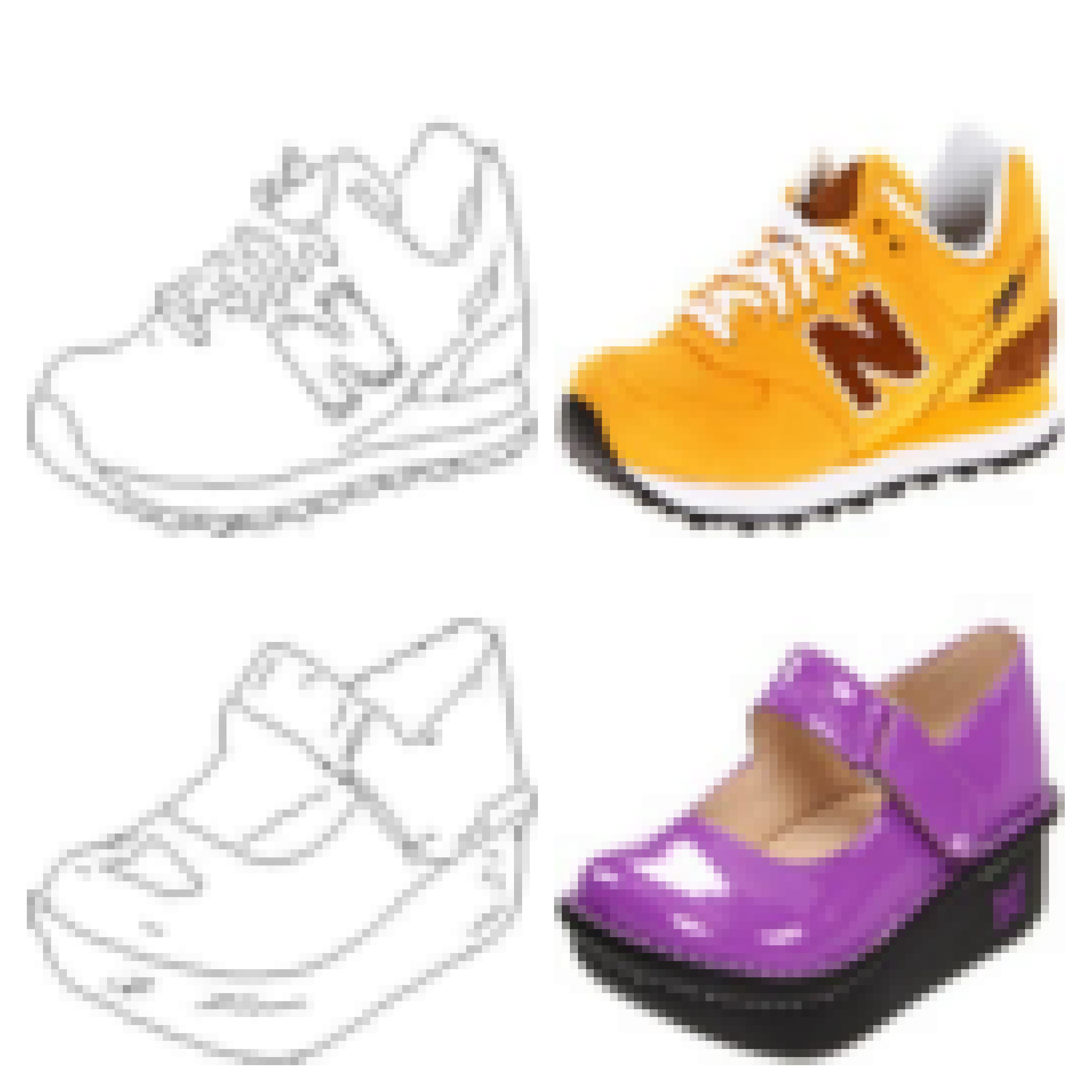} & \includegraphics[width=0.12\textwidth]{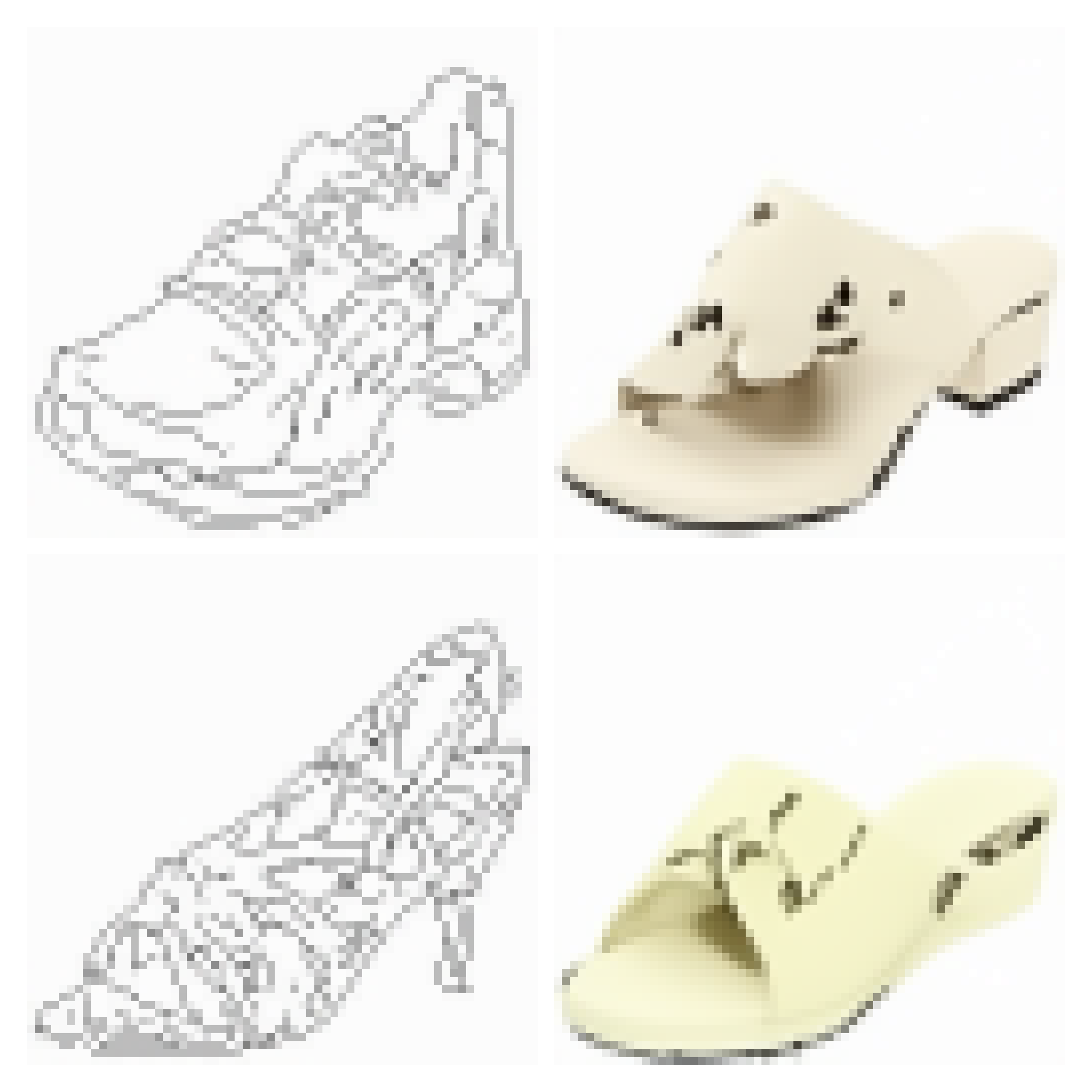} & \includegraphics[width=0.12\textwidth]{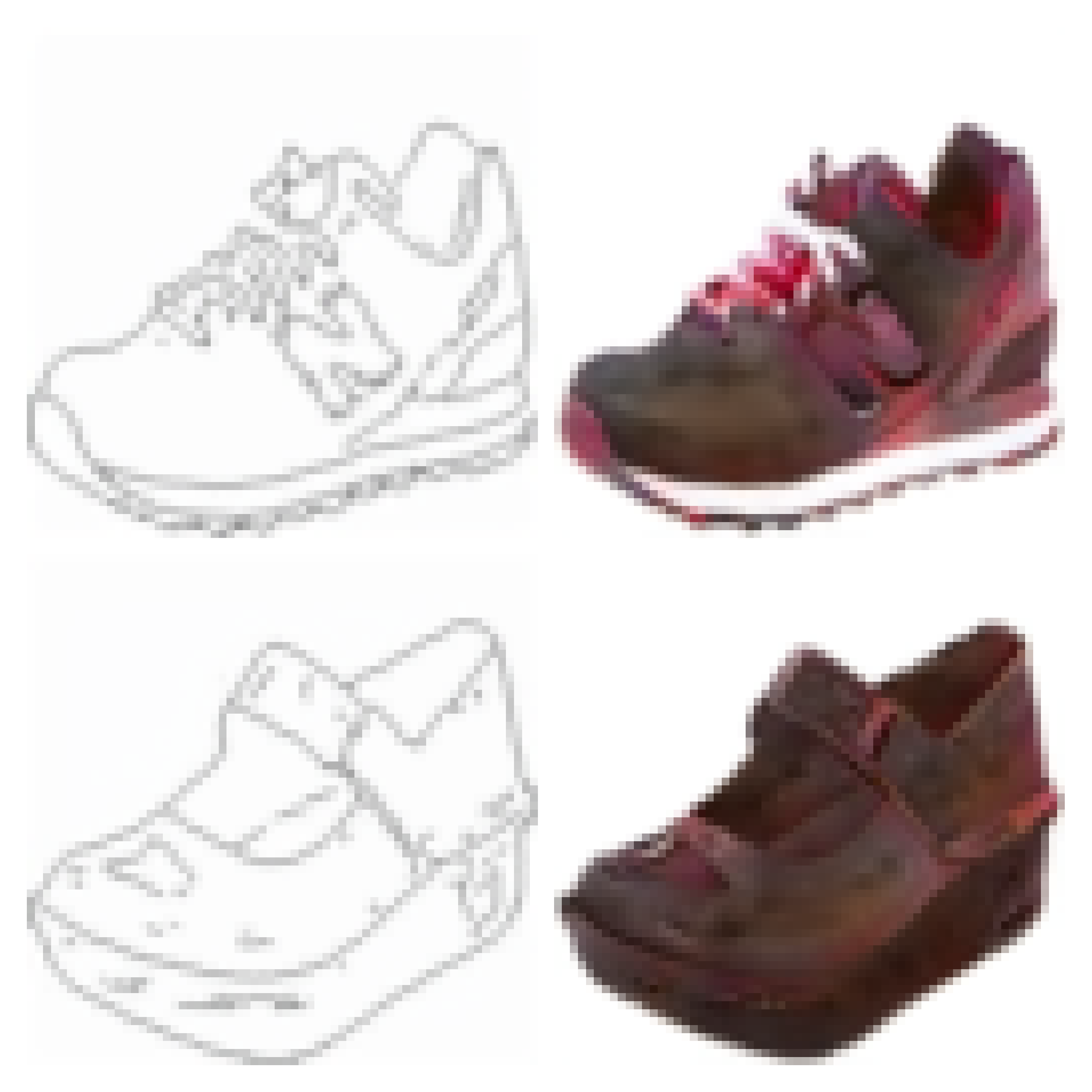} & \includegraphics[width=0.12\textwidth]{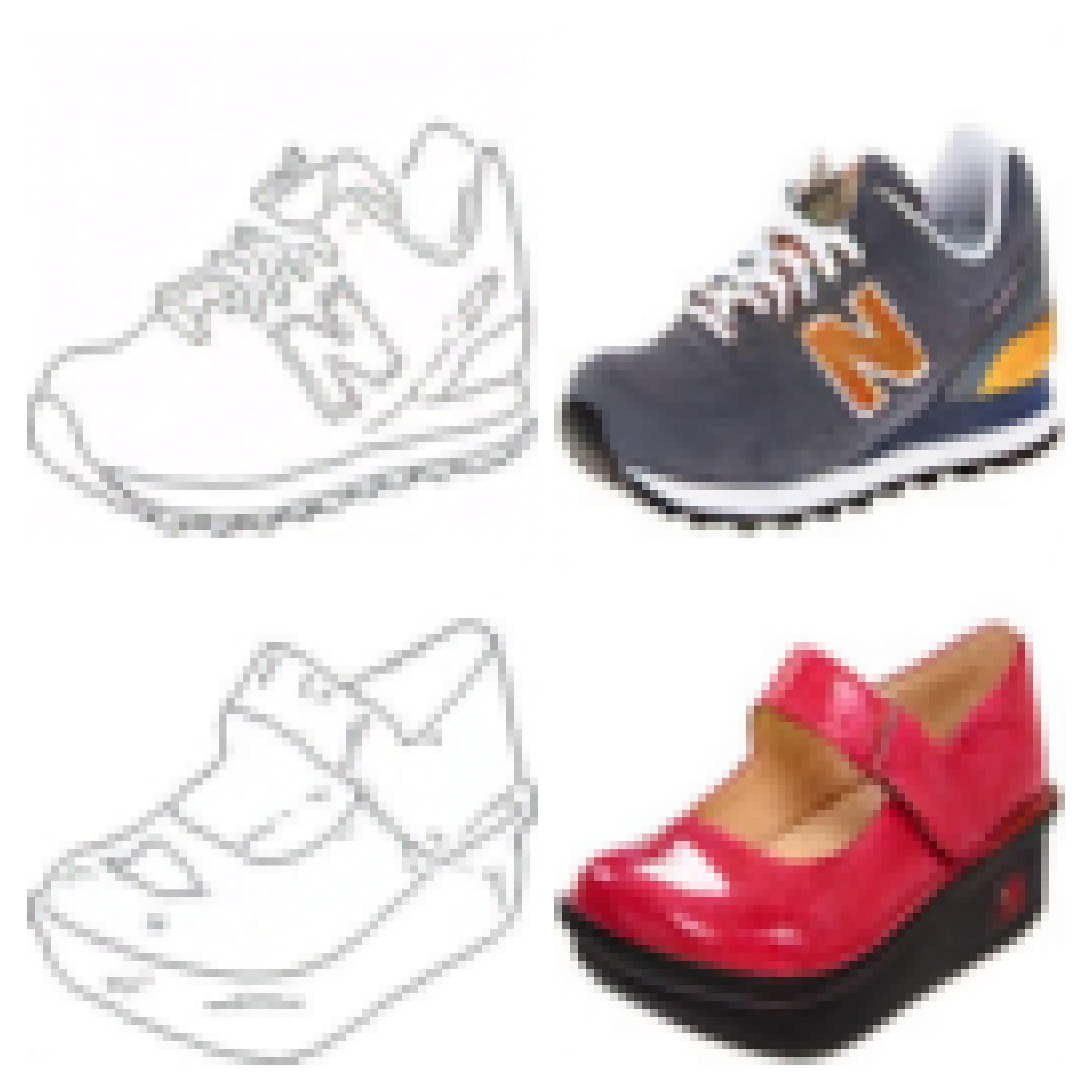}\tabularnewline
\includegraphics[width=0.12\textwidth]{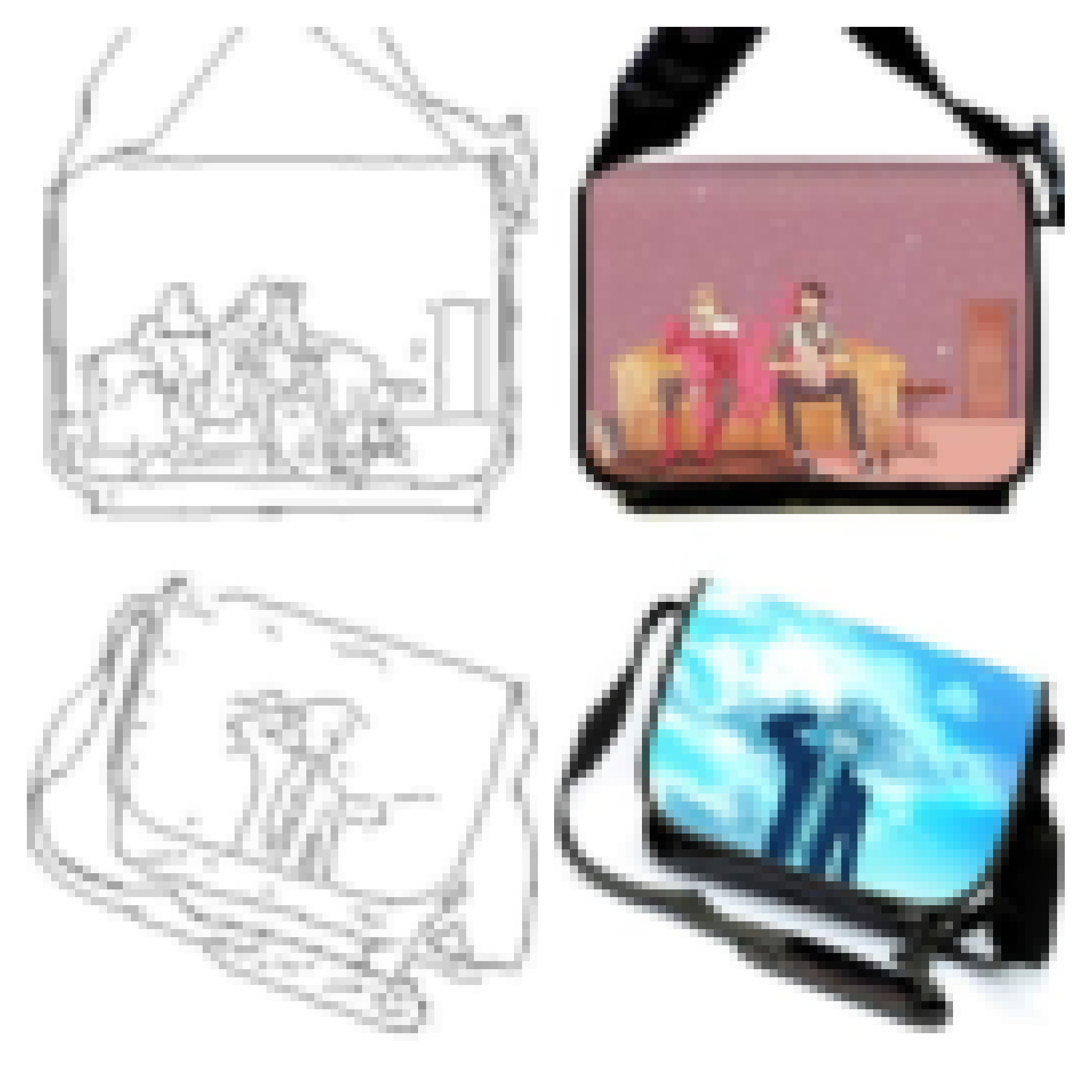} & \includegraphics[width=0.12\textwidth]{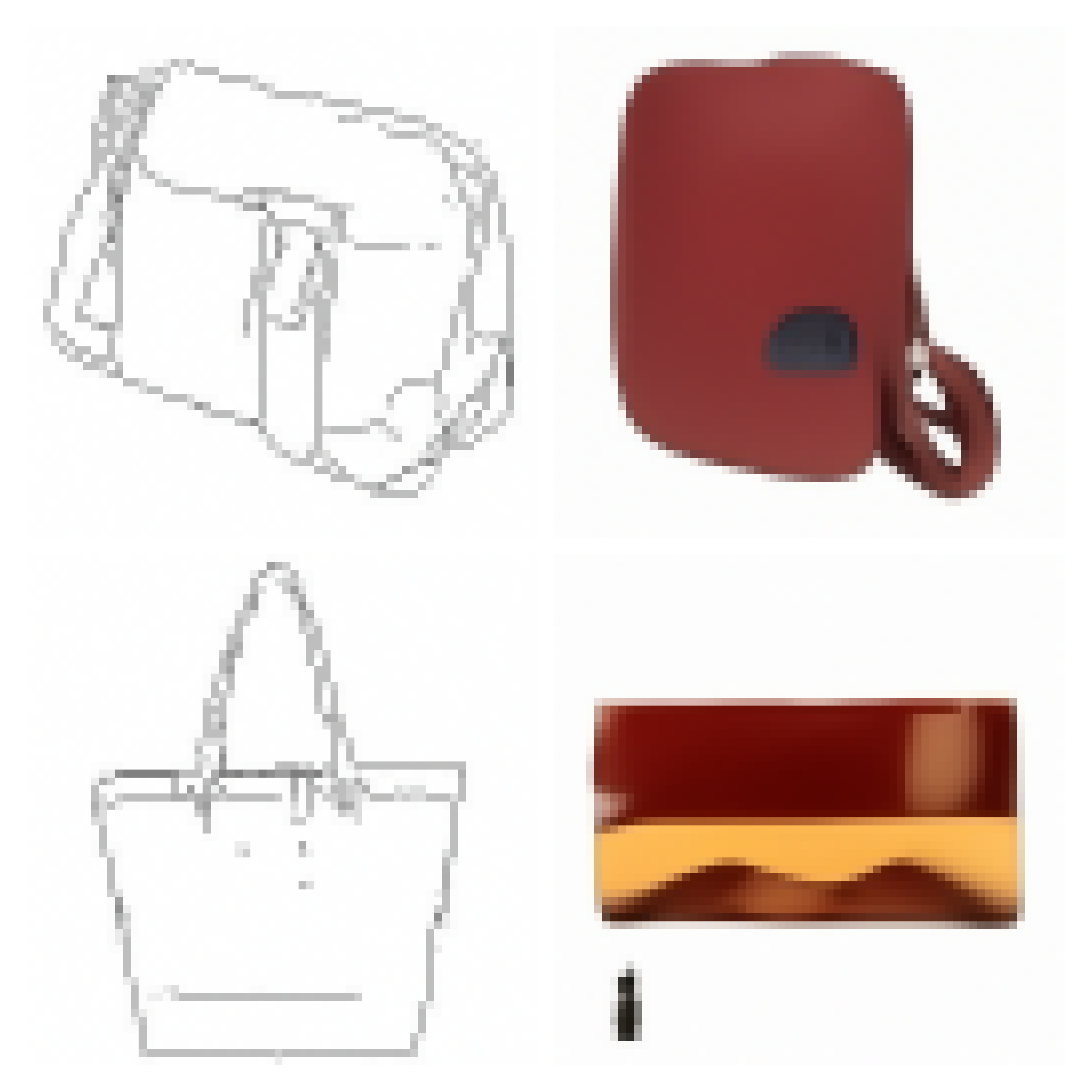} & \includegraphics[width=0.12\textwidth]{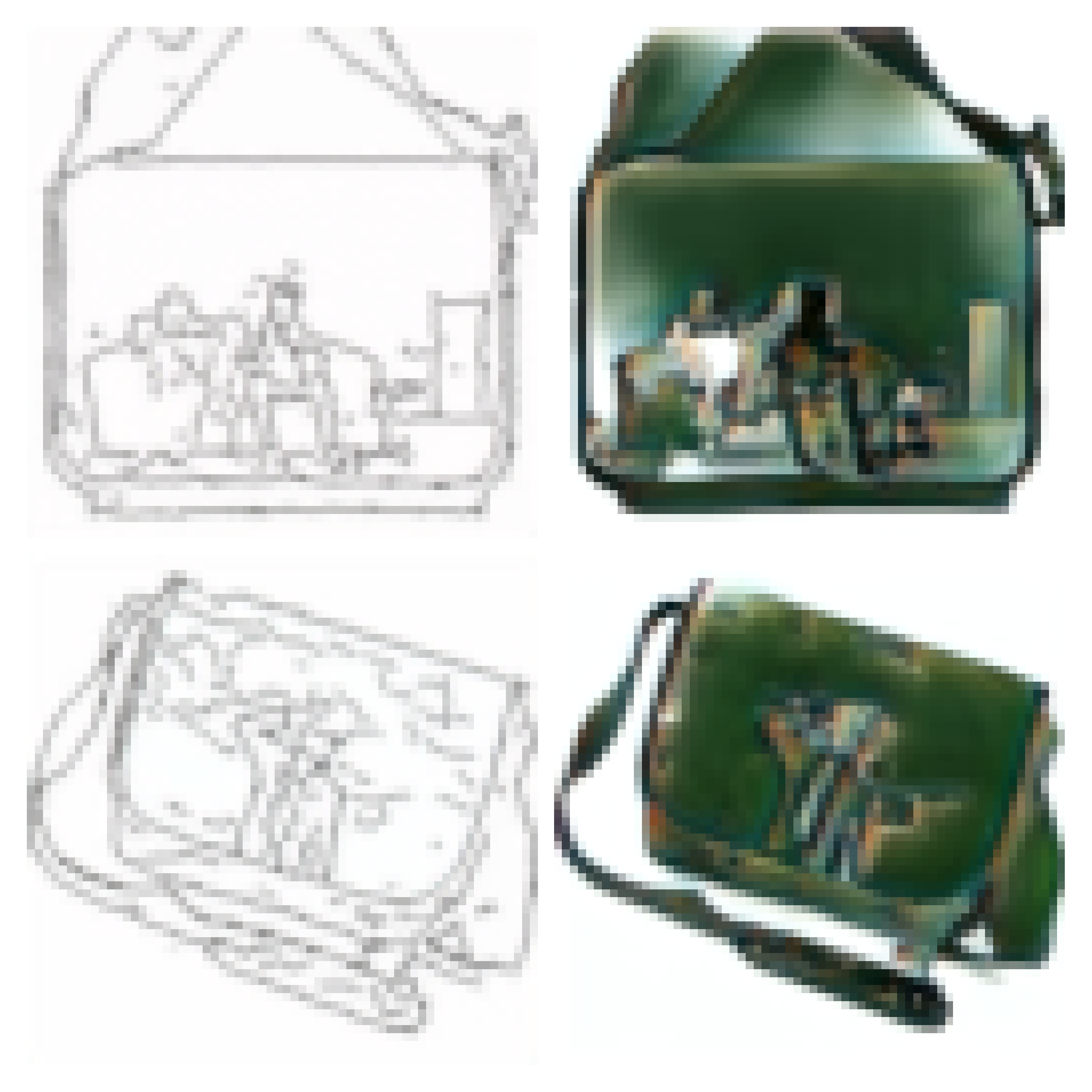} & \includegraphics[width=0.12\textwidth]{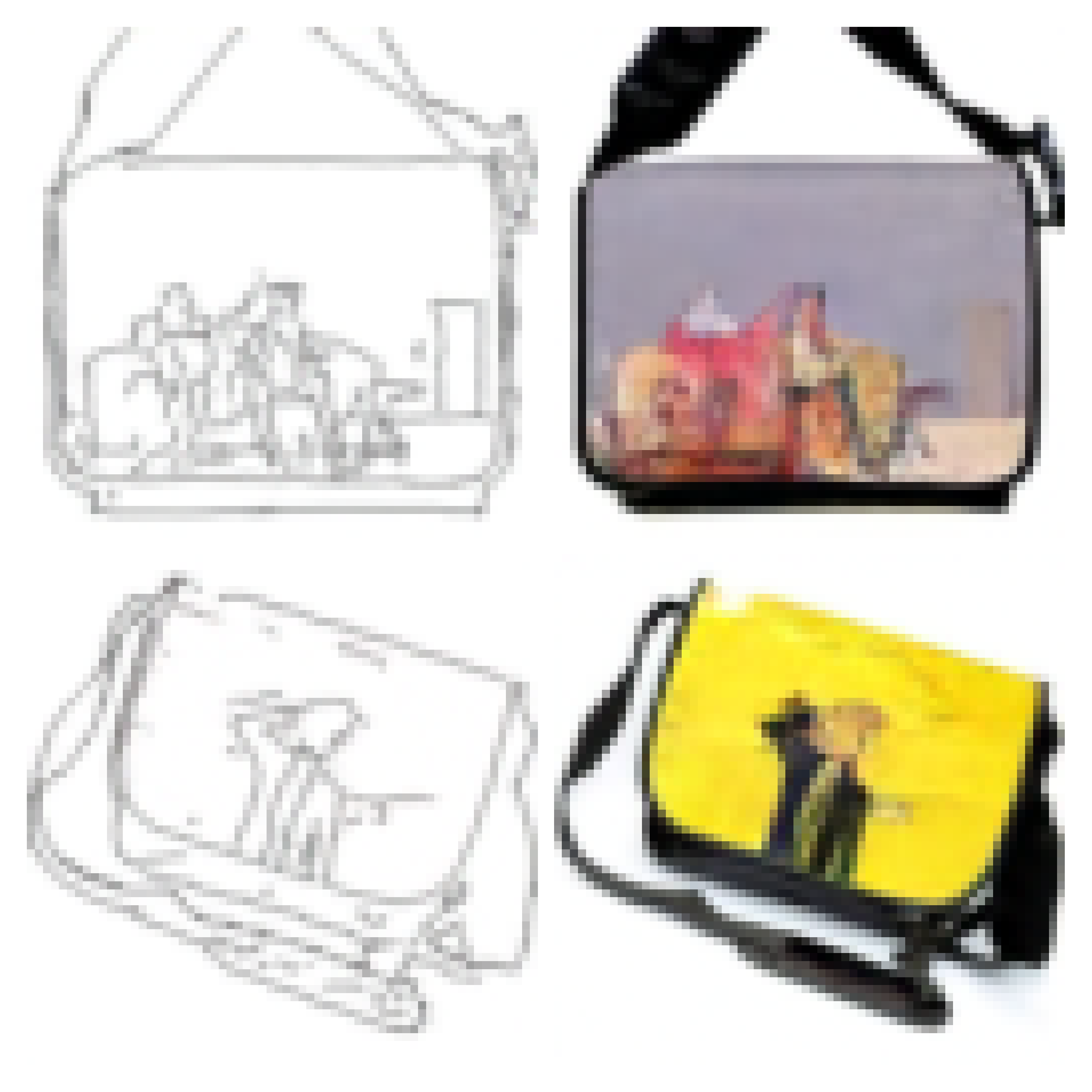}\tabularnewline
\end{tabular}}}
\par\end{centering}
\caption{Images generated by BDBM and bidirectional baselines on Edges$\leftrightarrow$Shoes
and Edges$\leftrightarrow$Handbags. ``Reference'' column shows
reference images of the two domains. \label{fig:bidirect_qualitative}}
\end{figure}

We compare BDBM with bidirectional baselines DDIB and RF, presenting
quantitative and qualitative results in Table~\ref{tab:main_result_2}
and Fig.~\ref{fig:bidirect_qualitative}. BDBM outperforms the two
baselines by large margins for translations in both directions. DDIB
struggles to maintain pair consistency between boundary samples due
to random mapping into shared Gaussian latent samples, resulting in
translations that often differ greatly from the ground truth. Meanwhile,
RF performs reasonably well for the color-to-sketch translation but
poorly for the reverse. This is because different color images can
have very similar sketch images. This causes the learned velocity
for the sketch-to-color translation to point toward the average of
multiple target color images associated with a source sketch image,
as evident in Fig.~\ref{fig:bidirect_qualitative}.

\noindent 

\subsection{Ablation Study}

\subsubsection{Impacts of different parameterizations}

\noindent 
\begin{table*}
\begin{centering}
\begin{tabular}{ccccccccc}
\toprule 
\multirow{2}{*}{Prediction} & \multicolumn{4}{c}{Edges$\rightarrow$Shoes$\times64$} & \multicolumn{4}{c}{Edges$\rightarrow$Handbags$\times64$}\tabularnewline
\cmidrule{2-9} 
 & FID $\downarrow$ & IS $\uparrow$ & LPIPS $\downarrow$ & Diversity $\uparrow$ & FID $\downarrow$ & IS $\uparrow$ & LPIPS $\downarrow$ & Diversity $\uparrow$\tabularnewline
\midrule
\midrule 
$z$ & 1.06 & 3.28 & 0.02 & 6.90 & 3.06 & 3.74 & 0.08 & 9.01\tabularnewline
\midrule 
$x_{T}+x_{0}$ & 1.51 & 3.25 & 0.04 & 2.21 & 3.71 & 3.75 & 0.11 & 7.54\tabularnewline
\midrule 
$\left(x_{T},x_{0}\right)$ & 1.49 & 3.24 & 0.01 & 1.97 & 3.49 & 3.77 & 0.12 & 7.88\tabularnewline
\bottomrule
\end{tabular}
\par\end{centering}
\caption{Results of our method w.r.t. different parameterizations.\label{tab:Results across different param}}
\end{table*}

As discussed in Section~\ref{subsec:Bidirectional-Diffusion-Bridge-Models},
the transition kernel of BDBM can be modeled by predicting the noise
$z$ or endpoints (either by predicting $x_{0}+x_{T}$ and inferring
the missing endpoint given the known one, or by directly predicting
one endpoint given the other). We compare the effectiveness of these
approaches on the Edges$\rightarrow$Shoes and Edges$\rightarrow$Handbags
translation tasks, with results shown in Table~\ref{tab:Results across different param}.
In addition to FID and LPIPS metrics, we evaluate Diversity \cite{batzolis2021conditional,LiX0L23},
which measures the average pixel-wise standard deviation of multiple
color images generated from a single sketch on a held-out test set
of 200 samples. We observe that predicting noise achieves slightly
better FID scores and produces more diverse samples than predicting
endpoints. We hypothesize that since $x_{0}$, $x_{T}$ are fixed
while $z$ is sampled randomly during training, predicting endpoints
tends to have less variance than predicting noise, which results in
less diverse samples.

\subsubsection{Effect of the noise variance $\sigma_{t}^{2}$\label{subsec:noise_variance_ablation}}

In Section~\ref{subsec:Model-and-training}, the noise variance $\sigma_{t}^{2}$
of BDBM is defined as $\sigma_{t}^{2}=k\frac{t}{T}\left(1-\frac{t}{T}\right)$,
which means we can control $\sigma_{t}^{2}$ by changing the value
of $k$. Table~\ref{tab:ablation_k} shows the results on Edges$\rightarrow$Shoes
for different values of $k\in\left\{ 1,2,4,8\right\} $. Increasing
$k$ generally yields more diverse samples but worsens FID and LPIPS
scores. This trade-off occurs because higher $k$ values increase
the variance of the distribution $q\left(x_{t}|x_{0},x_{T}\right)$,
enlarging the path space and consequently making the model optimization
more challenging. Conversely, when $k$ is too small, the noise variance
becomes insufficient to corrupt domain information for effective translation.
Our results indicate that $k=2$ offers the best balance between diversity
and quality.

\begin{table}
\begin{centering}
\begin{tabular}{cccc}
\toprule 
\multirow{2}{*}{$k$} & \multicolumn{3}{c}{Edges$\rightarrow$Shoes$\times64$}\tabularnewline
\cmidrule{2-4} 
 & FID $\downarrow$ & LPIPS $\downarrow$ & Diversity $\uparrow$\tabularnewline
\midrule
\midrule 
1 & 2.07 & 0.04 & 4.61\tabularnewline
\midrule 
2 & 1.06 & 0.02 & 6.90\tabularnewline
\midrule 
4 & 2.35 & 0.03 & 7.26\tabularnewline
\midrule 
8 & 3.52 & 0.05 & 7.81\tabularnewline
\bottomrule
\end{tabular}
\par\end{centering}
\caption{Results of BDBM on Edges$\rightarrow$Shoes w.r.t. different values
of $k$ controlling the variance $\sigma_{t}^{2}$.\label{tab:ablation_k}}
\end{table}

\subsubsection{Effect of the variance $\delta_{s,t}^{2}$ of the transition kernel\label{subsec:transition_variance_ablation}}

\begin{table}
\begin{centering}
\begin{tabular}{ccccccc}
\toprule 
 &  & \multicolumn{5}{c}{Edges$\rightarrow$Shoes$\times64$}\tabularnewline
\midrule 
\multicolumn{2}{c}{NFE} & 20 & 50 & 100 & 200 & 1000\tabularnewline
\midrule
\midrule 
\multirow{4}{*}{$\eta$} & 0.0 & 4.16 & 2.98 & 2.47 & 2.15 & 1.87\tabularnewline
\cmidrule{2-7} 
 & 0.2 & 3.37 & 2.31 & 1.79 & 1.42 & 1.14\tabularnewline
\cmidrule{2-7} 
 & 0.5 & 2.63 & 1.69 & 1.38 & 1.10 & 0.96\tabularnewline
\cmidrule{2-7} 
 & 1.0 & 2.11 & 1.52 & 1.25 & 1.06 & 0.92\tabularnewline
\bottomrule
\end{tabular}
\par\end{centering}
\caption{FID scores of BDBM on Edges$\rightarrow$Shoes w.r.t. different values
of $\eta$ controlling the variance $\delta_{s,t}^{2}$ and different
numbers of sampling steps.\label{tab:ablation_eta}}
\end{table}

\noindent 
\begin{figure}
\noindent \begin{centering}
\includegraphics[width=0.3\textwidth]{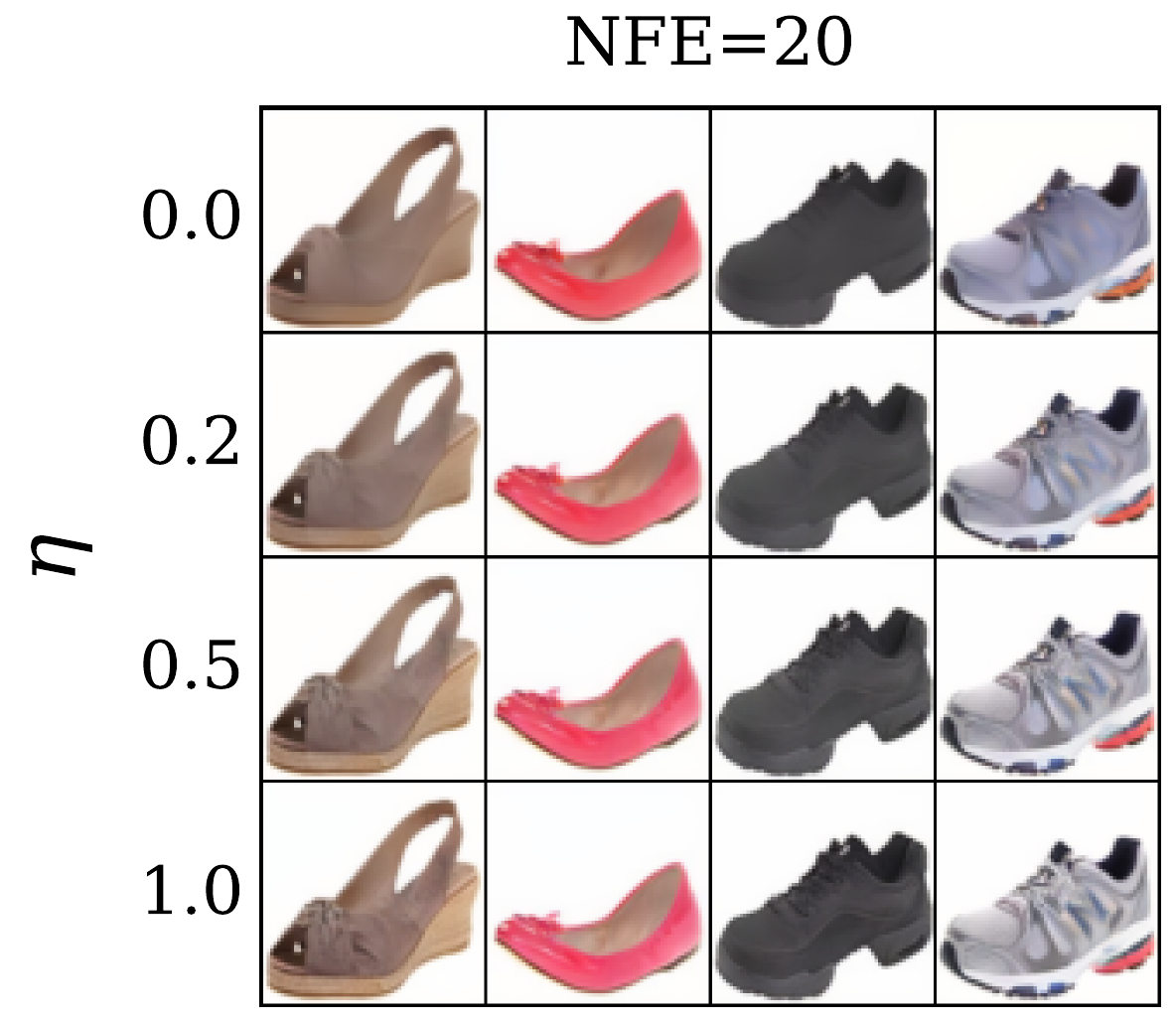}\hspace*{0.04\columnwidth}\includegraphics[width=0.3\textwidth]{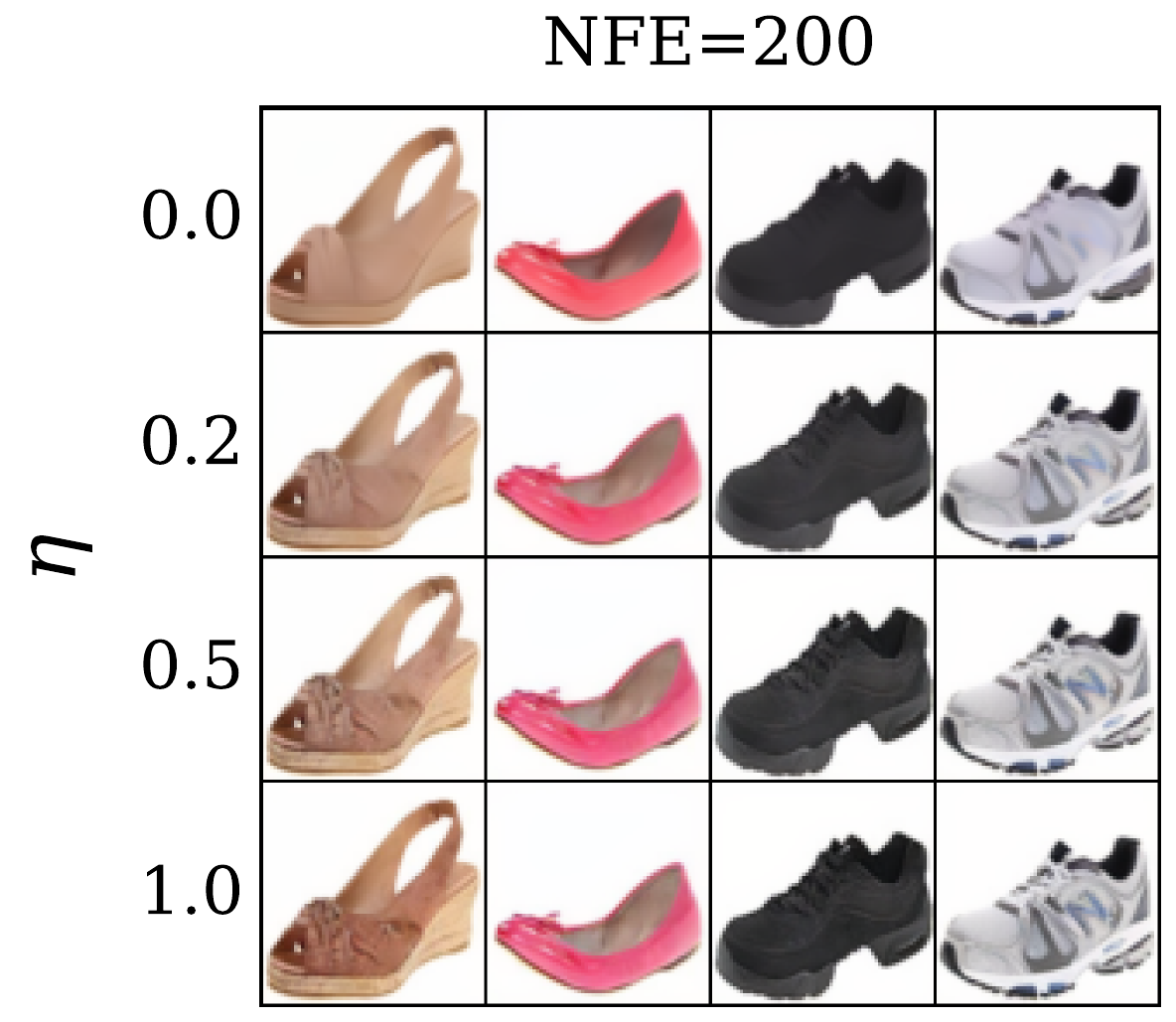}
\par\end{centering}
\caption{Samples generated by BDBM when translating from sketches to shoes
using NFE=20 and NFE=200 for w.r.t. different values of $\eta$.\label{fig:ablation_eta_quantitative}}
\end{figure}

We study the impact of varying the variance $\delta_{s,t}^{2}$ of
the transition kernel via changing $\eta$ (Section~\ref{subsec:Model-and-training})
on generation quality, with the results presented in Table~\ref{tab:ablation_eta}
and Fig.~\ref{fig:ablation_eta_quantitative}. We observe that increasing
$\eta$ from 0 to 1 consistently improves the quality of generated
results, regardless of the number of sampling steps. The reason is
that the target bridge process connecting boundary points from the
two domains is stochastic and corresponds to $\eta=1$. Consequently,
higher $\eta$ values make $x_{t}$ more likely to be a sample from
the target distribution at time $t$, leading to better results.

\subsubsection{Translation in latent spaces}

\begin{table}
\centering{}%
\begin{tabular}{cccc}
\toprule 
\multirow{2}{*}{Model} & \multicolumn{3}{c}{Day$\rightarrow$Night$\times256$}\tabularnewline
\cmidrule{2-4} 
 & FID $\downarrow$ & IS $\uparrow$ & LPIPS $\downarrow$\tabularnewline
\midrule
\midrule 
RF & 12.38 & 3.90 & 0.37\tabularnewline
\midrule 
DDIB & 226.9 & 2.11 & 0.79\tabularnewline
\midrule 
$\text{I}^{2}\text{SB}$ & 15.56 & 4.03 & 0.36\tabularnewline
\midrule 
DDBM & 27.63 & 3.92 & 0.55\tabularnewline
\midrule
\midrule 
BDBM (ours) & \textbf{6.63} & \textbf{4.18} & \textbf{0.34}\tabularnewline
\bottomrule
\end{tabular}\caption{Comparison of BDBM and baseline methods on the Day$\rightarrow$Night
translation task in latent spaces. Baseline results are sourced from
\cite{zhou2024denoising}\label{tab:ablation_latent}}
\end{table}

To validate BDBM's translation capability in latent spaces, we adopt
the Day$\rightarrow$Night translation experiment from \cite{zhou2024denoising}.
For a fair comparison, we maintain the same experimental settings
as in \cite{zhou2024denoising}, including the model architecture,
training iterations, and NFE=53 for sample generation. We also follow
\cite{zhou2024denoising} and compute metrics using the reconstructed
versions of the ground-truth target images. This helps mitigate the
impact of the VQ-GAN decoding process and ensures that the results
accurately reflect the translation quality. Table~\ref{tab:ablation_latent}
presents the results of BDBM and baseline methods, with the baseline
results taken from \cite{zhou2024denoising}. It is evident that BDBM
significantly outperforms the baselines, demonstrating its consistent
performance in both pixel and latent spaces. We also observed that
BDBM effectively captures the statistics of the two domains, where
in the dataset, nighttime images are much less diverse than daytime
ones, leading to the generation of duplicated nighttime images when
using different random seeds, as illustrated in Fig.~\ref{fig:Additional-results-NightDay}.

\section{Related Work}

\subsection{Schrödinger Bridges and Diffusion Bridges}

Recent bridge models can broadly be classified into Schrödinger bridges
(SB) and diffusion bridges (DB). The Schrödinger Bridge problem \cite{schrodinger1931umkehrung,pavon1991free}
aims to find a stochastic process that connects two arbitrary marginal
distributions $p_{A}$, $p_{B}$ while remaining as close as possible
to a reference process. When the reference process is a diffusion
process initialized at $p_{A}$, the solution to the SB problem can
be characterized by two coupled partial differential equations (PDEs)
governing the forward and backward diffusion processes initialized
at $p_{A}$ and $p_{B}$, respectively \cite{leonard2014survey,vargas2021solving,bortoli2021diffusion,chen2022likelihood,LiuCST22}. 

SB models are typically trained using iterative proportional fitting
which requires expensive simulation of the forward and backward processes
\cite{fortet1940resolution,bortoli2021diffusion}. Several approaches
have been proposed \cite{peluchetti2023diffusion,shi2023diffusion,tong2023simulation}
to improve the scalability of training SB models by leveraging the
score and flow matching frameworks \cite{hyvarinen2005estimation,song2020score,lipman2022flow,liu2022flow}.
However, SB models overlook the relationships between samples from
the two domains, making them unsuitable for paired translation tasks. 

Diffusion bridges simplify Schrödinger bridges by assuming a Dirac
distribution at one endpoint, allowing them to model the coupling
between the two domains for paired translations. I$^{2}$SB \cite{LiuVHTNA23}
is a diffusion bridge derived from the general theory of SBs. On the
other hand, methods like SB{\scriptsize{}ALIGN} \cite{SomnathPHM0B23},
$\Omega$-bridge \cite{liu2022let,LiuW0l23}, and DDBM \cite{zhou2024denoising}
leverage Doob's $h$-transform to obtain the formula of a continuous-time
$h$-transformed process that converges almost surely to a specific
target sample while aligning closely with the reference diffusion
process. SB{\scriptsize{}ALIGN} and $\Omega$-bridge create a $h$-transform
process that generates data and learn the drift of this process, whereas
DDBM designs a $h$-transformed process that converges to a latent
sample. For data generation, DDBM learns the score with respect to
the reverse process via conditional score matching, following the
approach in \cite{song2020score}. BBDM \cite{LiX0L23} extends the
unconditional variational framework for discrete-time diffusion processes
\cite{ho2020denoising,kingma2021variational,song2020denoising} to
a conditional variational framework for Brownian bridges. It then
uses the new framework to model the transition kernel of the data
generation process.

Different from the aforementioned methods, our method is built on
the Chapman-Kolmogorov equation (CKE) for bridges and has a novel
design that supports bidirectional transition between the two domains
using a single model.

\subsection{Diffusion and Flow Models for I2I}

Diffusion models (DMs) \cite{sohl2015deep,song2019generative,ho2020denoising,song2020score}
are powerful generative models that progressively denoise latent samples
from a standard Gaussian distribution to generate images. For image-to-image
(I2I) translation, DMs can incorporate source images as conditions
through either classifier-based \cite{dhariwal2021diffusion} or classifier-free
\cite{ho2021classifier} guidance techniques during the denoising
process to generate corresponding target images \cite{sasaki2021unit,saharia2022palette,zhao2022egsde,wolleb2022swiss}.
However, since one of the two boundary distributions in DMs is always
a standard Gaussian, bidirectional translation requires training two
distinct DMs conditioned on source and target images. DDIB \cite{su2023dual}
exemplifies this approach by combining two separate diffusion models
for source and target domains through a shared Gaussian latent space
for bidirectional translation.

Flow models (FMs) \cite{liu2022flow,lipman2022flow,albergo2023stochastic,do2024variational}
build an ODE map between two arbitrary boundary distributions and
can be trained via the flow matching loss \cite{lipman2022flow} related
to the score matching loss for diffusion models \cite{song2020score}.
FMs can be viewed as special cases of diffusion bridges where the
variance of the transition kernel is zero. Due to their deterministic
nature, FMs are less suitable for capturing the coupling between two
domains, as demonstrated by our experimental results in Sections~\ref{subsec:Bidirectional-I2I-Translation}
and \ref{subsec:noise_variance_ablation}. Nonetheless, FMs can be
useful for unpaired translation and can be specially designed to represent
optimal transport maps \cite{liu2022rectified,lipman2022flow,tong2023improving}.

\section{Conclusion}

We introduced the Bidirectional Diffusion Bridge Model (BDBM), a novel
framework for bidirectional image-to-image (I2I) translation using
a single network. By leveraging the Chapman-Kolmogorov Equation, BDBM
models the shared components of forward and backward transitions,
enabling efficient bidirectional generation with minimal computational
overhead. Empirical results demonstrated that BDBM consistently outperforms
existing I2I translation methods across diverse datasets.

Despite these strengths, BDBM has so far been applied exclusively
to the image domain. Extending it to other domains, such as text,
presents an exciting direction for future research. In particular,
exploring BDBM for multimodal tasks like image$\leftrightarrow$text
generation would be a promising avenue.

\clearpage{}

\bibliographystyle{plain}

\addtocontents{toc}{\protect\setcounter{tocdepth}{2}}

\clearpage{}

\appendix
\tableofcontents{}

\section{Theoretical Results}

\subsection{Derivation of the backward CKE in Eq.~\ref{eq:backward_CKE_2} from
Eq.~\ref{eq:backward_CKE}\label{subsec:Derivation-of-the-backward-CKE}}

\noindent According to Bayes' rule, we have:
\begin{align}
p\left(x_{t}|x_{T}\right) & =\frac{p\left(x_{T}|x_{t}\right)p\left(x_{t}\right)}{p\left(x_{T}\right)}\label{eq:backward_CKE_deriv_1}\\
 & =\frac{p\left(x_{t}\right)}{p\left(x_{T}\right)}\int p\left(x_{T}|x_{t+1}\right)p\left(x_{t+1}|x_{t}\right)dx_{t+1}\label{eq:backward_CKE_deriv_2}\\
 & =\int\frac{p\left(x_{t}\right)}{p\left(x_{T}\right)}p\left(x_{T}|x_{t+1}\right)p\left(x_{t+1}|x_{t}\right)dx_{t+1}\\
 & =\int p\left(x_{t+1},x_{t}\right)\frac{p\left(x_{T}|x_{t+1}\right)}{p\left(x_{T}\right)}dx_{t+1}\\
 & =\int p\left(x_{t}|x_{t+1}\right)\frac{p\left(x_{T}|x_{t+1}\right)p\left(x_{t+1}\right)}{p\left(x_{T}\right)}dx_{t+1}\\
 & =\int p\left(x_{t}|x_{t+1}\right)p\left(x_{t+1}|x_{T}\right)dx_{t+1}\label{eq:backward_CKE_deriv_6}
\end{align}

Here, $p\left(x_{T}|x_{t}\right)=\int p\left(x_{T}|x_{t+1}\right)p\left(x_{t+1}|x_{t}\right)dx_{t+1}$
(from Eq.~\ref{eq:backward_CKE_deriv_1} to Eq.~\ref{eq:backward_CKE_deriv_2})
is the backward CKE in Eq.~\ref{eq:backward_CKE}. The result in
Eq.~\ref{eq:backward_CKE_deriv_6} is the backward CKE in Eq.~\ref{eq:backward_CKE_2}.

\subsection{Chapman-Kolmogorov equations for bridges\label{subsec:Proof_Chapman-Kolmogorov-equations-for-bridges}}

The CKE for bridges in Eq.~\ref{eq:bridge_CKE_not_learnable} can
be derived from the CKE for conditional Markov process in Eq.~\ref{eq:bridge_CKE}
by choosing $v$ to be $T$. However, deriving the CKE in Eq.~\ref{eq:bridge_CKE_learnable}
from Eq.~\ref{eq:bridge_CKE} is not straightforward as it involves
the integration w.r.t. $dx_{t}$ rather than $dx_{s}$ ($t<s$). It
suggests that we should consider the reverse of the original conditional
Markov process. Since it is another conditional Markov process (conditioned
on $\hat{x}_{0}=y_{A}$), it can be characterized by the following
CKE:
\begin{align}
p\left(\hat{x}_{T}|x_{s},\hat{x}_{0}\right) & =\int p\left(\hat{x}_{T}|x_{t},\hat{x}_{0}\right)p\left(x_{t}|x_{s},\hat{x}_{0}\right)dx_{t}\label{eq:derived_CKE_learnable_1}\\
\Rightarrow\frac{p\left(x_{s}|\hat{x}_{T},\hat{x}_{0}\right)p\left(\hat{x}_{T}|\hat{x}_{0}\right)}{p\left(x_{s}|\hat{x}_{0}\right)} & =\int\frac{p\left(x_{t}|\hat{x}_{T},\hat{x}_{0}\right)p\left(\hat{x}_{T}|\hat{x}_{0}\right)}{p\left(x_{t}|\hat{x}_{0}\right)}p\left(x_{t}|x_{s},\hat{x}_{0}\right)dx_{t}\label{eq:derived_CKE_learnable_2}\\
\Rightarrow p\left(x_{s}|\hat{x}_{T},\hat{x}_{0}\right) & =\int\frac{p\left(x_{t}|x_{s},\hat{x}_{0}\right)p\left(x_{s}|\hat{x}_{0}\right)}{p\left(x_{t}|\hat{x}_{0}\right)}p\left(x_{t}|\hat{x}_{T},\hat{x}_{0}\right)dx_{t}\label{eq:derived_CKE_learnable_3}\\
\Rightarrow p\left(x_{s}|\hat{x}_{T},\hat{x}_{0}\right) & =\int p\left(x_{s}|x_{t},\hat{x}_{0}\right)p\left(x_{t}|\hat{x}_{T},\hat{x}_{0}\right)dx_{t}\label{eq:derived_CKE_learnable_4}
\end{align}
Here, $\hat{x}_{T}\sim p\left(\hat{x}_{T}|\hat{x}_{0}\right)$ with
$p\left(\hat{x}_{T}=y_{B}\right)=p\left(y_{B}|y_{A}\right)$. By writing
Eq.~\ref{eq:derived_CKE_learnable_3} with slightly different notations,
we obtain Eq.~\ref{eq:bridge_CKE_learnable}.

\subsection{Tweedie's formula for bridges\label{subsec:Tweedie's-formula-for-bridges}}

Assume that $x$ is sampled from a Gaussian distribution $p\left(x|y_{A},y_{B}\right)=\Normal\left(\alpha y_{A}+\beta y_{B},\sigma^{2}\mathrm{I}\right)$.
The posterior expectation of $y_{B}$ given $x$ and $y_{A}$ can
be computed as follows:
\begin{equation}
\tilde{y}_{B}=\Expect_{p\left(y_{B}|x,y_{A}\right)}\left[y_{B}\right]=x-\alpha y_{A}+\sigma^{2}\nabla\log p\left(x|y_{A}\right)\label{eq:Tweedie_bridge}
\end{equation}
where $p\left(x|y_{A}\right)=\Expect_{p\left(y_{B}|y_{A}\right)}\left[p\left(x|y_{A},y_{B}\right)\right]$.
We refer to Eq.~\ref{eq:Tweedie_bridge} as Tweedie's formula for
bridges.

We start by representing $\nabla\log p\left(x|y_{A}\right)$ as follows:
\begin{align}
 & \nabla\log p\left(x|y_{A}\right)\\
=\  & \frac{1}{p\left(x|y_{A}\right)}\nabla p\left(x|y_{A}\right)\\
=\  & \frac{1}{p\left(x|y_{A}\right)}\nabla\int p\left(x|y_{A},y_{B}\right)p\left(y_{B}|y_{A}\right)dy_{B}\\
=\  & \frac{1}{p\left(x|y_{A}\right)}\int p\left(y_{B}|y_{A}\right)\nabla p\left(x|y_{A},y_{B}\right)dy_{B}\\
=\  & \int\frac{p\left(y_{B}|y_{A}\right)p\left(x|y_{A},y_{B}\right)}{p\left(x|y_{A}\right)}\nabla\log p\left(x|y_{A},y_{B}\right)dy_{B}\\
=\  & \int p\left(y_{B}|x,y_{A}\right)\left(\frac{\alpha y_{A}+\beta y_{B}-x}{\sigma^{2}}\right)dy_{B}\\
=\  & \frac{\alpha y_{A}+\beta\Expect_{p\left(y_{B}|x,y_{A}\right)}\left[y_{B}\right]-x}{\sigma^{2}}\label{eq:Tweedie_bridge_score_6}
\end{align}
Rearrange Eq.~\ref{eq:Tweedie_bridge_score_6}, we have:
\begin{equation}
\tilde{y}_{B}=\Expect_{p\left(y_{B}|x,y_{A}\right)}\left[y_{B}\right]=\frac{1}{\beta}\left(x-\alpha y_{A}+\sigma^{2}\nabla\log p\left(x|y_{A}\right)\right)\label{eq:Tweedie_yB_approx}
\end{equation}
Since $p\left(x|y_{A},y_{B}\right)=\Normal\left(\alpha y_{A}+\beta y_{B},\sigma^{2}\mathrm{I}\right)$,
$x$ can be represented as $x=\alpha y_{A}+\beta y_{B}+\sigma z$,
which means:
\begin{equation}
y_{B}=\frac{1}{\beta}\left(x-\alpha y_{A}-\sigma z\right)\label{eq:Tweedie_yB_orig}
\end{equation}
Eqs.~\ref{eq:Tweedie_yB_approx}, \ref{eq:Tweedie_yB_orig} suggest
that $-\sigma\nabla\log p\left(x|y_{A}\right)$ is the least square
approximation of $z$. This means $z_{\theta}\left(t,x_{t},x_{0}\right)$
in Eq.~\ref{eq:loss_bridge_2} should equal to $-\sigma\nabla\log p\left(x|x_{0}\right)$.

\subsection{Connection between the CKE framework and other frameworks for bridges\label{subsec:Connection-between-the-CKE-framework}}

\subsubsection{Link to variational inference}

If we assume the generative process is a discrete-time \emph{conditional
Markov process} running from time $0$ to time $T$ with the initial
distribution $p\left(x_{0}|\hat{x}_{0}\right)$ being a Dirac distribution
at $\hat{x}_{0}$ (i.e., $p\left(x_{0}|\hat{x}_{0}\right)=\delta_{\hat{x}_{0}}$),
the generative distribution over all time steps will be given below:
\begin{equation}
p_{\theta}\left(x_{0:T}|\hat{x}_{0}\right)=p\left(x_{0}|\hat{x}_{0}\right)\prod_{t=0}^{T-1}p_{\theta}\left(x_{t+1}|x_{t},\hat{x}_{0}\right)
\end{equation}
Here, $x_{0}$, ..., $x_{T-1}$ are regarded as latent variables and
$x_{T}$ is regarded as an observed variable. The (variational) inference
distribution $q\left(x_{0:T-1}|\hat{x}_{T},\hat{x}_{0}\right)$ can
be factorized as follows:
\begin{align}
 & q\left(x_{0:T-1}|\hat{x}_{T},\hat{x}_{0}\right)\nonumber \\
=\  & q\left(x_{T-1}|\hat{x}_{T},\hat{x}_{0}\right)\prod_{t=0}^{T-2}q\left(x_{t}|x_{t+1},\hat{x}_{T},\hat{x}_{0}\right)\\
=\  & q\left(x_{T-1}|\hat{x}_{T},\hat{x}_{0}\right)\prod_{t=0}^{T-2}\frac{q\left(x_{t+1}|x_{t},\hat{x}_{T},\hat{x}_{0}\right)q\left(x_{t}|\hat{x}_{T},\hat{x}_{0}\right)}{q\left(x_{t+1}|\hat{x}_{T},\hat{x}_{0}\right)}\\
=\  & q\left(x_{0}|\hat{x}_{T},\hat{x}_{0}\right)\prod_{t=0}^{T-2}q\left(x_{t+1}|x_{t},\hat{x}_{T},\hat{x}_{0}\right)
\end{align}
which characterizes a double conditional Markov process with Dirac
distributions $\delta_{\hat{x}_{0}}$ and $\delta_{\hat{x}_{T}}$
at both ends and the transition kernel $q\left(x_{t+1}|x_{t},\hat{x}_{T},\hat{x}_{0}\right)$.

We can learn $\theta$ by minimizing the negative variational lower
bound below:
\begin{align}
 & -\Expect_{p\left(\hat{x}_{0}\right)p\left(\hat{x}_{T}|\hat{x}_{0}\right)}\left[\text{ELBO}\left(\hat{x}_{T},\hat{x}_{0}\right)\right]\nonumber \\
=\  & \Expect_{p\left(\hat{x}_{0}\right)p\left(\hat{x}_{T}|\hat{x}_{0}\right)}\left[\Expect_{q\left(x_{0:T-1}|\hat{x}_{T},\hat{x}_{0}\right)}\left[-\log\frac{p_{\theta}\left(x_{0:T}|\hat{x}_{0}\right)}{q\left(x_{0:T-1}|\hat{x}_{T},\hat{x}_{0}\right)}\right]\right]\\
=\  & -\log p_{\theta}\left(x_{T}|x_{T-1},\hat{x}_{0}\right)\nonumber \\
 & +\sum_{t=1}^{T-1}D_{\text{KL}}\left(q\left(x_{t+1}|x_{t},\hat{x}_{T},\hat{x}_{0}\right)\|p_{\theta}\left(x_{t+1}|x_{t},\hat{x}_{0}\right)\right)\nonumber \\
 & +D_{\text{KL}}\left(q\left(x_{0}|\hat{x}_{T},\hat{x}_{0}\right)\|p\left(x_{0}|\hat{x}_{0}\right)\right)\label{eq:ELBO_2}
\end{align}
The KL term in Eq.~\ref{eq:ELBO_2} is the discrete-time version
of our loss in Eq.~\ref{eq:loss_bridge_1}.

\subsubsection{Link to score matching}

When the Markov process between $\hat{x}_{0}$, $\hat{x}_{T}$ is
a continuous-time diffusion process, the problem of matching $p_{\theta}\left(x_{s}|x_{t},\hat{x}_{0}\right)$
to $q\left(x_{s}|x_{t},\hat{x}_{T},\hat{x}_{0}\right)$ in Eq.~\ref{eq:loss_bridge_1}
can be reformulated in the differential form as matching $\frac{\partial}{\partial t}p_{\theta}\left(x_{t}|\hat{x}_{0}\right)$
to $\frac{\partial}{\partial t}q\left(x_{t}|\hat{x}_{T},\hat{x}_{0}\right)$
where $q\left(x_{t}|\hat{x}_{T},\hat{x}_{0}\right)$ is the marginal
distribution at time $t$ of the diffusion process between $\hat{x}_{0}$,
$\hat{x}_{T}$. Given the connection between $\frac{\partial p}{\partial t}$
and $\nabla p$ via the KBE (Eq.~\ref{eq:KBE_1}), we can instead
match $\nabla p_{\theta}\left(x_{t}|\hat{x}_{0}\right)$ to $\nabla q\left(x_{t}|\hat{x}_{T},\hat{x}_{0}\right)$,
which is similar to matching $\nabla\log p_{\theta}\left(x_{t}|\hat{x}_{0}\right)$
to $\nabla\log q\left(x_{t}|\hat{x}_{T},\hat{x}_{0}\right)$.

\subsubsection{Link to Doob's $h$-transform}

We consider a slightly different setting for bridges: Instead of starting
a Markov process from a specific initial sample $\hat{x}_{0}=y_{A}$
and ensure that the final distribution $p\left(x_{T}|\hat{x}_{0}\right)$
will satisfy $p\left(x_{T}=y_{B}|\hat{x}_{0}\right)=p\left(y_{B}|y_{A}\right)$,
we start the process from an initial distribution of $x_{0}$ and
force it to hit a predetermined sample $\hat{x}_{T}=y_{B}$ at time
$T$ almost surely. If the initial distribution $p\left(x_{0}\right)$
is chosen such that $p\left(x_{0}=y_{A}\right)=p\left(y_{A}|y_{B}\right)$,
then the two settings are statistically equivalent when all samples
from the two domains $A$, $B$ are counted.

Let $p\left(x_{t}\right)$ be the marginal distribution at time $t$
corresponding to a Markov process starting from the initial distribution
$p\left(x_{0}\right)$. Also assume that $p\left(x_{t}\right)$ has
support over the entire sample space. Then, we have:
\begin{align}
p\left(\hat{x}_{T}\right) & =\int p\left(\hat{x}_{T}|x_{t}\right)p\left(x_{t}\right)dx_{t}\label{eq:Doob_integral_t}
\end{align}
Interestingly, we can define a \emph{new} marginal distribution of
$x_{t}$ as $\tilde{p}\left(x_{t}\right)=\frac{p\left(\hat{x}_{T}|x_{t}\right)p\left(x_{t}\right)}{p\left(\hat{x}_{T}\right)}$,
and if this distribution converges to a Dirac distribution at time
$T$ then, under some mild conditions\footnote{A key condition here is that $\hat{p}\left(x_{t}=y_{B}\right)$ does
not vanish $\forall$ $t$.}, this Dirac distribution should center around $\hat{x}_{T}=y_{B}$.

At time $s\neq t$, Eq.~\ref{eq:Doob_integral_t} becomes: 
\begin{align}
p\left(\hat{x}_{T}\right) & =\int p\left(\hat{x}_{T}|x_{s}\right)p\left(x_{s}\right)dx_{s}\\
 & =\int p\left(\hat{x}_{T}|x_{s}\right)\left(\int p\left(x_{s}|x_{t}\right)p\left(x_{t}\right)dx_{t}\right)dx_{s}\\
 & =\int\int p\left(\hat{x}_{T}|x_{s}\right)p\left(x_{s}|x_{t}\right)p\left(x_{t}\right)dx_{t}dx_{s}\\
 & =\int\left(\int p\left(\hat{x}_{T}|x_{s}\right)p\left(x_{s}|x_{t}\right)dx_{s}\right)p\left(x_{t}\right)dx_{t}\label{eq:Doob_integral_s4}
\end{align}
Since $p\left(\hat{x}_{T}\right)$ in Eq.~\ref{eq:Doob_integral_t}
is the same as in Eq.~\ref{eq:Doob_integral_s4}, the CKE below should
hold for every $s\neq t$:

\begin{align}
p\left(\hat{x}_{T}|x_{t}\right) & =\int p\left(\hat{x}_{T}|x_{s}\right)p\left(x_{s}|x_{t}\right)dx_{s}\label{eq:Doob_CKE_1}\\
 & =\Expect\left[p\left(\hat{x}_{T}|X_{s}\right)|X_{t}=x_{t}\right]\label{eq:Doob_CKE_2}
\end{align}
Here, we focus on the generative setting with $0<t<s$ and rewrite
Eq.~\ref{eq:Doob_CKE_1} as follows:
\begin{equation}
1=\int p\left(x_{s}|x_{t}\right)\frac{p\left(\hat{x}_{T}|x_{s}\right)}{p\left(\hat{x}_{T}|x_{t}\right)}dx_{s}\label{eq:Doob_transition_kernel}
\end{equation}
Eq.~\ref{eq:Doob_transition_kernel} suggests that we can set $p\left(x_{s}|x_{t}\right)\frac{p\left(\hat{x}_{T}|x_{s}\right)}{p\left(\hat{x}_{T}|x_{t}\right)}$
to be a distribution over $x_{s}$. Let us denote $\tilde{p}\left(x_{s}|x_{t}\right)=p\left(x_{s}|x_{t}\right)\frac{p\left(\hat{x}_{T}|x_{s}\right)}{p\left(\hat{x}_{T}|x_{t}\right)}$,
then $\tilde{p}\left(x_{s}|x_{t}\right)$ can be viewed as the transition
kernel of another Markov process derived from the original Markov
process. Interestingly, $\tilde{p}\left(x_{t}\right)$ is the marginal
distribution at time $t$ of this process, and since $\tilde{p}\left(\hat{x}_{T}\right)$
is a Dirac distribution at $\hat{x}_{T}$, this process converges
to $\hat{x}_{T}=y_{B}$ almost surely. Please refer to the last part
of this subsection for detail proofs.

It is worth noting that in Eq.~\ref{eq:Doob_integral_t}, the term
$p\left(x_{t}\right)$ is fixed since it is the marginal distribution
of the (predefined) original Markov process while the term $p\left(\hat{x}_{T}|x_{t}\right)$
can vary freely as long as it satisfies Eq.~\ref{eq:Doob_CKE_1}.
Therefore, if we let $h\left(\cdot,\cdot,T,\hat{x}_{T}\right)$ be
any function such that:
\begin{align}
h\left(t,x_{t},T,\hat{x}_{T}\right) & =\int h\left(s,x_{s},T,\hat{x}_{T}\right)p\left(x_{s}|x_{t}\right)dx_{s}\label{eq:Doob_h_CKE_1}\\
 & =\Expect\left[h\left(s,X_{s},T,\hat{x}_{T}\right)|X_{t}=x_{t}\right]\label{eq:Doob_h_CKE_2}
\end{align}
and $h\left(T,x_{T},T,\hat{x}_{T}\right)=\delta_{\hat{x}_{T}}\left(x_{T}\right)$
then by setting $\tilde{p}\left(x_{s}|x_{t}\right)=p\left(x_{s}|x_{t}\right)\frac{h\left(s,x_{s},T,\hat{x}_{T}\right)}{h\left(t,x_{t},T,\hat{x}_{T}\right)}$,
we obtain a new Markov process called \emph{Doob's $h$-transform
process} that converges to $\hat{x}_{T}=y_{B}$ almost surely. This
is the main idea behind Doob's $h$-transform \cite{doob1984classical}.

In the continuous-time setting, Eq.~\ref{eq:Doob_CKE_1} can be written
in the differential form below:
\begin{equation}
\begin{cases}
\mathcal{A}_{t}h\left(t,x_{t},T,\hat{x}_{T}\right)=0\\
h\left(T,x_{T},T,\hat{x}_{T}\right)=\delta_{\hat{x}_{T}}\left(x_{T}\right)
\end{cases}\label{eq:Doob_h_cond}
\end{equation}
where $\mathcal{A}_{t}$ is the generator operator defined as $\mathcal{A}_{t}f\left(t,x_{t}\right)\triangleq\lim_{\Delta t\downarrow0}\frac{\Expect\left[f\left(t+\Delta t,X_{t+\Delta t}\right)|X_{t}=x_{t}\right]-f\left(t,x_{t}\right)}{\Delta t}$.
The above equation is in fact a KBE. When the original Markov process
is a continuous-time diffusion process described by the SDE $dX_{t}=\mu\left(t,X_{t}\right)dt+\sigma\left(t\right)dW_{t}$,
given any real-valued function $f\left(t,x\right)$, $\mathcal{A}_{t}f\left(t,x\right)$
can be represented as follows:
\begin{align*}
\mathcal{A}_{t}f & =\frac{\partial f}{\partial t}+\nabla f\cdot\mu+\frac{\sigma^{2}}{2}\Delta f\\
 & =\frac{\partial f}{\partial t}+\mathcal{G}f
\end{align*}
The generator $\mathcal{A}_{t}^{h}$ of the \emph{Doob's $h$-transform
process} can be derived from $\mathcal{A}_{t}$ as follows:
\begin{align*}
\mathcal{A}_{t}^{h}f & =\frac{1}{h}\mathcal{A}_{t}\left(fh\right)
\end{align*}
By leveraging the fact that $\mathcal{A}_{t}h=0$ in Eq.~\ref{eq:Doob_h_cond},
$\mathcal{A}_{t}^{h}f$ can be expressed as follows:
\[
\mathcal{A}_{t}^{h}f=\frac{\partial f}{\partial t}+\nabla f\cdot\left(\mu+\sigma^{2}\nabla\log h\right)+\frac{\sigma^{2}}{2}\Delta f
\]
It implies that this diffusion process is described by the SDE:
\[
dX_{t}=\left(\mu\left(t,X_{t}\right)+\sigma^{2}\nabla\log h\left(t,X_{t},T,\hat{x}_{T}\right)\right)dt+\sigma\left(t\right)dW_{t}
\]

\paragraph{Proofs for some properties of $\tilde{p}\left(x_{s}|x_{t}\right)$
and $\tilde{p}\left(x_{t}\right)$}

For any times $0\leq t<r<s$, we have:
\begin{align}
 & \int\tilde{p}\left(x_{s}|x_{r}\right)\tilde{p}\left(x_{r}|x_{t}\right)dx_{r}\nonumber \\
=\  & \int p\left(x_{s}|x_{r}\right)\frac{p\left(\hat{x}_{T}|x_{s}\right)}{\cancel{p\left(\hat{x}_{T}|x_{r}\right)}}p\left(x_{r}|x_{t}\right)\frac{\cancel{p\left(\hat{x}_{T}|x_{r}\right)}}{p\left(\hat{x}_{T}|x_{t}\right)}dx_{r}\\
=\  & \frac{p\left(\hat{x}_{T}|x_{s}\right)}{p\left(\hat{x}_{T}|x_{t}\right)}\int p\left(x_{s}|x_{r}\right)p\left(x_{r}|x_{t}\right)dx_{r}\\
=\  & \frac{p\left(\hat{x}_{T}|x_{s}\right)}{p\left(\hat{x}_{T}|x_{t}\right)}p\left(x_{s}|x_{t}\right)\\
=\  & \tilde{p}\left(x_{s}|x_{t}\right)
\end{align}
The last equation implies that $\tilde{p}\left(x_{s}|x_{t}\right)$
satisfies the CKE and is the transition probability of a Markov process.
Besides, we have:
\begin{align}
 & \int\tilde{p}\left(x_{s}|x_{t}\right)\tilde{p}\left(x_{t}\right)dx_{t}\nonumber \\
=\  & \int p\left(x_{s}|x_{t}\right)\frac{p\left(\hat{x}_{T}|x_{s}\right)}{\cancel{p\left(\hat{x}_{T}|x_{t}\right)}}\frac{\cancel{p\left(\hat{x}_{T}|x_{t}\right)}p\left(x_{t}\right)}{p\left(\hat{x}_{T}\right)}dx_{t}\\
=\  & \frac{p\left(\hat{x}_{T}|x_{s}\right)}{p\left(\hat{x}_{T}\right)}\int p\left(x_{s}|x_{t}\right)p\left(x_{t}\right)dx_{t}\\
=\  & \frac{p\left(\hat{x}_{T}|x_{s}\right)p\left(x_{s}\right)}{p\left(\hat{x}_{T}\right)}\\
=\  & \tilde{p}\left(x_{s}\right)
\end{align}
which means $\tilde{p}\left(x_{t}\right)$ is the marginal distribution
at time $t$ of the Markov process characterized by $\tilde{p}\left(x_{s}|x_{t}\right)$.

\subsection{Derivation of transitions in Eq.~\ref{eq:ref_forward_mean_1} and
Eq.~\ref{eq:ref_backward_mean_1}}

We consider the case where marginal distributions at timestep $t$
and $s$ (with $t<s$) are $q\left(x_{t}|x_{0},x_{T}\right)=\Normal\left(\alpha_{t}x_{0}+\beta_{t}x_{T},\sigma_{t}^{2}\mathrm{I}\right)$
and $q\left(x_{s}|x_{0},x_{T}\right)=\Normal\left(\alpha_{s}x_{0}+\beta_{s}x_{T},\sigma_{s}^{2}\mathrm{I}\right)$,
respectively. We detail the derivation of our proposed forward transition
distribution, denoted as $q\left(x_{s}|x_{t},x_{0},x_{T}\right)$,
and backward transition distribution, denoted as $q\left(x_{t}|x_{s},x_{0},x_{T}\right)$.

\subsubsection{Derivation of forward transition $q\left(x_{s}|x_{t},x_{0},x_{T}\right)$
in Eq.~\ref{eq:ref_forward_mean_1}}

Recall that the forward CKE, from $t$ to $s$, given two endpoints
$x_{0}$ and $x_{T}$ is given by:
\[
q\left(x_{s}|x_{0},x_{T}\right)=\int q\left(x_{s}|x_{t},x_{0},x_{T}\right)q\left(x_{t}|x_{0},x_{T}\right)dx_{t}
\]

\noindent where $q\left(x_{s}|x_{t},x_{0},x_{T}\right)$ replace $p_{\theta}\left(x_{s}|x_{t},x_{0}\right)$
in Eq.~\ref{eq:bridge_CKE_learnable} in case we align $p_{\theta}\left(x_{s}|x_{t},\hat{x}_{0}\right)$
with $q\left(x_{s}|x_{t},\hat{x}_{T},\hat{x}_{0}\right)$. Following
\cite{bishop2006pattern} (Eq. 2.115), we assume that $q\left(x_{s}|x_{t},x_{0},x_{T}\right)=\mathcal{N}\left(ax_{t}+bx_{0}+cx_{T}+d,\delta_{s,t}^{2}\mathrm{I}\right)$
and we have:
\begin{align}
\Expect\left[\begin{array}{c}
x_{t}|x_{0},x_{T}\\
x_{s}|x_{t},x_{0},x_{T}
\end{array}\right] & =\left(\begin{array}{c}
\alpha_{t}x_{0}+\beta_{t}x_{T}\\
a\left(\alpha_{t}x_{0}+\beta_{t}x_{T}\right)+bx_{0}+cx_{T}+d
\end{array}\right)\\
\text{Cov} & =\left(\begin{array}{cc}
\diag\left(\sigma_{t}^{2}\right) & \diag\left(a\sigma_{t}^{2}\right)\\
\diag\left(a\sigma_{t}^{2}\right) & \diag\left(\delta_{s,t}^{2}+a^{2}\sigma_{t}^{2}\right)
\end{array}\right)
\end{align}

\noindent Compare the mean and covariance with that of $q\left(x_{s}\mid x_{0},x_{T}\right)$,
we have:
\begin{equation}
\begin{cases}
d=0\\
a\left(\alpha_{t}x_{0}+\beta_{t}x_{T}\right)+bx_{0}+cx_{T}=\alpha_{s}x_{0}+\beta_{s}x_{T}\\
\delta_{s,t}^{2}+a^{2}\sigma_{t}^{2}=\sigma_{s}^{2}
\end{cases}
\end{equation}

\noindent 
\begin{equation}
\Rightarrow\begin{cases}
a= & \frac{\sqrt{\sigma_{s}^{2}-\delta_{s,t}^{2}}}{\sigma_{t}}\\
b= & \alpha_{s}-\alpha_{t}\frac{\sqrt{\sigma_{s}^{2}-\delta_{s,t}^{2}}}{\sigma_{t}}\\
c= & \beta_{s}-\beta_{t}\frac{\sqrt{\sigma_{s}^{2}-\delta_{s,t}^{2}}}{\sigma_{t}}
\end{cases}
\end{equation}
\begin{align}
 & q\left(x_{s}|x_{t},x_{0},x_{T}\right)\nonumber \\
=\  & \mathcal{N}\left(\left(\frac{\sqrt{\sigma_{s}^{2}-\delta_{s,t}^{2}}}{\sigma_{t}}\right)x_{t}+\left(\alpha_{s}-\alpha_{t}\frac{\sqrt{\sigma_{s}^{2}-\delta_{s,t}^{2}}}{\sigma_{t}}\right)x_{0}+\left(\beta_{s}-\beta_{t}\frac{\sqrt{\sigma_{s}^{2}-\delta_{s,t}^{2}}}{\sigma_{t}}\right)x_{T},\delta_{s,t}^{2}\mathrm{I}\right)\\
=\  & \mathcal{N}\left(\alpha_{s}x_{0}+\beta_{s}x_{T}+\sqrt{\sigma_{s}^{2}-\delta_{s,t}^{2}}\frac{\left(x_{t}-\alpha_{t}x_{0}-\beta_{t}x_{T}\right)}{\sigma_{t}},\delta_{s,t}^{2}\mathrm{I}\right)
\end{align}

\subsubsection{Derivation of backward transition $q\left(x_{t}|x_{s},x_{0},x_{T}\right)$
in Eq.~\ref{eq:ref_backward_mean_1}}

Recall that from \ref{eq:Bayes_rule_bridge}, we can derive $q\left(x_{t}|x_{s},x_{0},x_{T}\right)$
from Bayes' rule:
\begin{equation}
q\left(x_{t}|x_{s},x_{0},x_{T}\right)=q\left(x_{s}|x_{t},x_{0},x_{T}\right)\frac{q\left(x_{t}|x_{0},x_{T}\right)}{q\left(x_{s}|x_{0},x_{T}\right)}
\end{equation}

\noindent With: 
\begin{align}
 & q\left(x_{s}\mid x_{t},x_{0},x_{T}\right)\nonumber \\
=\  & \frac{1}{\sqrt{2\pi}\delta_{s,t}}\exp\left(-\frac{1}{2}\frac{\left(x_{s}-\left(\frac{\sqrt{\sigma_{s}^{2}-\delta_{s,t}^{2}}}{\sigma_{t}}x_{t}+\left(\alpha_{s}-\alpha_{t}\frac{\sqrt{\sigma_{s}^{2}-\delta_{s,t}^{2}}}{\sigma_{t}}\right)x_{0}+\left(\beta_{s}-\beta_{t}\frac{\sqrt{\sigma_{s}^{2}-\delta_{s,t}^{2}}}{\sigma_{t}}\right)x_{T}\right)\right)^{2}}{\delta_{s,t}^{2}}\right)
\end{align}
\begin{align}
q\left(x_{t}\mid x_{0},x_{T}\right) & =\frac{1}{\sqrt{2\pi}\sigma_{t}}\exp\left(-\frac{1}{2}\frac{\left(x_{t}-\left(\alpha_{t}x_{0}+\beta_{t}x_{T}\right)\right)^{2}}{\sigma_{t}^{2}}\right)\\
q\left(x_{s}\mid x_{0},x_{T}\right) & =\frac{1}{\sqrt{2\pi}\sigma_{s}}\exp\left(-\frac{1}{2}\frac{\left(x_{s}-\left(\alpha_{s}x_{0}+\beta_{s}x_{T}\right)\right)^{2}}{\sigma_{s}^{2}}\right)
\end{align}

\noindent Then we know: 
\begin{align}
 & q\left(x_{t}|x_{s},x_{0},x_{T}\right)\nonumber \\
=\  & \frac{1}{\sqrt{2\pi}\frac{\delta_{s,t}\sigma_{t}}{\sigma_{s}}}\exp\left[-\frac{1}{2}\left(\frac{\left(x_{s}-\left(\frac{\sqrt{\sigma_{s}^{2}-\delta_{s,t}^{2}}}{\sigma_{t}}x_{t}+\left(\alpha_{s}-\alpha_{t}\frac{\sqrt{\sigma_{s}^{2}-\delta_{s,t}^{2}}}{\sigma_{t}}\right)x_{0}+\left(\beta_{s}-\beta_{t}\frac{\sqrt{\sigma_{s}^{2}-\delta_{s,t}^{2}}}{\sigma_{t}}\right)x_{T}\right)\right)^{2}}{\delta_{s,t}^{2}}\right.\right.\nonumber \\
+ & \left.\left.\frac{\left(x_{t}-\left(\alpha_{t}x_{0}+\beta_{t}x_{T}\right)\right)^{2}}{\sigma_{t}^{2}}-\frac{\left(x_{s}-\left(\alpha_{s}x_{0}+\beta_{s}x_{T}\right)\right)^{2}}{\sigma_{s}^{2}}\right)\right]\\
=\  & \frac{1}{\sqrt{2\pi}\frac{\delta_{s,t}\sigma_{t}}{\sigma_{s}}}\exp\left(-\frac{\left(x_{t}-\tilde{\mu}_{t}\right)^{2}}{2\left(\frac{\delta_{s,t}\sigma_{t}}{\sigma_{s}}\right)^{2}}\right)
\end{align}

\noindent where
\begin{align}
\tilde{\mu}_{t}= & \left(\frac{\sigma_{t}\sqrt{\sigma_{s}^{2}-\delta_{s,t}^{2}}}{\sigma_{s}^{2}}\right)x_{s}+\left(\alpha_{t}-\alpha_{s}\frac{\sigma_{t}\sqrt{\sigma_{s}^{2}-\delta_{s,t}^{2}}}{\sigma_{s}^{2}}\right)x_{0}+\left(\beta_{t}-\beta_{s}\frac{\sigma_{t}\sqrt{\sigma_{s}^{2}-\delta_{s,t}^{2}}}{\sigma_{s}^{2}}\right)x_{T}\nonumber \\
= & \alpha_{t}x_{0}+\beta_{t}x_{T}+\sigma_{t}\sqrt{\sigma_{s}^{2}-\delta_{s,t}^{2}}\frac{\left(x_{s}-\alpha_{s}x_{0}-\beta_{s}x_{T}\right)}{\sigma_{s}^{2}}
\end{align}

\subsection{Special variants of BDBM\label{subsec:Special-cases-of-BDBM}}

Below, we discuss several important variants of BDBM. These variants
mainly correspond to the variability of $\delta_{s,t}$ within the
interval $[0,\sigma_{s})$.

\subsubsection{$\delta_{s,t}=0$}

If we set $\delta_{s,t}=0$, then $p_{\theta}\left(x_{s}|x_{t},x_{0}\right)$
will become the deterministic mapping $\mu_{\theta}\left(s,t,x_{t},x_{0}\right)$
from $x_{t}$, $x_{0}$ to $x_{s}$. Similarly, $p_{\phi}\left(x_{t}|x_{s},x_{T}\right)$
will become the deterministic mapping $\tilde{\mu}_{\phi}\left(t,s,x_{s},x_{T}\right)$
from $x_{s}$, $x_{T}$ to $x_{t}$. This variant links to the deterministic
mapping from $x_{t}$ ($x$) to $x_{}$ ($x_{t}$) in DDIM \cite{song2020denoising}.

\subsubsection{$\delta_{s,t}=\sqrt{\sigma_{s}^{2}-\sigma_{t}^{2}\frac{\beta_{s}^{2}}{\beta_{t}^{2}}}$}

When $\delta_{s,t}=\sqrt{\sigma_{s}^{2}-\sigma_{t}^{2}\frac{\beta_{s}^{2}}{\beta_{t}^{2}}}$,
$\mu\left(s,t,x_{t},x_{0},x_{T}\right)$ (Eq.~\ref{eq:ref_forward_mean_1})
and $\tilde{\mu}\left(t,s,x_{s},x_{0},x_{T}\right)$ (Eq.~\ref{eq:ref_backward_mean_1})
become:
\begin{align}
\mu\left(s,t,x_{t},x_{0},x_{T}\right) & =\frac{\beta_{s}}{\beta_{t}}x_{t}+\left(\alpha_{s}-\alpha_{t}\frac{\beta_{s}}{\beta_{t}}\right)x_{0}\label{eq:mu_special_1}\\
\tilde{\mu}\left(t,s,x_{s},x_{0},x_{T}\right) & =\frac{\sigma_{t}^{2}}{\sigma_{s}^{2}}\frac{\beta_{s}}{\beta_{t}}x_{s}+\left(\alpha_{t}-\alpha_{s}\frac{\sigma_{t}^{2}}{\sigma_{s}^{2}}\frac{\beta_{s}}{\beta_{t}}\right)x_{0}+\left(\beta_{t}-\beta_{s}\frac{\sigma_{t}^{2}}{\sigma_{s}^{2}}\frac{\beta_{s}}{\beta_{t}}\right)x_{T}\label{eq:mu_tilde_special_1}
\end{align}

Although the term containing $x_{T}$ in Eq.~\ref{eq:mu_special_1}
vanishes, $\mu\left(s,t,x_{t},x_{0},x_{T}\right)$ still depends on
$x_{T}$ since $x_{t}$ depends on $x_{T}$ via Eq.~\ref{eq:ref_bridge_xt_decomp}.
In this case, if $x_{T}$ is modeled directly via $x_{T,\theta}$,
then setting $x_{t}=x_{0}$ at the initial sampling step $t=0$ will
lead to poor generation results since $\mu_{\theta}\left(t,x_{t},x_{0}\right)$
no longer depends on $x_{T,\theta}\left(t,x_{t},x_{0}\right)$. Instead,
we have to set $x_{t}=\alpha_{\epsilon}x_{0}+\beta_{\epsilon}x_{T,\theta}\left(\epsilon,x_{0},x_{0}\right)$
where $\epsilon$ is a small value such that $\beta_{\epsilon}\neq\beta_{0}=0$.
This will ensure that $\mu_{\theta}\left(t,x_{t},x_{0}\right)$ uses
the knowledge from $x_{T,\theta}\left(\epsilon,x_{0},x_{0}\right)$. 

The term containing $x_{T}$ in Eq.~\ref{eq:mu_tilde_special_1}
is unlikely to vanish because otherwise, this will lead to $\frac{\beta_{t}}{\beta_{s}}=\frac{\sigma_{t}}{\sigma_{s}}$
for every time pair $\left(t,s\right)$. This equation does not hold
since if we choose $t=T$ and choose $s$ such that $\beta_{s},\sigma_{s}\neq0$,
we have $\frac{\beta_{T}}{\beta_{s}}=\frac{1}{\beta_{s}}\neq\frac{0}{\sigma_{s}}=\frac{\sigma_{T}}{\sigma_{s}}$.
The term containing $x_{0}$ in Eq.~\ref{eq:mu_tilde_special_1},
by contrast, can vanish if $\alpha_{t}-\alpha_{s}\frac{\sigma_{t}^{2}}{\sigma_{s}^{2}}\frac{\beta_{s}}{\beta_{t}}=0$,
or equivalently, $\sigma_{t}^{2}=k\alpha_{t}\beta_{t}$ where $k>0$
is a constant w.r.t. $t$.

\subsubsection{$\delta_{s,t}=\sqrt{\sigma_{s}^{2}-\sigma_{t}^{2}\frac{\alpha_{s}^{2}}{\alpha_{t}^{2}}}$}

When $\delta_{s,t}=\sqrt{\sigma_{s}^{2}-\sigma_{t}^{2}\frac{\alpha_{s}^{2}}{\alpha_{t}^{2}}}$,
$\mu\left(s,t,x_{t},x_{0},x_{T}\right)$ and $\tilde{\mu}\left(t,s,x_{s},x_{0},x_{T}\right)$
become:
\begin{align}
\mu\left(s,t,x_{t},x_{0},x_{T}\right) & =\frac{\alpha_{s}}{\alpha_{t}}x_{t}+\left(\beta_{s}-\beta_{t}\frac{\alpha_{s}}{\alpha_{t}}\right)x_{T}\label{eq:mu_special_2}\\
\tilde{\mu}\left(t,s,x_{s},x_{0},x_{T}\right) & =\frac{\sigma_{t}^{2}}{\sigma_{s}^{2}}\frac{\alpha_{s}}{\alpha_{t}}x_{s}+\left(\alpha_{t}-\alpha_{s}\frac{\sigma_{t}^{2}}{\sigma_{s}^{2}}\frac{\alpha_{s}}{\alpha_{t}}\right)x_{0}+\left(\beta_{t}-\beta_{s}\frac{\sigma_{t}^{2}}{\sigma_{s}^{2}}\frac{\alpha_{s}}{\alpha_{t}}\right)x_{T}\label{eq:mu_tilde_special_2}
\end{align}
In this case, there will be no problem during sampling with $\mu_{\theta}\left(t,x_{t},x_{0}\right)$
and $\mu_{\phi}\left(s,x_{s},x_{T}\right)$ since they always use
the knowledge from $x_{T,\theta}\left(t,x_{t},x_{0}\right)$ and $x_{0,\phi}\left(s,x_{s},x_{T}\right)$,
respectively. Note that the term containing $x_{T}$ in Eq.~\ref{eq:mu_tilde_special_2}
can vanish if $\sigma_{t}^{2}=k\alpha_{t}\beta_{t}$ ($k>0$) but
this does not affect sampling.

\subsubsection{Brownian Bridge\label{subsec:Brownian-Bridge_settings}}

A Brownian Bridge \cite{LiX0L23} is a special case of the generalized
diffusion bridge in which: 
\begin{align*}
\beta_{t} & =\frac{t}{T}\\
\alpha_{t} & =1-\beta_{t}=1-\frac{t}{T}\\
\sigma_{t}^{2} & =k\alpha_{t}\beta_{t}=k\frac{t}{T}\left(1-\frac{t}{T}\right)
\end{align*}
With this choice of $\alpha_{t}$, $\beta_{t}$, and $\sigma_{t}$,
we can easily prove that $\sigma_{s}^{2}-\sigma_{t}^{2}\frac{\alpha_{s}^{2}}{\alpha_{t}^{2}}\geq0$
for all $t<s$ as follows:
\begin{align*}
 & \sigma_{s}^{2}-\sigma_{t}^{2}\frac{\alpha_{s}^{2}}{\alpha_{t}^{2}}\geq0\\
\Leftrightarrow\  & \frac{\sigma_{s}^{2}}{\sigma_{t}^{2}}\geq\frac{\alpha_{s}^{2}}{\alpha_{t}^{2}}\\
\Leftrightarrow\  & \frac{\beta_{s}}{\beta_{t}}\geq\frac{\alpha_{s}}{\alpha_{t}}\\
\Leftrightarrow\  & \frac{s}{t}\geq\frac{T-s}{T-t}\\
\Leftrightarrow\  & sT\geq tT\\
\Leftrightarrow\  & s\geq t
\end{align*}
Therefore, we can set $\delta_{s,t}=\sqrt{\eta\left(\sigma_{s}^{2}-\sigma_{t}^{2}\frac{\alpha_{s}^{2}}{\alpha_{t}^{2}}\right)}$
with $\eta\in\left[0,1\right]$. 

When $\eta=1$, $\mu\left(s,t,x_{t},x_{0},x_{T}\right)$ and $\tilde{\mu}\left(t,s,x_{s},x_{0},x_{T}\right)$
become:
\begin{align}
\mu\left(s,t,x_{t},x_{0},x_{T}\right) & =\frac{\alpha_{s}}{\alpha_{t}}x_{t}+\left(\beta_{s}-\beta_{t}\frac{\alpha_{s}}{\alpha_{t}}\right)x_{T}\\
 & =\frac{T-s}{T-t}x_{t}+\frac{\left(s-t\right)T}{T-t}x_{T}
\end{align}
\begin{align}
\tilde{\mu}\left(t,s,x_{s},x_{0},x_{T}\right) & =\frac{\sigma_{t}^{2}}{\sigma_{s}^{2}}\frac{\alpha_{s}}{\alpha_{t}}x_{s}+\left(\alpha_{t}-\alpha_{s}\frac{\sigma_{t}^{2}}{\sigma_{s}^{2}}\frac{\alpha_{s}}{\alpha_{t}}\right)x_{0}\\
 & =\frac{\beta_{t}}{\beta_{s}}x_{s}+\left(\alpha_{t}-\alpha_{s}\frac{\beta_{t}}{\beta_{s}}\right)x_{0}\\
 & =\frac{t}{s}x_{s}+\frac{\left(s-t\right)T}{s}x_{0}
\end{align}

\subsection{Training and sampling algorithms for BDBM\label{subsec:Training-and-sampling_alg}}

In Algos.~\ref{alg:Training-BDBM},\ref{alg:Sampling-forward}, and
\ref{alg:Sampling-backward}, we provide the detailed training, forward
sampling and backward sampling algorithms for our proposed BDBM with
$z_{\varphi}\left(t,x_{t},\left(1-m\right)*x_{0},m*x_{T}\right)$
as the model.

\begin{algorithm}
\begin{algorithmic}[1]
\State {\bfseries Input:} $\alpha_t$, $\beta_t$, $\sigma_t$ as continuously differentiable functions of $t$ satisfying  $\alpha_{0}=\beta_{T}=1$ and $\alpha_{T}=\beta_{0}=\sigma_{0}=\sigma_{T}=0$     
\Repeat     
	\State $t \sim \mathcal{U} \left(0,T \right)$ 
	\State $x_0, x_T \sim p(y_A, y_B)$     
	\State $z \sim \mathcal{N}(0, \mathrm{I})$     
	\State $x_t = \alpha_t x_0 + \beta_t x_T + \sigma_t z$     
	\State $m \sim \mathcal{B}(0.5)$     
	\State Update $\varphi$ by minimizing $\mathcal{L}(\varphi) = \left\Vert z_{\varphi}\left(t,x_{t},\left(1-m\right)*x_{0},m*x_{T}\right)-z\right\Vert _{2}^{2}$     
\Until {converged} 
\end{algorithmic}\caption{Training BDBM \label{alg:Training-BDBM}}
\end{algorithm}

\begin{algorithm}
\begin{algorithmic}[1] 
\State {\bfseries Input:} $\alpha_t$, $\beta_t$, $\sigma_t$, $\delta_{s,t}$, trained $z_{\varphi}\left(t,x_{t},\left(1-m\right)*x_{0},m*x_{T}\right)$, $x_0$
\State $m=0$
\For{$t = 0$ \textbf{to} $T - \Delta t$}
	\State $s = t + \Delta t$
	\State $z_{t|0} = z_{\varphi}\left(t,x_{t},x_{0},0\right)$
    \If{$s = T$}
		\State $\epsilon = 0$
    \Else         
		\State $\epsilon \sim \mathcal{N}(0, \mathrm{I})$
    \EndIf     
	\State $x_{s}= \frac{\beta_{s}}{\beta_{t}} x_{t} + \left(\alpha_{s} - \alpha_{t}\frac{\beta_{s}}{\beta_{t}}\right)x_{0} + \left(\sqrt{\sigma_{s}^{2}-\delta_{s,t}^{2}}-\sigma_{t}\frac{\beta_{s}}{\beta_{t}}\right) z_{t|0} + \delta_{s,t} \epsilon$
\EndFor
\State 
\Return $x_{s}$ 
\end{algorithmic}

\caption{Generating $x_{T}$ given $x_{0}$ (forward)\label{alg:Sampling-forward}}
\end{algorithm}

\begin{algorithm}
\begin{algorithmic}[1] 
\State {\bfseries Input:} $\alpha_t$, $\beta_t$, $\sigma_t$, $\delta_{s,t}$, trained $z_{\varphi}\left(t,x_{t},\left(1-m\right)*x_{0},m*x_{T}\right)$, $x_T$
\State $m=1$
\For{$s = T$ \textbf{to} $\Delta t$}
	\State $t = s - \Delta t$
	\State $z_{s|T} = z_{\varphi}\left(s,x_{s},0,x_{T}\right)$
    \If{$t = 0$}
		\State $\epsilon = 0$
    \Else
        \State $\epsilon \sim \mathcal{N}(0, \mathrm{I})$
    \EndIf
    \State $x_{t} = \frac{\alpha_{t}}{\alpha_{s}}x_{s}+\left(\beta_{t}-\beta_{s}\frac{\alpha_{t}}{\alpha_{s}}\right)x_{T}+\left(\frac{\sigma_{t}\sqrt{\sigma_{s}^{2}-\delta_{s,t}^{2}}}{\sigma_{s}}-\sigma_{s}\frac{\alpha_{t}}{\alpha_{s}}\right) z_{s|T} + \frac{\delta_{s,t}\sigma_{t}}{\sigma_{s}}\epsilon$
\EndFor
\State 
\Return $x_{t}$ 
\end{algorithmic}

\caption{Generating $x_{0}$ given $x_{T}$ (backward)\label{alg:Sampling-backward}}
\end{algorithm}

\section{Additional Experimental Results}

\subsection{Additional qualitative results of BDBM\label{subsec:Additional-qualitative-results}}

\noindent 
\begin{figure}
\begin{centering}
{\resizebox{1\textwidth}{!}{
\begin{tabular}{ccc}
\includegraphics{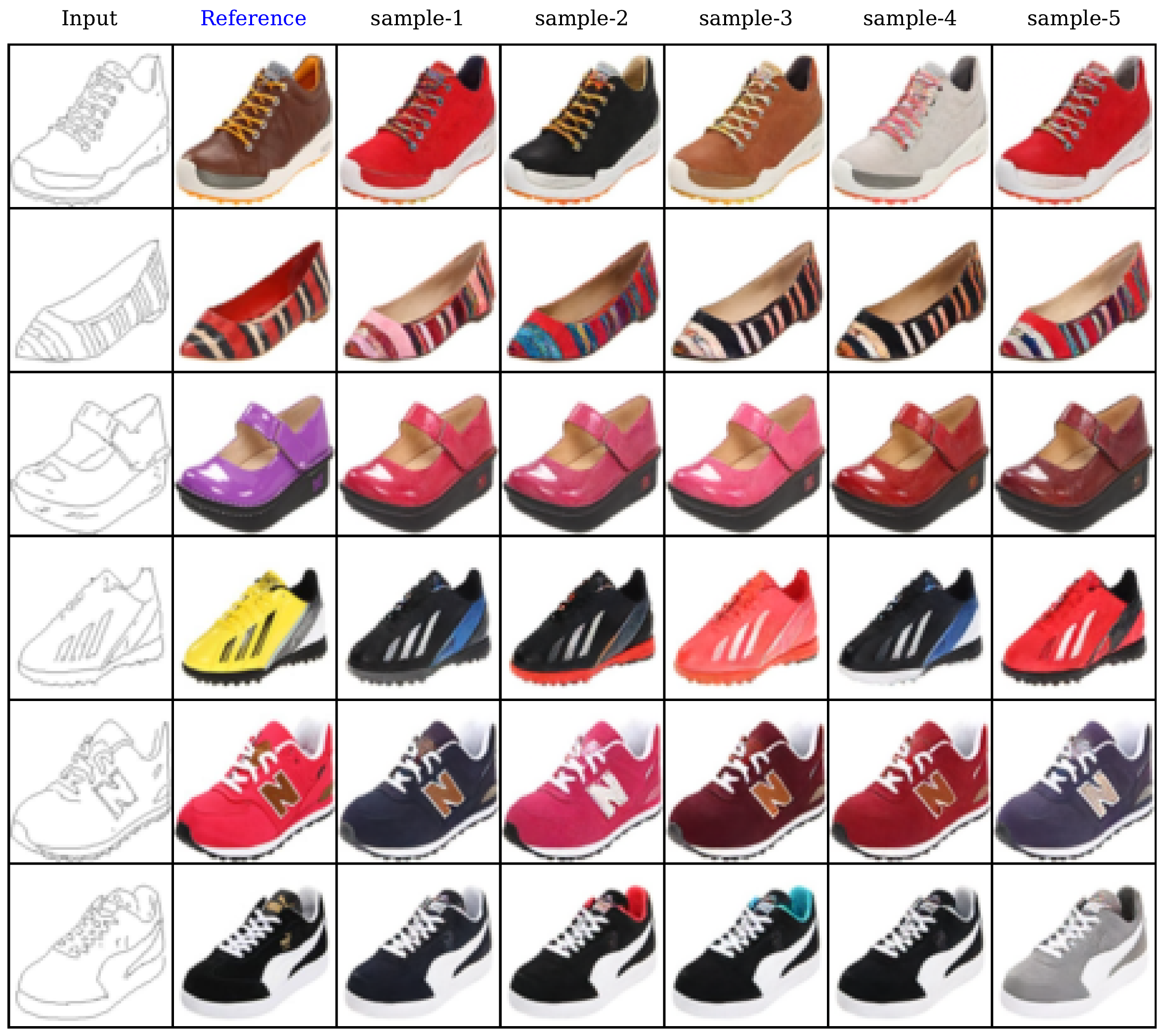} &  & \includegraphics{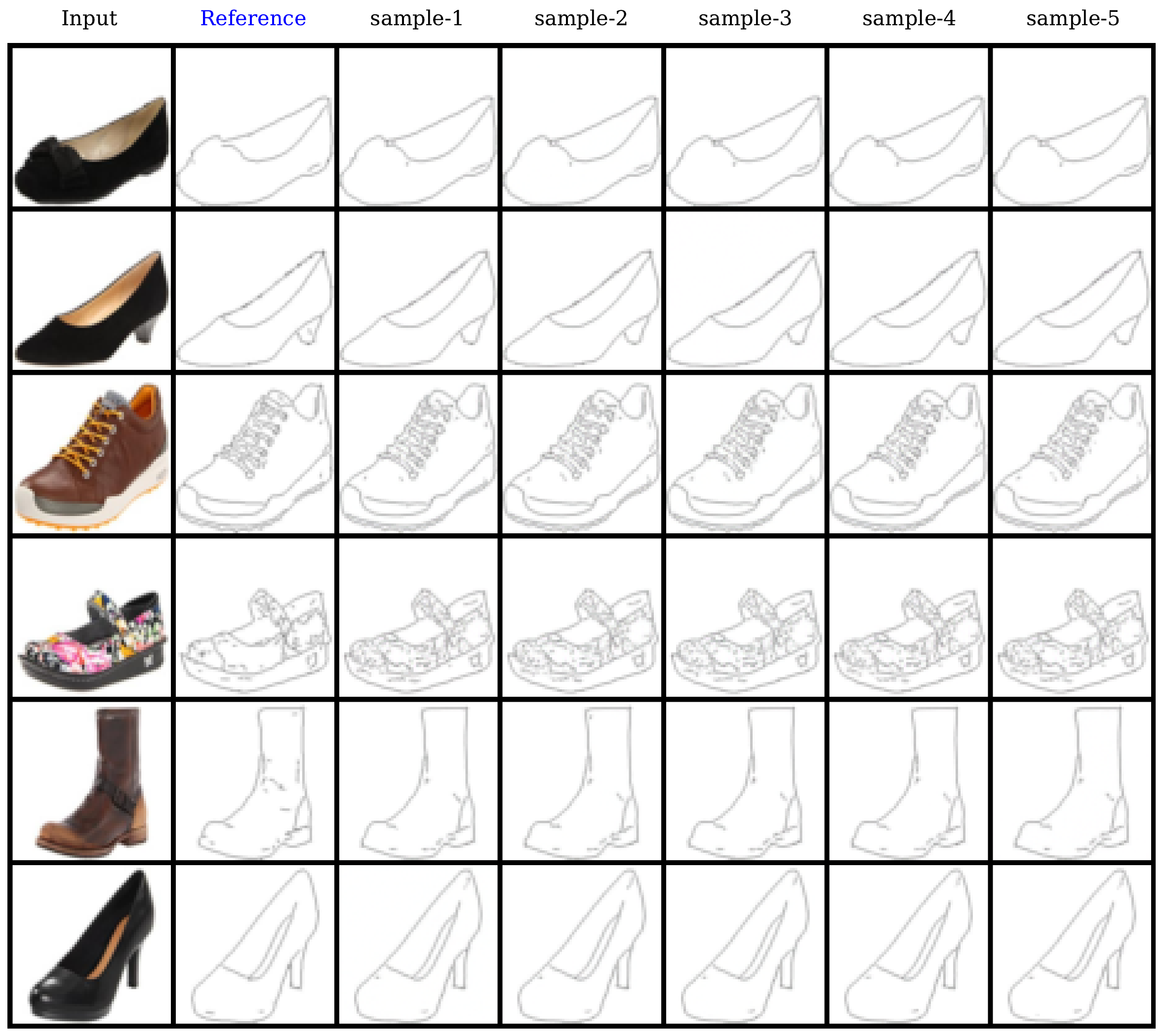}\tabularnewline
\end{tabular}}}
\par\end{centering}
\caption{Qualitative results of BDBM on a test set of Edges$\leftrightarrow$Shoes.\label{fig:Additional-results-EdgesShoes}}
\end{figure}

\noindent 
\begin{figure}
\begin{centering}
{\resizebox{1\textwidth}{!}{
\begin{tabular}{ccc}
\includegraphics{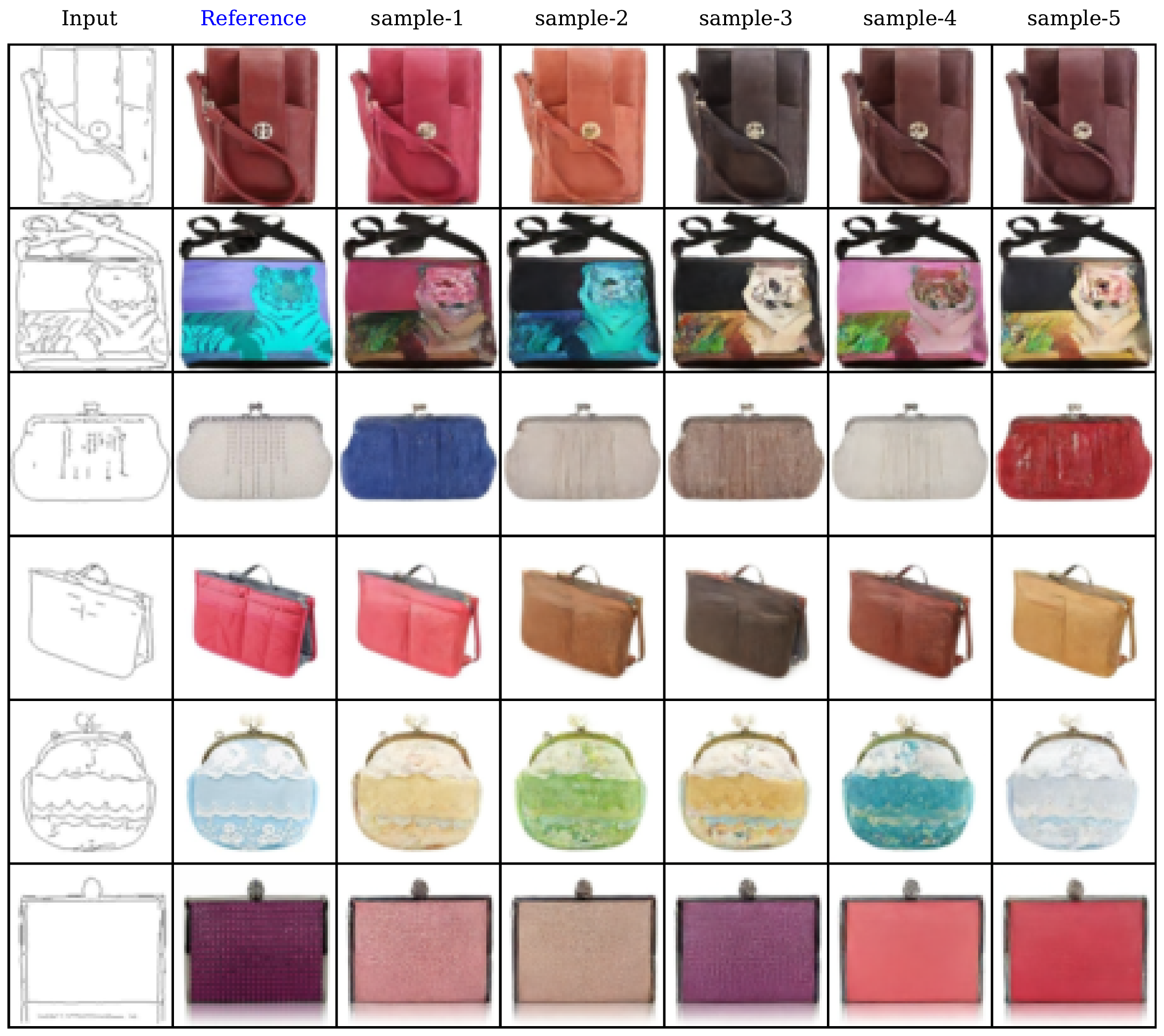} &  & \includegraphics{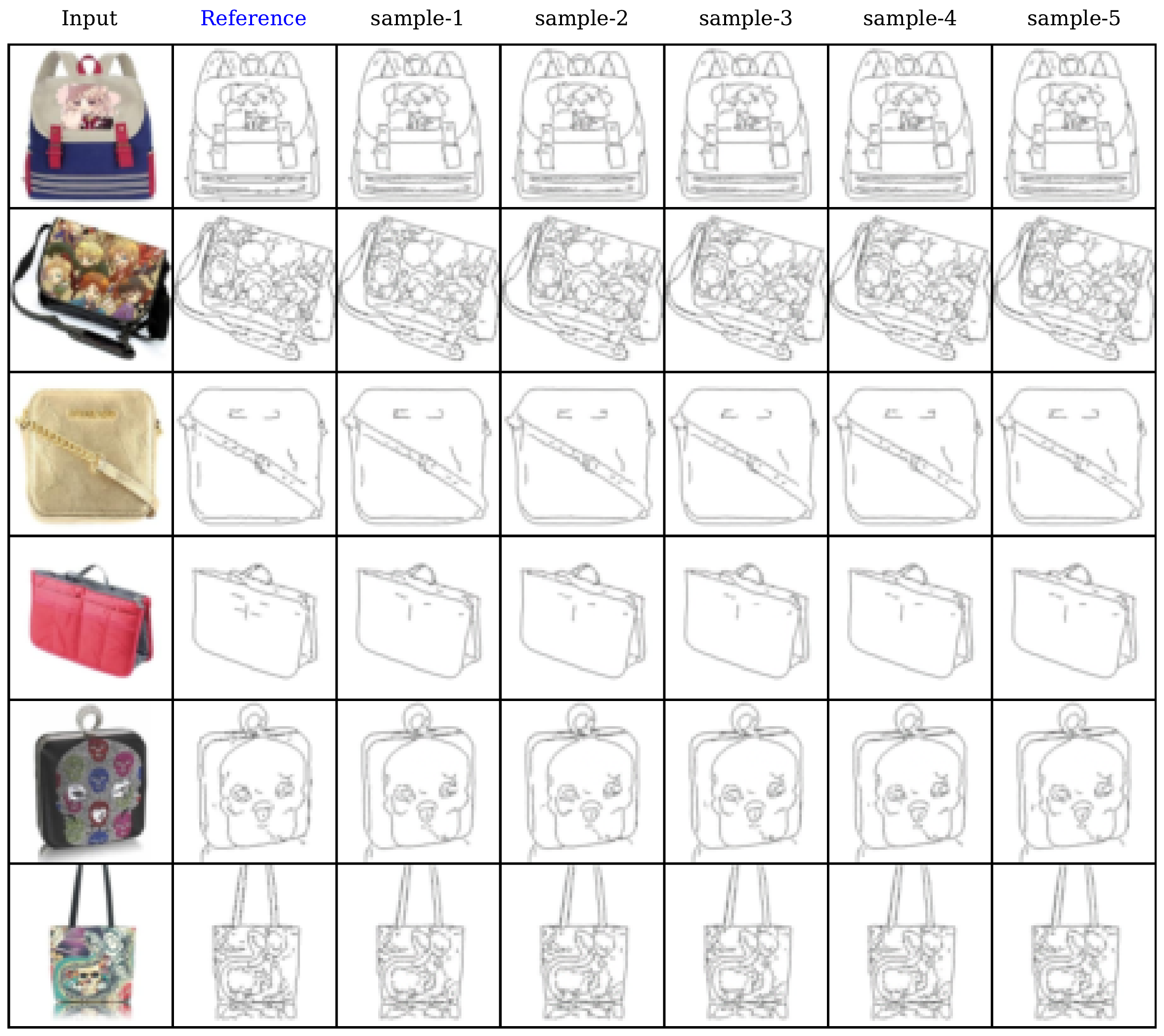}\tabularnewline
\end{tabular}}}
\par\end{centering}
\caption{Qualitative results of BDBM on a test set of Edges$\leftrightarrow$Handbag.\label{fig:Additional-results-EdgesHandbags}}
\end{figure}

\noindent 
\begin{figure}
\begin{centering}
{\resizebox{1\textwidth}{!}{
\begin{tabular}{ccc}
\includegraphics{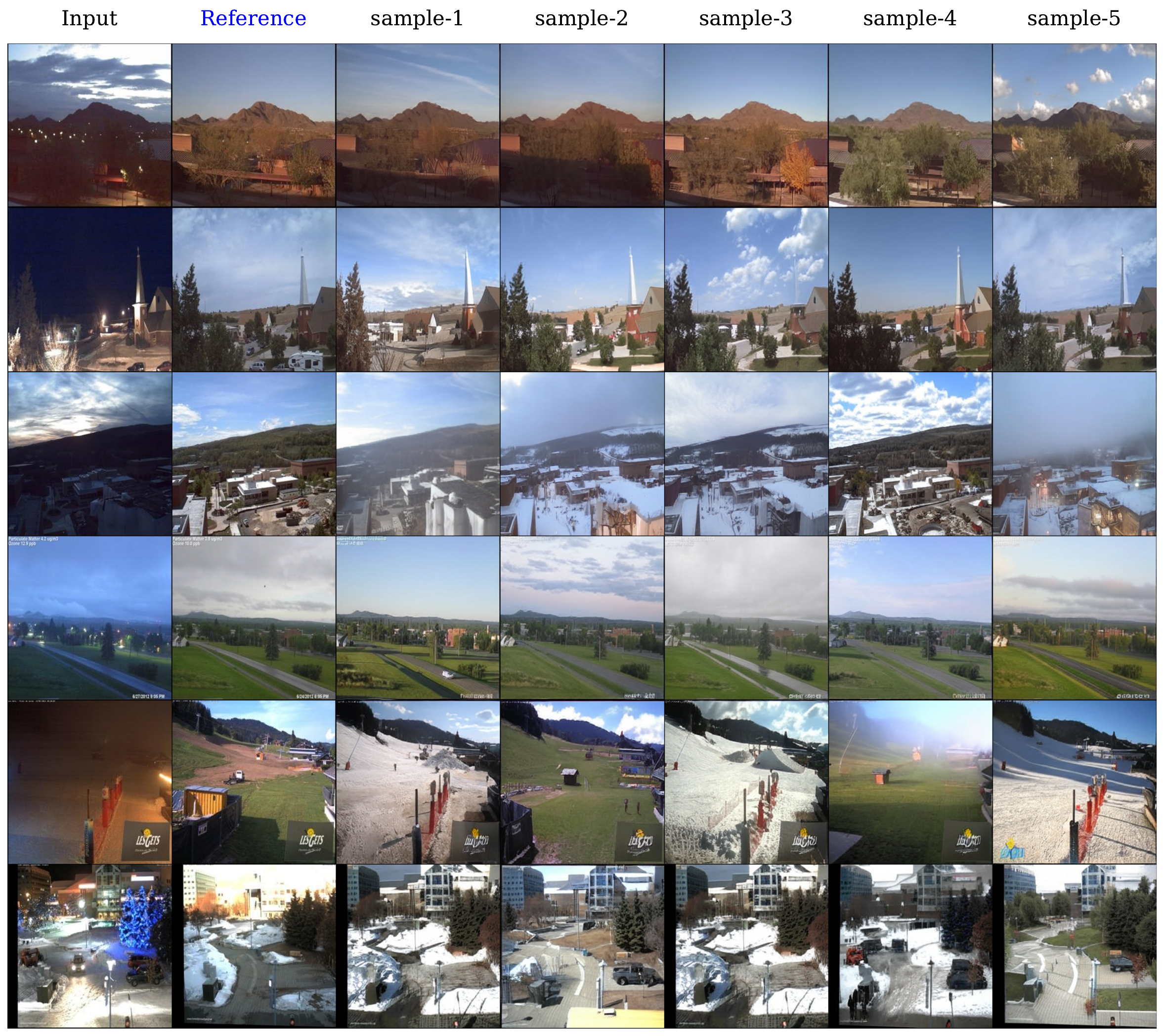} &  & \includegraphics{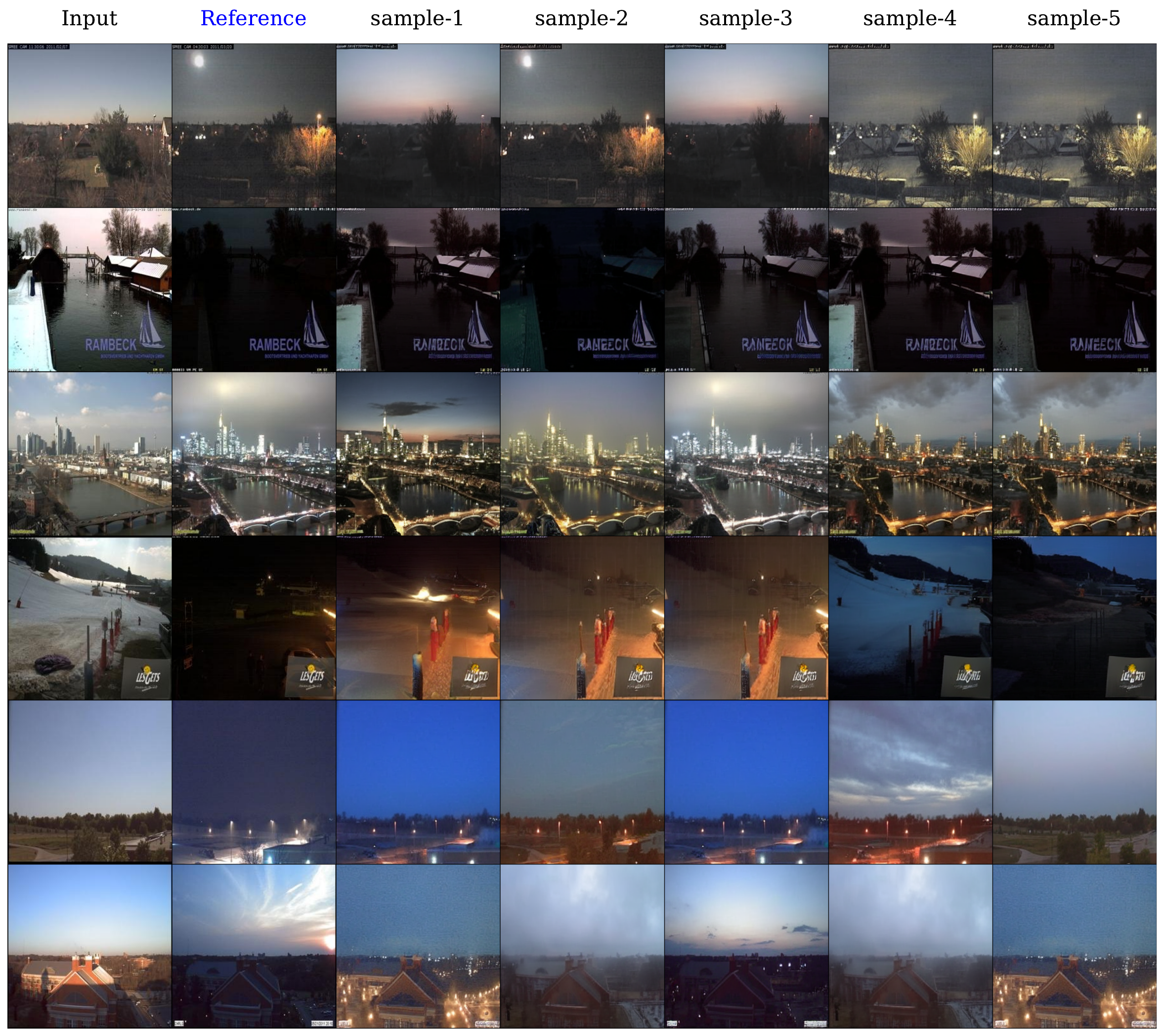}\tabularnewline
\end{tabular}}}
\par\end{centering}
\caption{Qualitative results of BDBM on a test set of Night$\leftrightarrow$Day.\label{fig:Additional-results-NightDay}}
\end{figure}

Figs.~\ref{fig:Additional-results-EdgesShoes}, \ref{fig:Additional-results-EdgesHandbags},
and \ref{fig:Additional-results-NightDay} showcase BDBM's generated
samples for both translation directions on the Edges$\leftrightarrow$Shoes,
Edges$\leftrightarrow$Handbag, and Night$\leftrightarrow$Day datasets.
Input samples are taken from a held-out test set not used during training.
The results demonstrate high-quality and diverse outputs, highlighting
BDBM's effectiveness in bidirectional translation.

\subsection{Unidirectional translation from color images to sketch/normal maps\label{subsec:Unidirectional-reverse-translation}}

\begin{table*}
\begin{centering}
\begin{tabular}{cccccccccc}
\toprule 
\multirow{2}{*}{Model} & \multicolumn{3}{c}{Shoes$\rightarrow$Edges$\times64$} & \multicolumn{3}{c}{Handbags$\rightarrow$Edges$\times64$} & \multicolumn{3}{c}{Outdoor$\rightarrow$Normal$\times256$}\tabularnewline
\cmidrule{2-10} 
 & FID $\downarrow$ & IS $\uparrow$ & LPIPS $\downarrow$ & FID $\downarrow$ & IS $\uparrow$ & LPIPS $\downarrow$ & FID $\downarrow$ & IS $\uparrow$ & LPIPS $\downarrow$\tabularnewline
\midrule
\midrule 
BBDM & \textbf{0.66} & \textbf{2.23} & \textbf{0.006} & \uline{1.54} & \textbf{3.11} & \textbf{0.010} & 18.87 & 5.82 & 0.122\tabularnewline
\midrule 
$\text{I}^{2}\text{SB}$ & 1.02 & 2.14 & 0.015 & 1.98 & 3.08 & 0.018 & \uline{11.54} & 5.97 & 0.229\tabularnewline
\midrule 
DDBM & 4.57 & 2.09 & 0.016 & 2.06 & 3.05 & 0.023 & 13.89 & \uline{6.15} & 0.237\tabularnewline
\midrule
\midrule 
BDBM-1 & \uline{0.71} & \uline{2.22} & \uline{0.007} & \textbf{1.51} & \uline{3.10} & \uline{0.011} & \textbf{9.88} & 5.98 & \textbf{0.054}\tabularnewline
\midrule 
BDBM & 0.98 & 2.20 & 0.009 & 1.87 & \uline{3.10} & 0.016 & 11.69 & \textbf{6.27} & \uline{0.069}\tabularnewline
\bottomrule
\end{tabular}
\par\end{centering}
\caption{Results of BDBM and unidirectional baselines for the color-to-sketch
and normal map translation tasks. The best results are highlighted
in bold, while the second-best results are underlined.\label{tab:quantitative_color2sketch}}
\end{table*}

\noindent 
\begin{figure}
\begin{centering}
{\resizebox{0.8\textwidth}{!}{%
\par\end{centering}
\begin{centering}
\begin{tabular}{c|c|c|c|c|c}
\textcolor{blue}{Input} & $\text{I}^{2}\text{SB}$ & BBDM & DDBM & BDBM-1 & BDBM\tabularnewline
 &  &  &  &  & \tabularnewline
\includegraphics[width=0.2\textwidth]{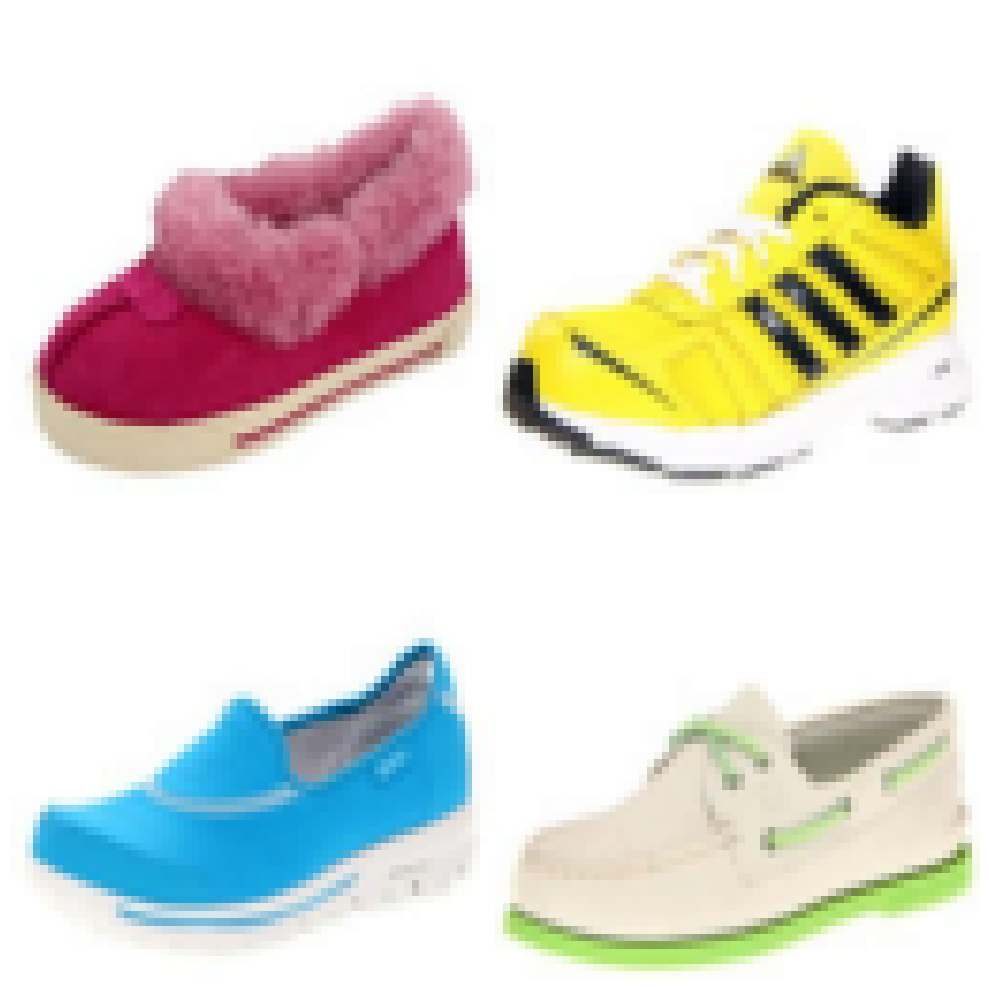} & \includegraphics[width=0.2\textwidth]{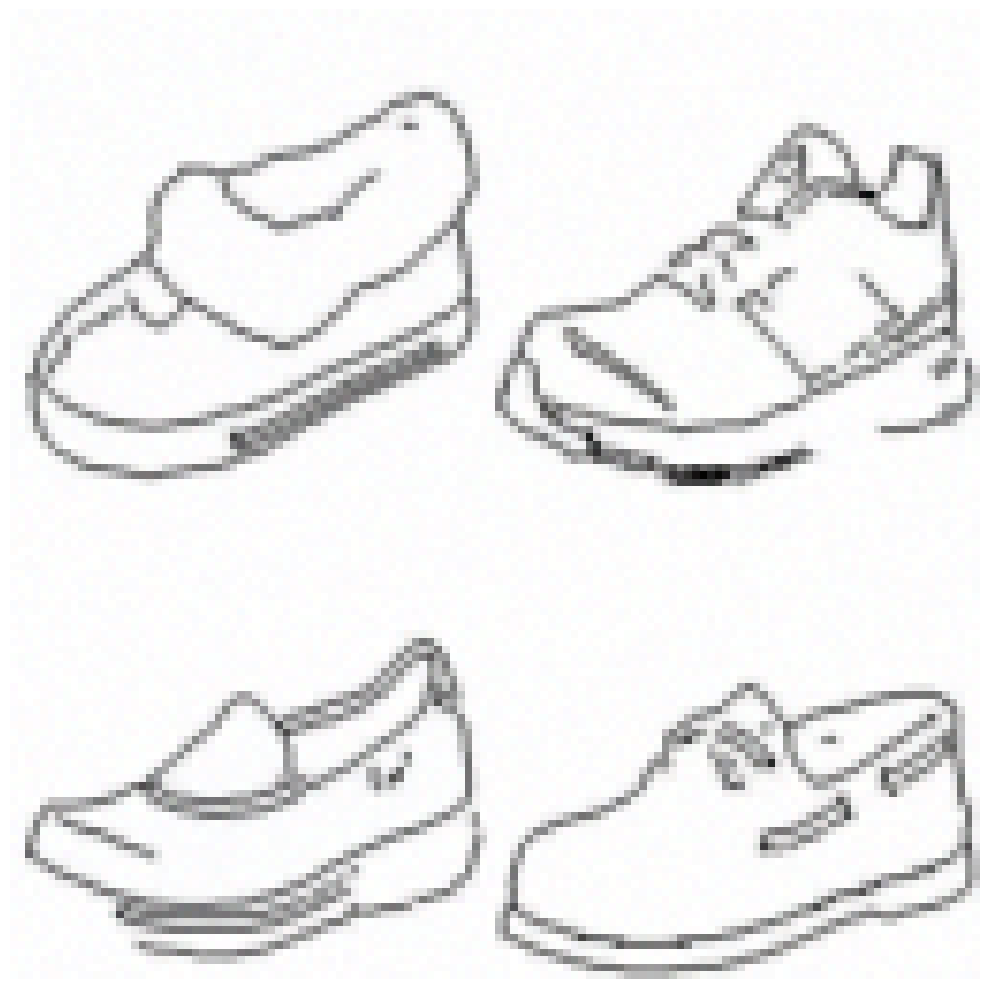} & \includegraphics[width=0.2\textwidth]{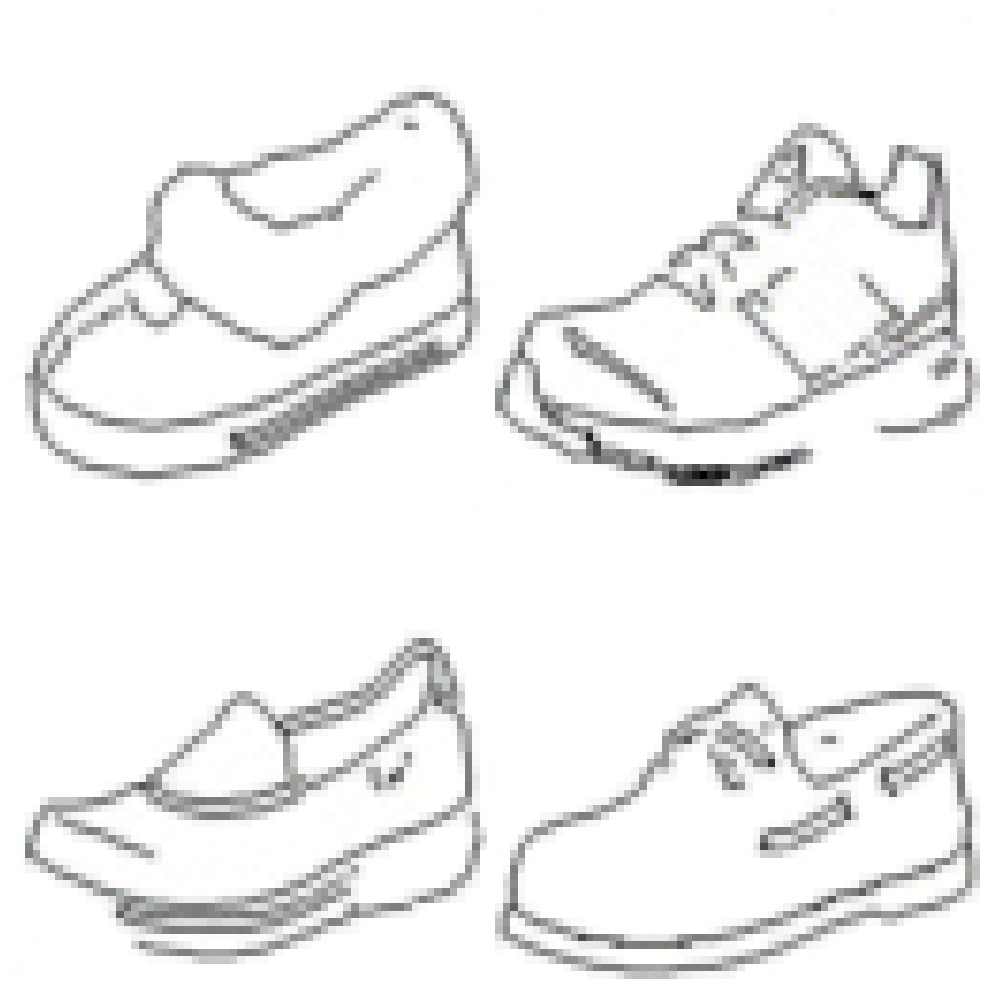} & \includegraphics[width=0.2\textwidth]{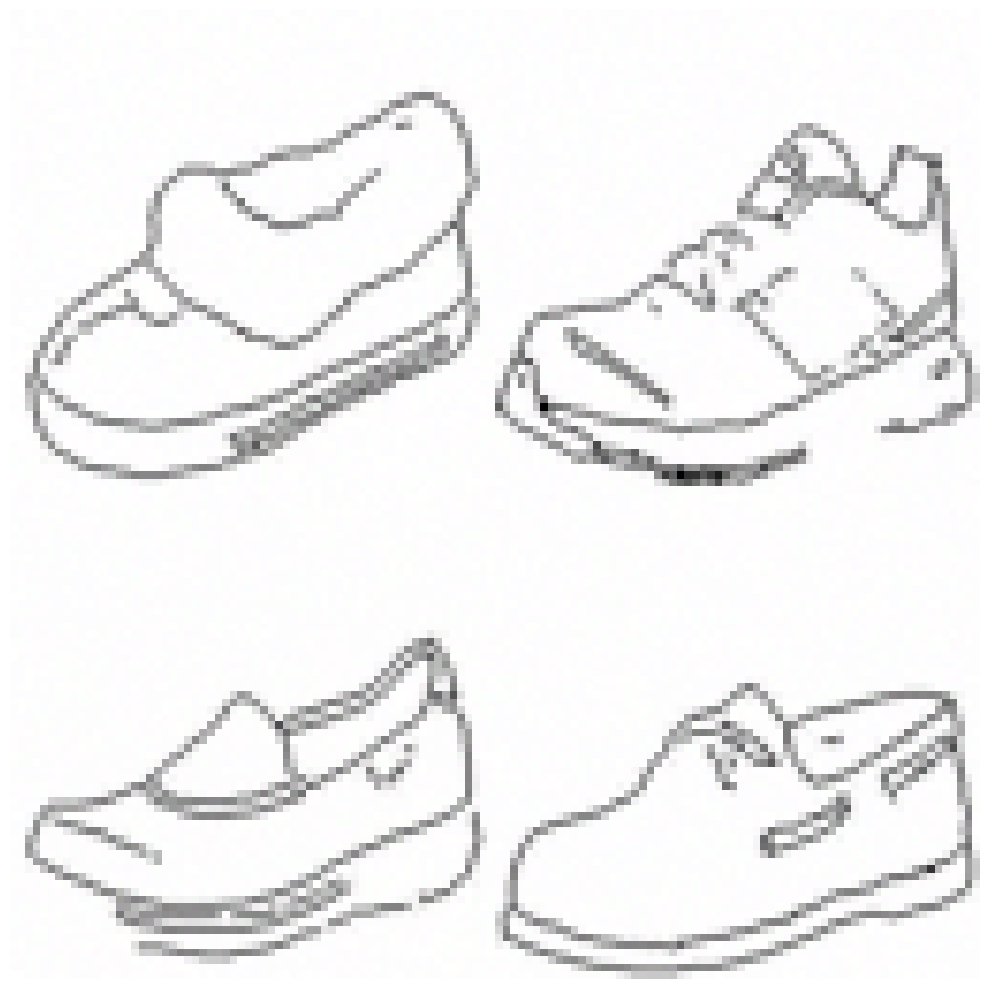} & \includegraphics[width=0.2\textwidth]{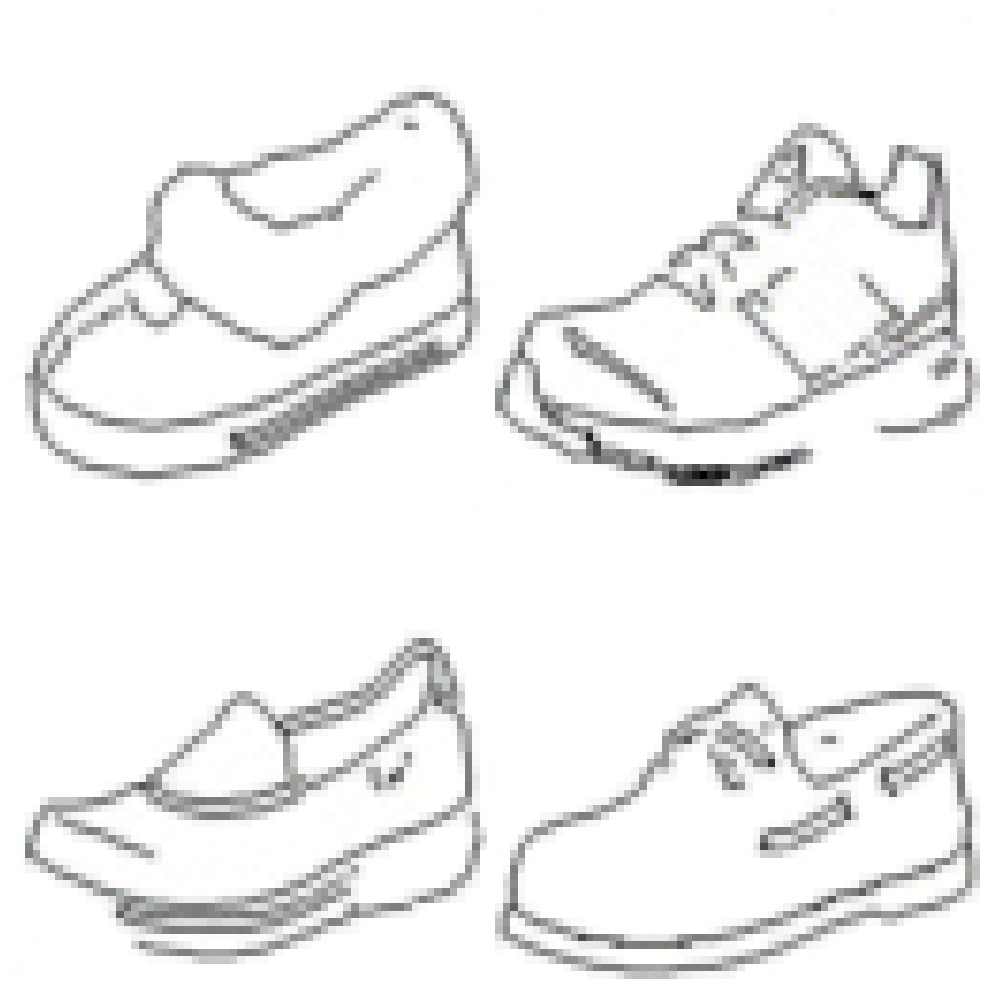} & \includegraphics[width=0.2\textwidth]{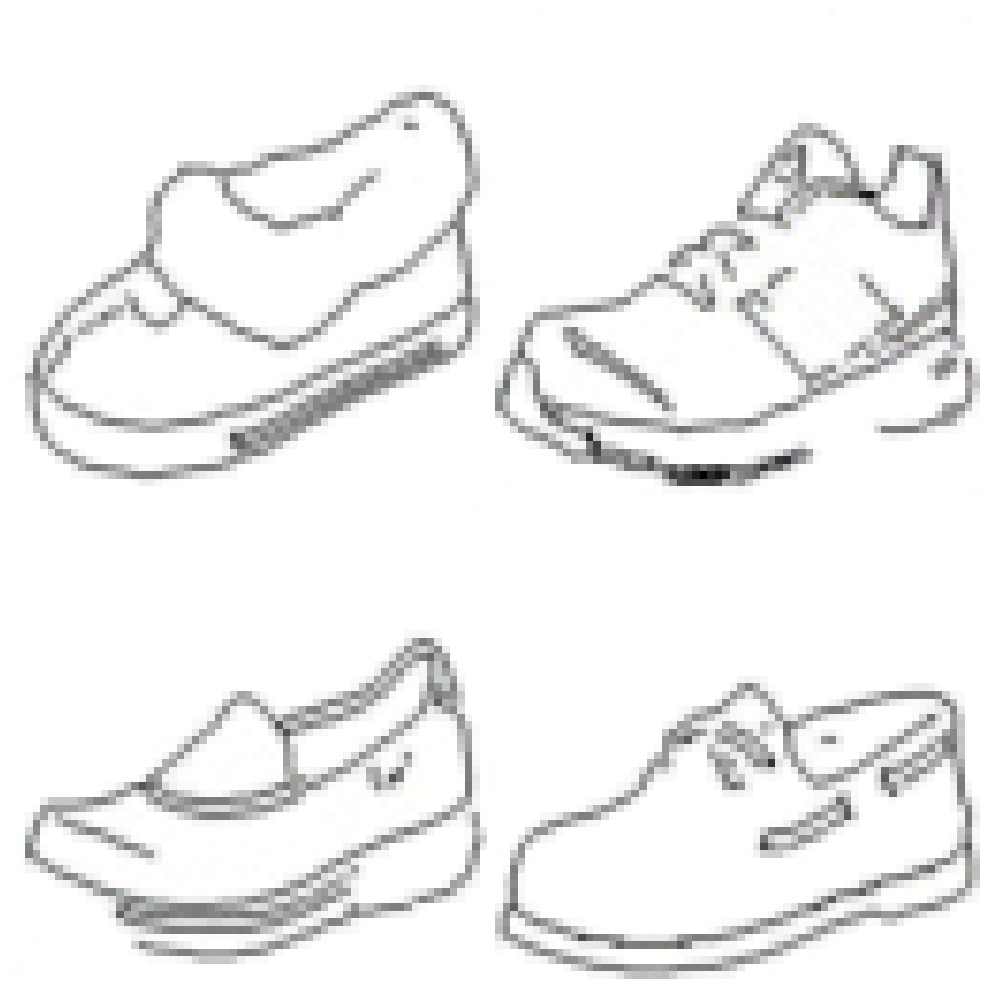}\tabularnewline
\hline 
 &  &  &  &  & \tabularnewline
\includegraphics[width=0.2\textwidth]{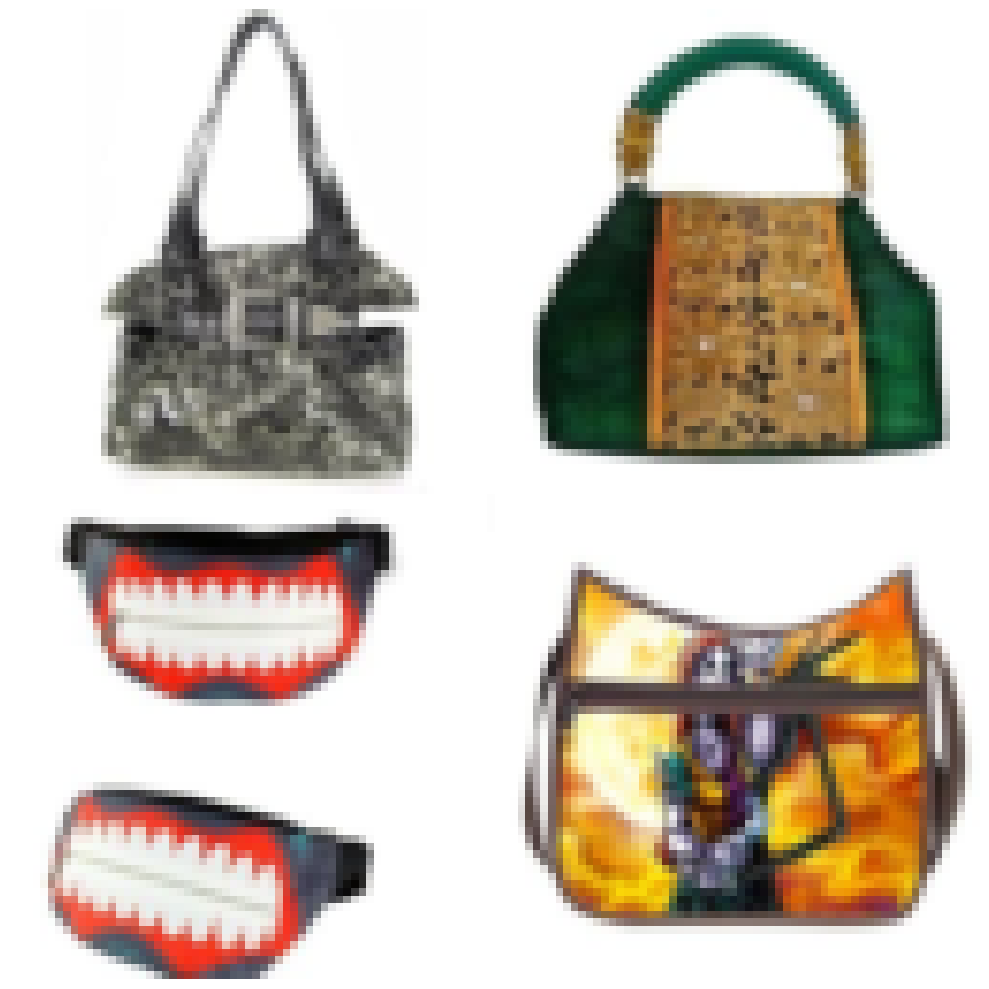} & \includegraphics[width=0.2\textwidth]{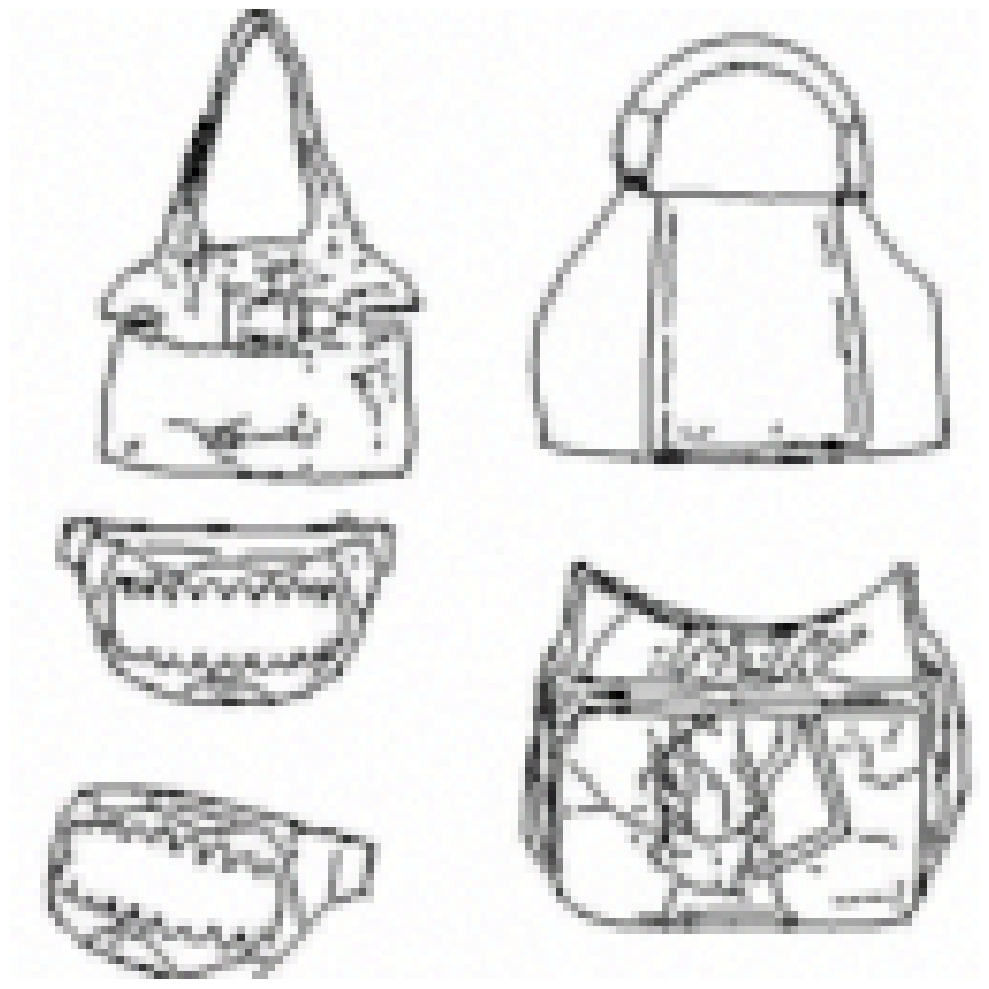} & \includegraphics[width=0.2\textwidth]{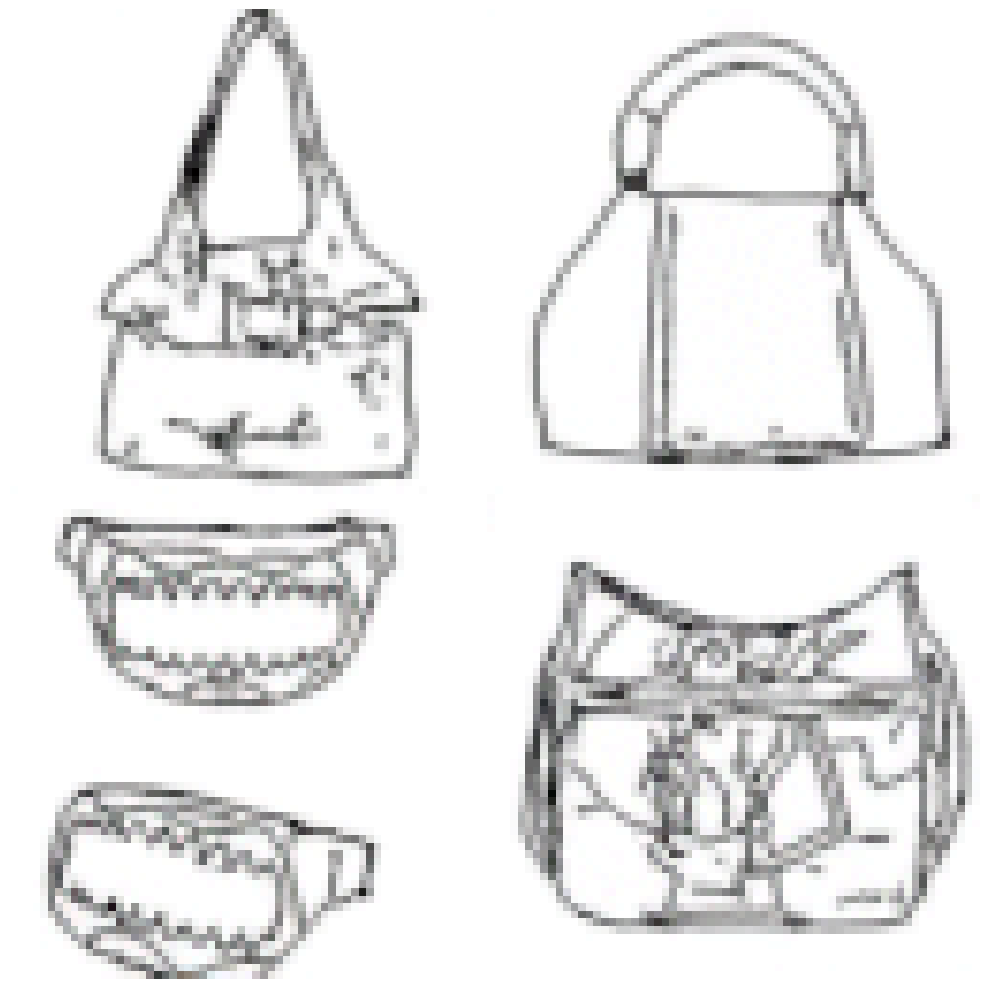} & \includegraphics[width=0.2\textwidth]{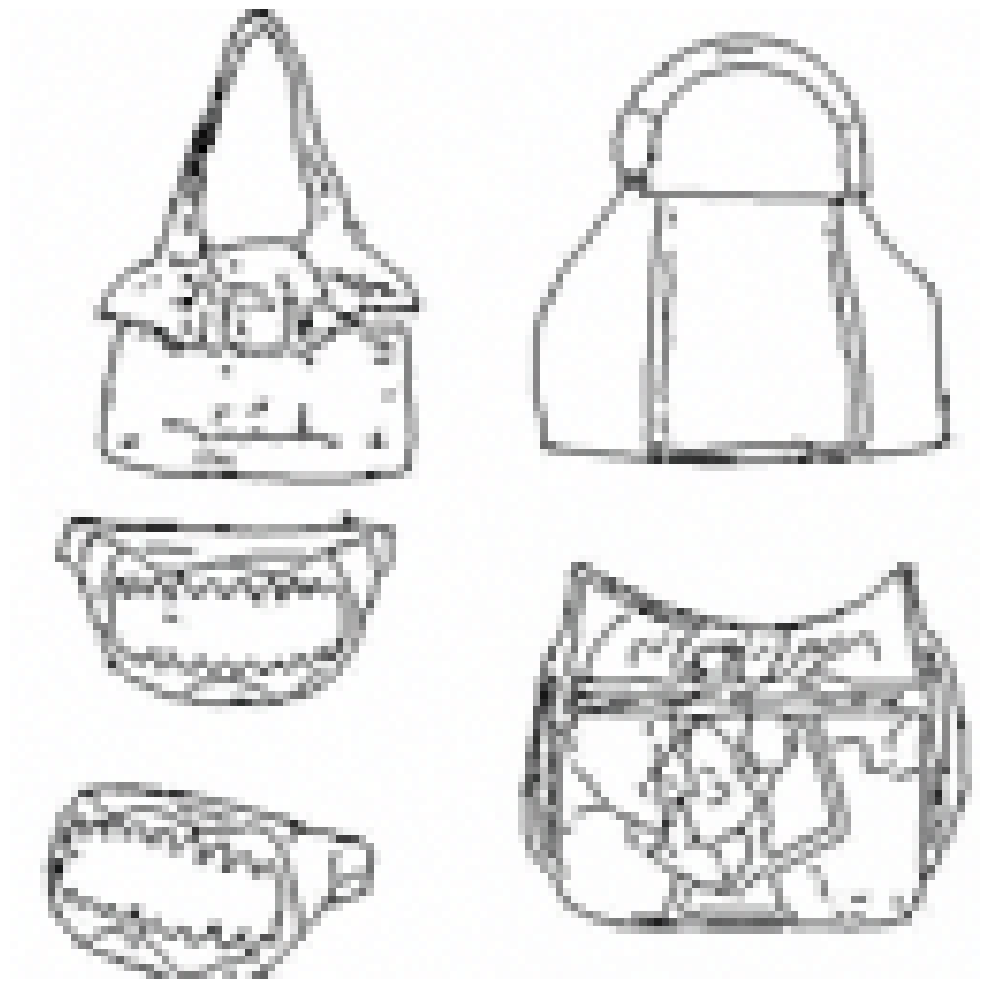} & \includegraphics[width=0.2\textwidth]{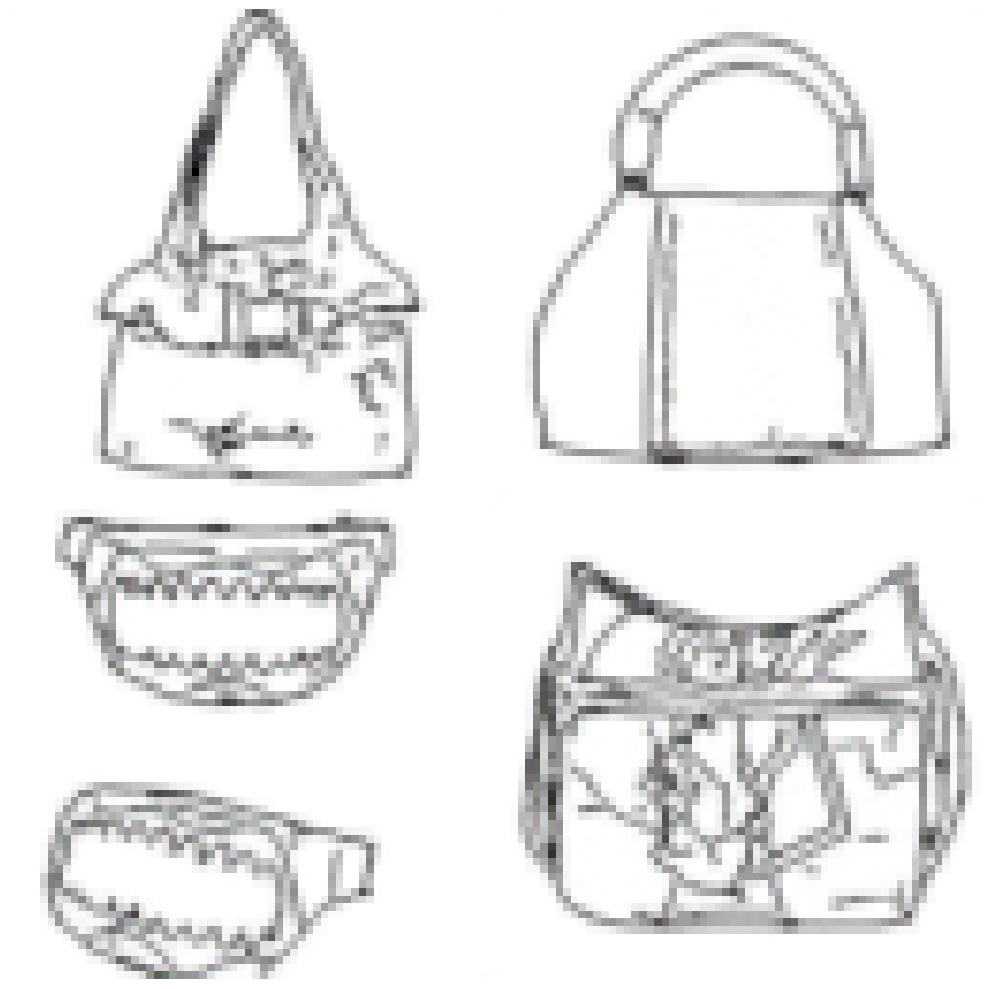} & \includegraphics[width=0.2\textwidth]{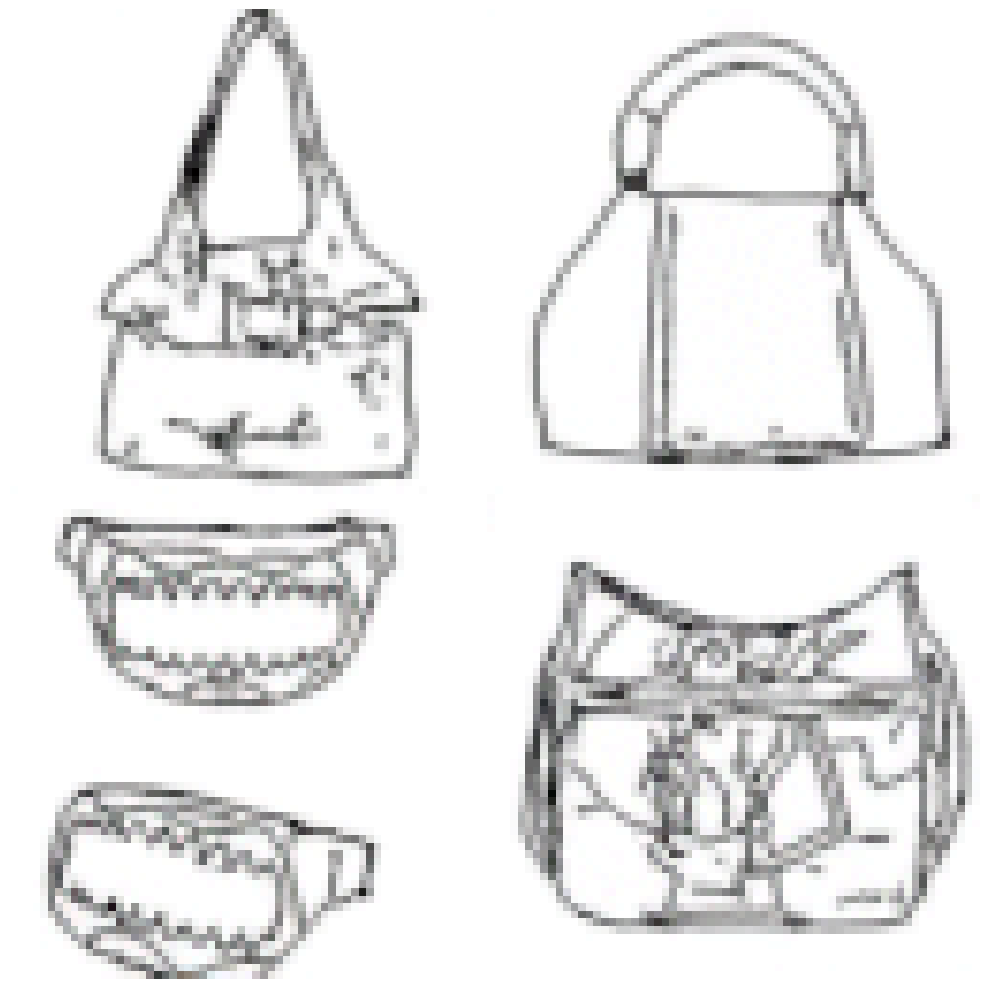}\tabularnewline
\hline 
 &  &  &  &  & \tabularnewline
\includegraphics[width=0.2\textwidth]{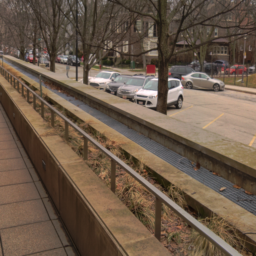} & \includegraphics[width=0.2\textwidth]{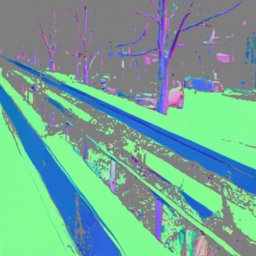} & \includegraphics[width=0.2\textwidth]{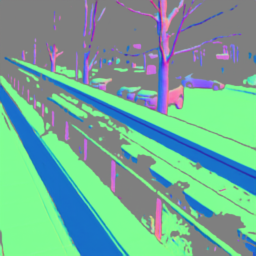} & \includegraphics[width=0.2\textwidth]{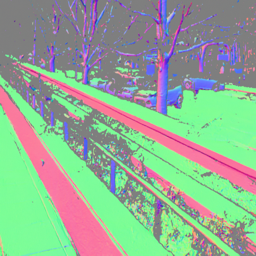} & \includegraphics[width=0.2\textwidth]{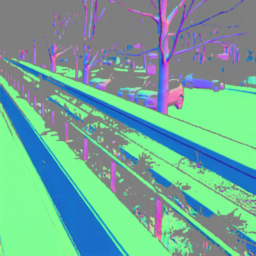} & \includegraphics[width=0.2\textwidth]{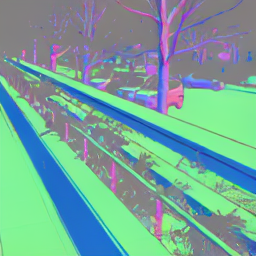}\tabularnewline
\end{tabular}}}
\par\end{centering}
\caption{Samples generated by BDBM and unidirectional baselines for color-to-sketch/normal
map translation.\label{fig:qualitative_color2sketch}}
\end{figure}

We further compare BDBM with unidirectional baselines BBDM, $\text{I}^{2}\text{SB}$,
DDBM, and BDBM-1 for color-to-sketch/normal map translation. For the
baselines, we trained new models using the same settings as described
in Section~\ref{subsec:Model-and-training} for this translation
direction, while for BDBM, we reused the checkpoints from Section~\ref{subsec:Unidirectional-I2I-Translation}
without retraining. 

As shown in Table~\ref{tab:quantitative_color2sketch}, BDBM performs
comparably to most baselines and even surpasses some on specific datasets,
despite using only \emph{half} of the training resources. Notably,
BDBM significantly outperforms DDBM and BBDM on the Shoes$\rightarrow$Edges
and Outdoor$\rightarrow$Normal datasets, respectively, highlighting
the computational efficiency of BDBM. 

Qualitative differences between methods, however, are less apparent,
as illustrated in Fig.~\ref{fig:qualitative_color2sketch}. This
is likely because sketches and normal maps contain fewer details than
color images, making the metrics \emph{more sensitive} to minor variations
even when the generated images are visually similar to the targets.

\subsection{Continuous-time BDBM vs. Discrete-time BDBM\label{subsec:Continuous-vs-Discrete}}

\begin{table}
\begin{centering}
\begin{tabular}{cccccc}
\toprule 
\multirow{2}{*}{Model} & \multirow{2}{*}{Time type} & \multicolumn{2}{c}{Edges$\leftrightarrow$Shoes$\times64$} & \multicolumn{2}{c}{Edges$\leftrightarrow$Handbags$\times64$}\tabularnewline
\cmidrule{3-6} 
 &  & FID $\downarrow$ & LPIPS $\downarrow$ & FID $\downarrow$ & LPIPS $\downarrow$\tabularnewline
\midrule
\midrule 
\multirow{2}{*}{BDBM-1} & discrete-time & 0.71/1.78 & 0.01/0.07 & 1.51/3.83 & 0.01/0.11\tabularnewline
 & continuous-time & 1.28/1.81 & 0.01/0.03 & 2.45/3.94 & 0.02/0.17\tabularnewline
\midrule
\midrule 
\multirow{2}{*}{BDBM} & discrete-time  & 0.98/1.06 & 0.01/0.02 & 1.87/3.06 & 0.02/0.08\tabularnewline
 & continuous-time & 2.38/2.41 & 0.01/0.04 & 2.88/3.79 & 0.04/0.16\tabularnewline
\bottomrule
\end{tabular}
\par\end{centering}
\caption{Comparison between discrete-time BDBM and continuous-time BDBM.\label{tab:disc_vs_cont_BDBM}}
\end{table}

\begin{figure}
\begin{centering}
\subfloat[Comparison between discrete-time BDBM and continuous-time BDBM on
Edges$\rightarrow$Shoes$\times64$.]{\begin{centering}
\includegraphics[width=1\textwidth]{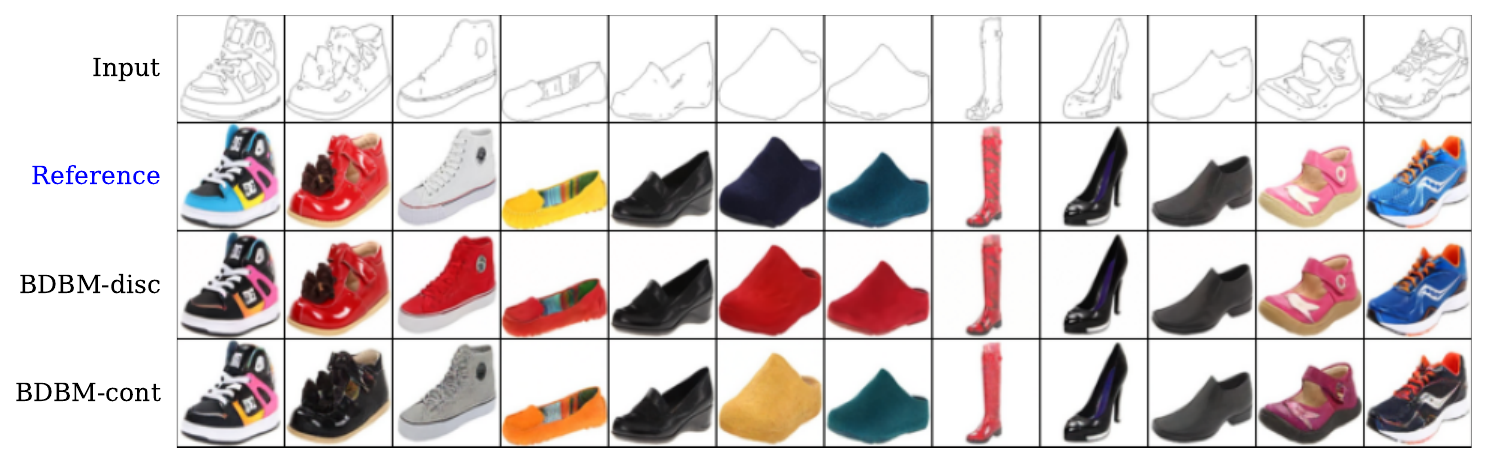}
\par\end{centering}
}
\par\end{centering}
\begin{centering}
\subfloat[Comparison between discrete-time BDBM and continuous-time BDBM on
Edges$\rightarrow$Handbags$\times64$.]{\centering{}\includegraphics[width=1\textwidth]{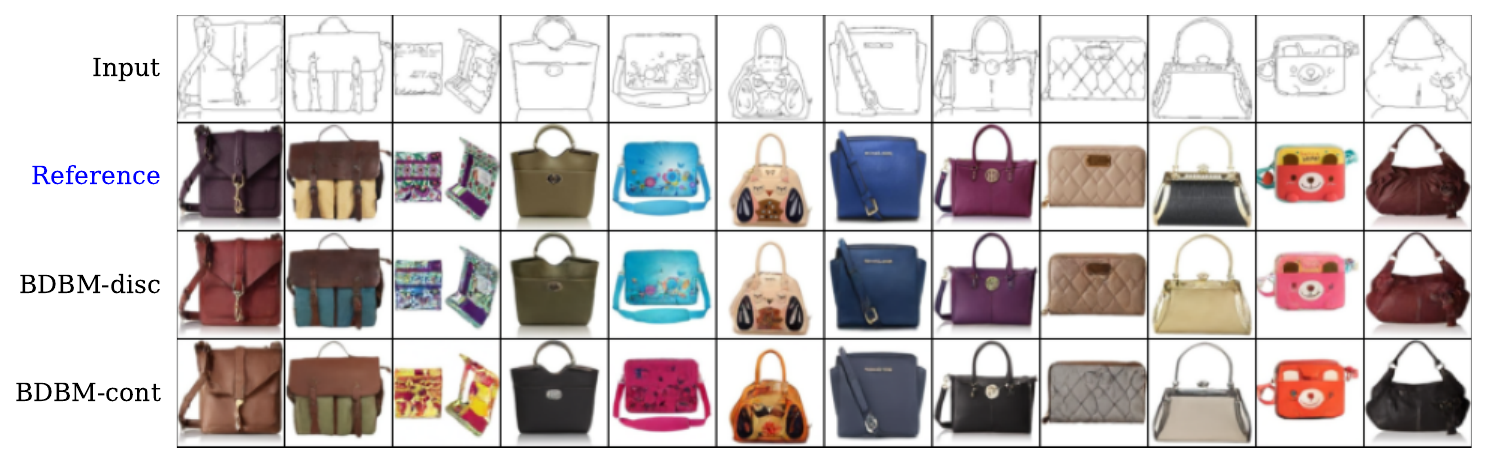}}
\par\end{centering}
\caption{Visualization of discrete-time BDBM and continuous-time BDBM accross
Edges$\rightarrow$Shoes$\times64$ and Edges$\rightarrow$Handbags$\times64$.
The first row shows the input images, the second row presents the
ground truth images, while the third and fourth rows display the outputs
of discrete-time and continuous-time BDBM, respectively. \label{fig:disc_vs_cont_BDBM_vis}}
\end{figure}

In this section, we compare discrete-time BDBM with its continuous-time
counterpart. Both models are evaluated under identical settings, except
that the continuous-time model allows $t$ to take any real value
in $\text{\ensuremath{\left[0,1\right]}}$, while the discrete-time
model restricts $t$ to integer values in $\left[0,1000\right]$. 

As shown in Table~\ref{tab:disc_vs_cont_BDBM} and Fig.~\ref{fig:disc_vs_cont_BDBM_vis},
discrete-time BDBM consistently outperforms its continuous-time counterpart.
The primary reason for this advantage is that discrete-time BDBM only
needs to predict noise for a fixed set of time steps, whereas the
continuous-time model must handle an infinite number of time steps.
As a result, given the same number of training iterations, discrete-time
BDBM can allocate more iterations to refining noise prediction at
each specific time step, leading to more accurate predictions. This
highlights the advantage of the discrete-time model when training
iterations are limited. However, we anticipate that with a sufficient
number of training iterations (as used for training continuous-time
diffusion models \cite{song2020score}), both models would likely
achieve comparable results.

\subsection{More visualization on generated samples by BDBM}

\begin{figure}
\begin{centering}
\includegraphics[width=1\textwidth]{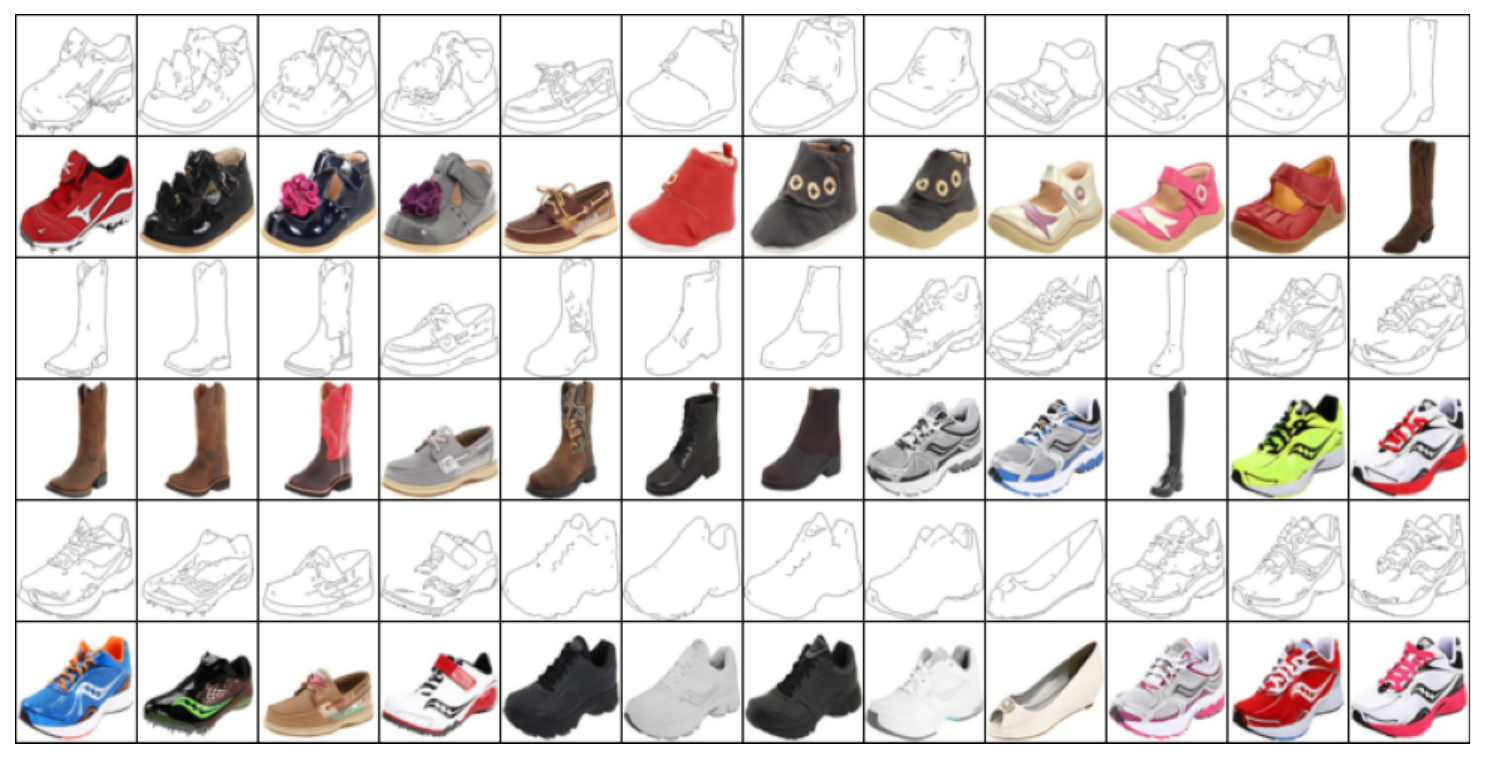}
\par\end{centering}
\caption{Additional qualitative results for Edges$\rightarrow$Shoes$\times64$,
where each pair of consecutive rows displaying the input image in
the ``Edges'' domain and its translation in the ``Shoes'' domain,
respectively.\label{fig:e2s_appdx}}

\end{figure}

\begin{figure}[H]
\begin{centering}
\includegraphics[width=1\textwidth]{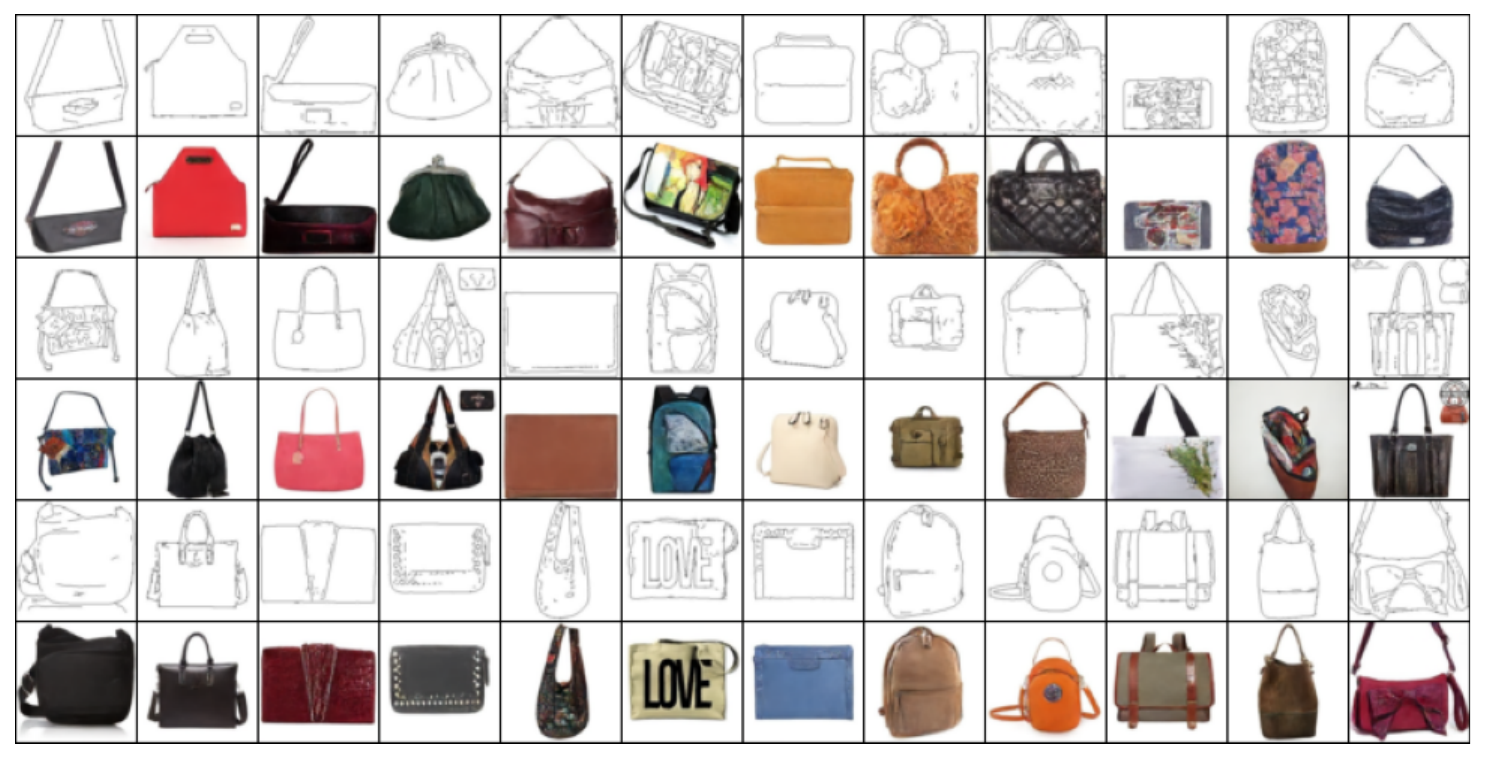}\caption{Additional qualitative results for Edges$\rightarrow$Handbags$\times64$,
where each pair of consecutive rows displaying the input image in
the ``Edges'' domain and its translation in the ``Handbags'' domain,
respectively.\label{fig:e2h_appdx}}
\par\end{centering}
\end{figure}

\begin{figure}[H]
\begin{centering}
\includegraphics[width=0.85\textwidth]{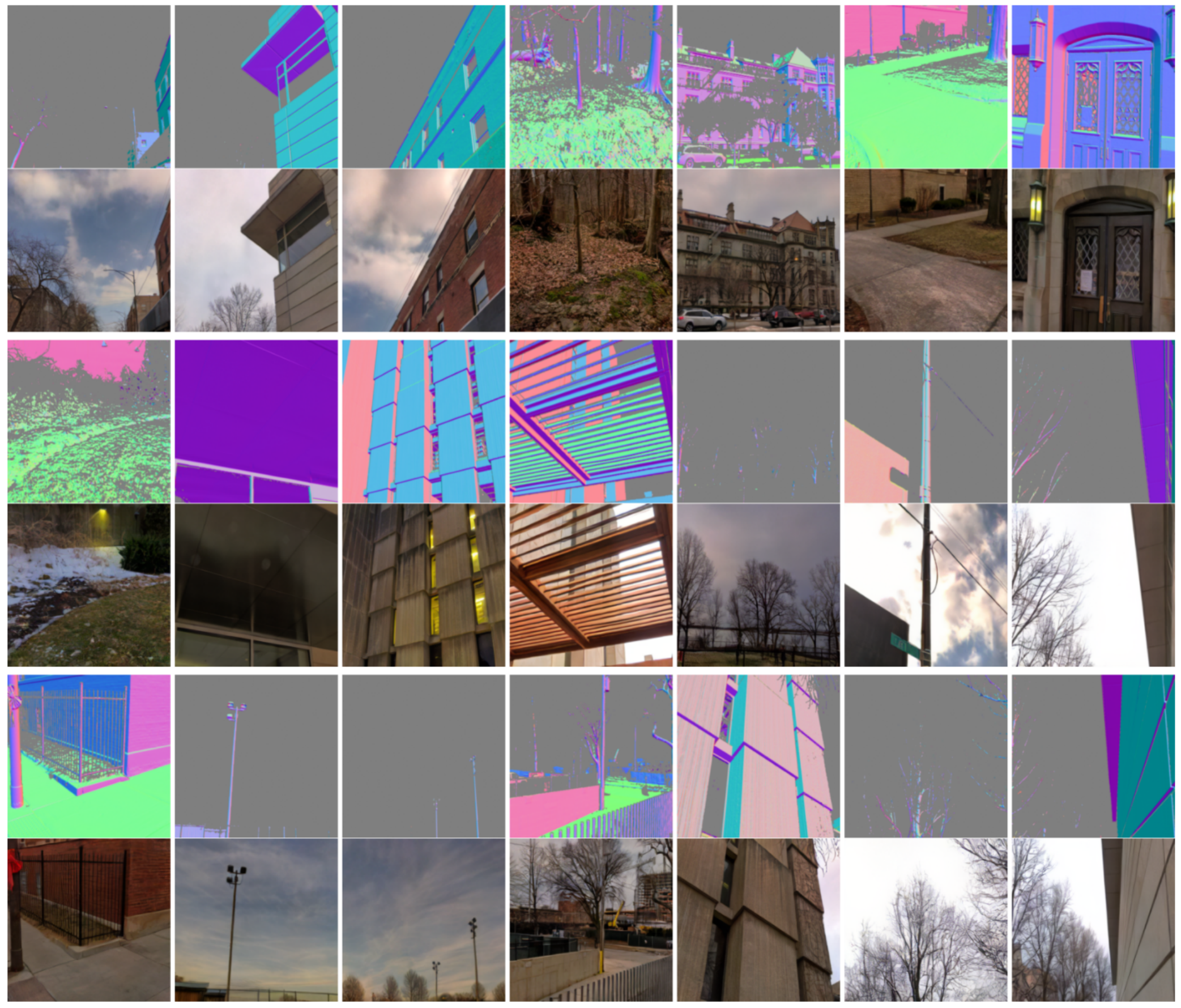}
\par\end{centering}
\caption{Additional qualitative results for DIODE Outdoor$\times256$, where
each pair of consecutive rows displaying the input image in the ``Normal
maps'' domain and its translation in the ``Color images'' domain,
respectively.\label{fig:diode_appdx}}
\end{figure}

We provide additional qualitative translation results for Edges$\rightarrow$Shoes$\times64$,
Edges$\rightarrow$Handbags$\times64$, and DIODE Outdoor$\times256$,
in Figs.~\ref{fig:e2s_appdx}, \ref{fig:e2h_appdx}, and \ref{fig:diode_appdx},
respectively.

\newpage{}

\end{document}